\documentclass[10pt,twocolumn,letterpaper]{article}

\usepackage{cvpr}              
\usepackage{amsmath,amssymb,amsthm,mathtools,bm,dsfont}
\usepackage{tikz-cd}
\usepackage{enumitem}
\usepackage{microtype}
\usepackage{siunitx}    
\usepackage{colortbl}   
\usepackage{tikz}       
\usepackage{multirow}
\usepackage{adjustbox}
\usepackage{tcolorbox}
\usepackage{fontawesome5}

\tcbuselibrary{skins}

\definecolor{findinggreen}{HTML}{E8F5E9}
\definecolor{findingbar}{HTML}{4CAF50}
\definecolor{findingicon}{HTML}{7CB342}
\definecolor{findingborder}{HTML}{A5D6A7}

\newtcolorbox{finding}{
  enhanced,
  colback=findinggreen,
  colframe=findinggreen,
  boxrule=0pt,
  borderline west={3pt}{0pt}{findingbar},
  arc=0pt,
  left=10pt, right=8pt, top=6pt, bottom=6pt,
  fontupper=\small,
  overlay={
    \node[draw=findingborder, fill=white, circle, inner sep=3pt, line width=0.6pt] 
      at (frame.north west) {\textcolor{findingicon}{\small\faLightbulb}};
  }
}

\definecolor{cvprblue}{rgb}{0.21,0.49,0.74}
\usepackage[pagebackref,breaklinks,colorlinks,allcolors=cvprblue]{hyperref}

\def\paperID{*****} 
\def\confName{CVPR}
\def\confYear{2026}

\title{Can Cross-Layer Transcoders Replace Vision Transformer Activations? \\ An Interpretable Perspective on Vision}

\author{
Gerasimos Chatzoudis$^{1}$\quad
Konstantinos D. Polyzos$^{2}$\quad
Zhuowei Li$^{1}$\thanks{Work done outside of Amazon.}\\
Difei Gu$^{1}$\quad
Gemma E. Moran$^{1}$\quad
Hao Wang$^{1}$\quad
Dimitris N. Metaxas$^{1}$\\[6pt]
$^{1}$Rutgers University\quad
$^{2}$University of California San Diego\\
{\tt\small gc745@scarletmail.rutgers.edu}\quad
{\tt\small kpolyzos@ucsd.edu}\quad
{\tt\small zl502@cs.rutgers.edu}\\
{\tt\small difei.gu@rutgers.edu}\quad
{\tt\small gm845@stat.rutgers.edu}\quad
{\tt\small hw488@cs.rutgers.edu}\quad
{\tt\small dnm@cs.rutgers.edu}
}

\begin{document}
\maketitle
\begin{abstract}
Understanding the internal activations of Vision Transformers (ViTs) is critical for building interpretable and trustworthy models. While Sparse Autoencoders (SAEs) have been used to extract human-interpretable features, they operate on individual layers and fail to capture the cross-layer computational structure of Transformers, as well as the relative significance of each layer in forming the last-layer representation. Alternatively, we introduce the adoption of Cross-Layer Transcoders (CLTs) as reliable, sparse, and depth-aware proxy models for MLP blocks in ViTs. CLTs use an encoder–decoder scheme to reconstruct each post-MLP activation from learned sparse embeddings of preceding layers, yielding a linear decomposition that transforms the final representation of ViTs from an opaque embedding into an additive, layer-resolved construction that enables faithful attribution and process-level interpretability. We train CLTs on CLIP ViT-B/32 and ViT-B/16 across CIFAR-100, COCO, and ImageNet-100. We show that CLTs achieve high reconstruction fidelity with post-MLP activations while preserving and even improving, in some cases, CLIP zero-shot classification accuracy. In terms of interpretability, we show that the cross-layer contribution scores provide faithful attribution, revealing that the final representation is concentrated in a smaller set of dominant layer-wise terms whose removal degrades performance and whose retention largely preserves it. These results showcase the significance of adopting CLTs as an alternative interpretable proxy of ViTs in the vision domain. 
\end{abstract}
    
\section{Introduction}

\begin{figure*}[t]
  \centering
  \includegraphics[width=\linewidth]{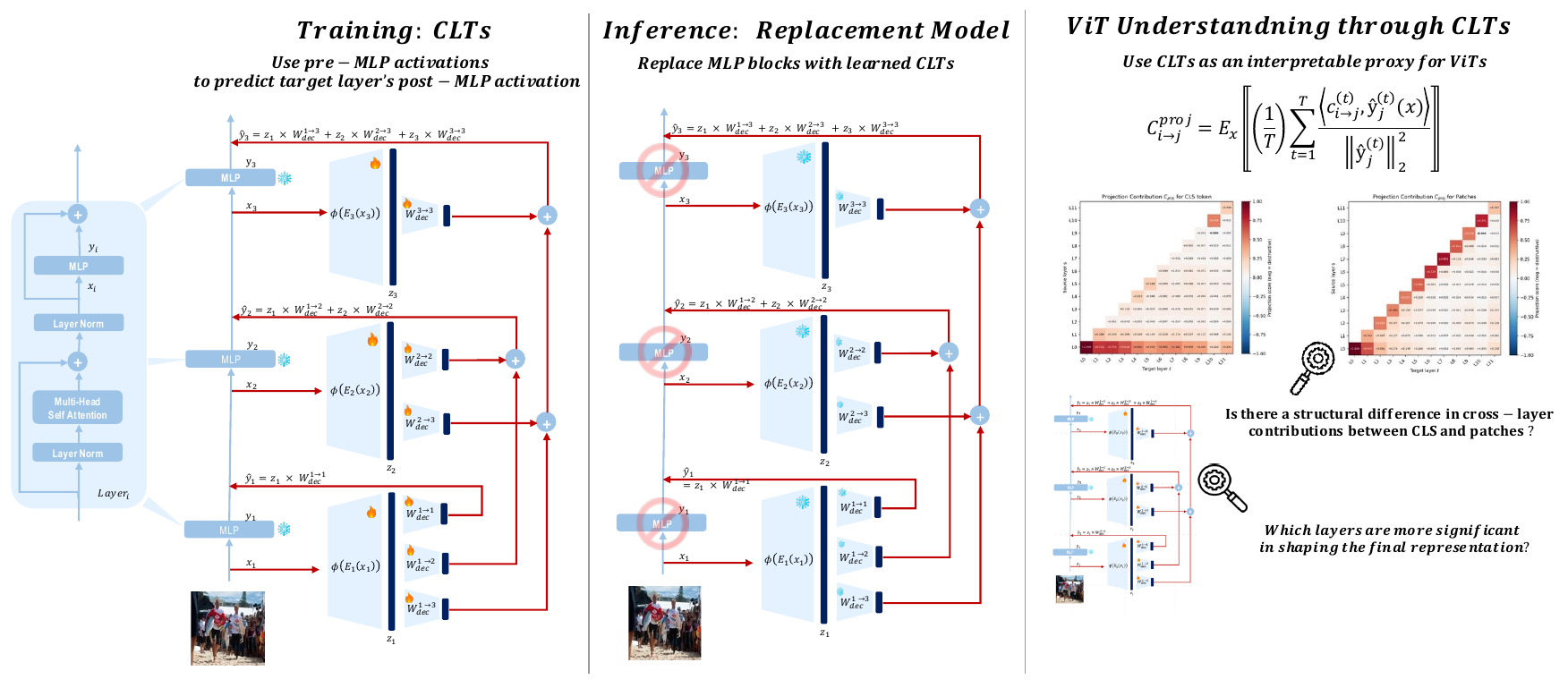}
  \caption{Overview of the Cross-Layer Transcoder (CLT) framework.
\textbf{Left}: Each CLT encodes LN2 activations into sparse codes $z_\ell$ and reconstructs MLP outputs $y_\ell$ via triangular decoders.
\textbf{Middle}: During inference, CLTs replace MLPs across layers, preserving zero-shot performance.
\textbf{Right}: Using CLTs as an interpretable proxy for ViTs to understand the cross-layer contributions of different token types and identify the most significant layers in shaping the ViT's final representations.}
  \label{fig:iz-trace}
\end{figure*}

Foundation Models (FMs) have shown strong generalization performance in diverse tasks ranging from classification to open-ended generation \citep{clip,sam,llama2,llava,gpt4}.
Despite their  empirical success, these models remain largely opaque, limiting interpretability as well as controllability and reliability in practice.
Particularly on the vision domain, understanding the internal representations of the widely-used Vision Transformers (ViTs) remains an open challenge for building interpretable, controllable, and verifiable models. Although existing approaches, including Sparse Autoencoders ~\cite{lim2024sparse, stevens2025sparse, bricken2023monosemanticity} have aimed to show that transformer activations can be organized into human-interpretable components, they operate \emph{locally}, learning features only within a single layer. Consequently, they fall short in capturing cross-layer interdependencies or the relative contribution of each layer to the final-layer representation, providing limited insight into how information is transformed \emph{across} network depth.

Aiming to advance interpretable vision models and inspired by their use in the language domain, we propose a novel interpretable perspective on vision by leveraging Cross-Layer Transcoders (CLTs). Specifically, our goal is to judiciously integrate CLTs as an alternative framework for analyzing the internal structure of transformer representations within Vision Transformers (ViTs). 
CLTs aim to reconstruct the post-MLP activations at a target layer from sparse features computed in earlier layers enabling a directed, cross-layer decomposition of each representation. This allows for a proper assessment of the contribution of each source layer to a given target layer.
When CLTs are accurate enough to replace the original MLP blocks, the resulting CLT-based `replacement' model can serve as a reliable proxy for analyzing the original transformer without compromising the ViT model performance on the downstream task.

While CLTs have been used to identify attribution graphs in Large Language Models (LLMs) \cite{ameisen2025circuittracing}, to the best of our knowledge their proper adoption for vision models as interpretable alternatives has not been explored yet.  Unlike autoregressive language models in the language domain, where all tokens
share a uniform sequential structure, ViTs operate over spatially structured patch tokens with
additional degrees of freedom, including varying patch granularity, two-dimensional spatial dependencies, and, in some ViTs, a global \texttt{[CLS]} token that serves a fundamentally different computational role. These fundamental differences of the language and vision domains naturally render the practical utility of CLTs to the vision domain a distinct and nontrivial question that is yet to be explored. To that end, in this paper we investigate the practical utility and benefits of CLTs in ViTs by addressing the following open research questions:
\medskip

\noindent
\textbf{RQ1: Functional Replacement.} \textit{Can CLTs functionally replace MLP blocks in Vision Transformers as an alternative interpretable proxy}? 
We aim to evaluate whether CLTs can faithfully reconstruct post-MLP activations from sparse features of earlier layers, and whether such replacements preserve downstream classification performance under different sparsity schemes and replacement strategies within CLTs.

\medskip
\noindent
\textbf{RQ2: Interpretable Cross-Layer Contribution.} \textit{Can CLTs provide faithful attribution as well as process-level interpretability?} 
We aim to investigate whether the sparse, depth-ordered structure of CLTs supports meaningful interpretation of internal ViT representations by quantifying layer-wise influence using cross-layer contribution scores. Specifically, our goal is to provide both quantitative and example-based evidence of how different layers shape the final representation, yielding interpretable cross-layer contributions.

\medskip

\noindent

Driven by these open research questions, the contributions of the present work can be summarized as follows:
\begin{itemize}
    \item \textbf{CLTs for Vision.} We investigate the effectiveness of Cross-Layer Transcoders in Vision Transformers, enabling sparse, cross-layer reconstruction of post-MLP activations from earlier layers.

    \item \textbf{Accurate MLP Replacement in key regimes.} CLTs can replace MLP blocks, especially in later layers or for [CLS] tokens across all layers, preserving and even improving in some cases zero-shot classification performance. Consequently, CLTs serve as a reliable proxy for the original ViT to investigate how earlier layers contribute across depth of the model. 
    \item \textbf{Interpretability via Cross-Layer Attribution.} Projection-based
contribution scores show that patch tokens exhibit
strongly diagonal-dominant attribution, with each layer
primarily explaining its own post-MLP output, while the
\texttt{[CLS]} token draws credit broadly across depth,
aggregating information from many preceding layers.

\item{ \textbf{Faithful attribution reveals concentrated
credit across depth.} Ablation experiments
demonstrate that the final-layer representation is
predominantly shaped by a small subset of source layers:
retaining only the top-4 out of 12 layers recovers
original model's accuracy, while removing the single
highest-scored layer causes substantial degradation.
This corroborates that the projection-based scores
faithfully identify the layers that are most significant in forming the output representation.}
\end{itemize}

\section{Related Work}
\label{sec:related}
\paragraph{\textbf{Mechanistic Interpretability and Sparse Autoencoders.}}

A wide range of methods have been proposed to interpret vision models, including feature visualization \citep{simonyan2014visualising, zeiler2014visualizing, olah2017feature} and network dissection \citep{bau2017network, oikarinen2022clip}. 
More recently, Mechanistic Interpretability has emerged as a systematic approach to analyzing and understanding neural networks~\citep{elhage2021mathematical, olah2020zoom}. A central challenge in this area is the presence of polysemantic neurons, i.e., units that respond to multiple, seemingly unrelated inputs, arising from superposition. Superposition is the phenomenon where networks encode more features than the available dimensions allow, forcing different concepts to share the same activations~\citep{elhage2022toy}. Recent efforts have leveraged SAEs to uncover interpretable features within LLMs ~\citep{templeton2024scaling, cunningham2023sparse, gao2024scaling}. Sparse Autoencoders (SAEs) have been employed to alleviate superposition by learning sparse, overcomplete representations of internal model activations via dictionary learning~\citep{sharkey2022taking, bricken2023monosemanticity}.

Although SAEs have been predominantly studied in the context of language models, recent efforts explore SAEs' utility in the vision domain.
In particular, SAEs have been applied to feature analysis, generative modeling, and concept disentanglement in the visual domain~\citep{ bhalla2024interpreting, stevens2025sparse, surkov2024one, fel2025archetypal, thasarathan2025universal}
 . 
Much of this work has centered on interpretability, revealing latent structure in deep visual representations. More recent studies have extended these insights to practical downstream applications, including the use of sparse features to analyze CLIP’s susceptibility to typographic attacks~\cite{joseph2025steering}, improve classification accuracy via class-conditioned latent masking and feature selection~\cite{lim2024sparse}, and guide model predictions through visual sparse steering~\cite{chatzoudis2025visual}.

\paragraph{\textbf{Feature Circuits and Cross-Layer Transcoders}}

Sparse Autoencoders have also been applied to uncover feature circuits, i.e., directed graphs that capture how interpretable features activate, interact, and causally contribute to a model’s final output~\cite{marks2024sparse, cunningham2023sparse, hanna2025circuit, conmy2023towards}.
In parallel to Sparse Autoencoders, recent work has investigated more structured sparsity-based models for interpretability. Transcoders, for example, are sparse, feedforward modules that operate within a single layer, reconstructing post-MLP activations from pre-MLP inputs using learned dictionaries~\cite{dunefsky2024transcoders}. 

While Transcoders focus on modeling transformations within a layer, other approaches extend this idea across layers to capture cross-layer causal structure. 
Specifically, Crosscoders generalize this idea by learning sparse codes that jointly reconstruct representations across multiple layers, or even across different models, enabling applications such as feature sharing and model diffing~\cite{lindsey2024crosscoders, minder2025robustly, minder2025overcoming}. 
Cross-Layer Transcoders (CLTs) represent a specialized instance of Crosscoders, forming a triangular, depth-ordered architecture that reconstructs each MLP output from the sparse codes of all preceding layers~\citep{ameisen2025circuittracing}. CLTs have been applied in LLMs to construct attribution graphs and replacement models that reveal how early-layer features influence downstream computations ~\citep{ameisen2025circuittracing}.
Our work explores the effectiveness of CLTs in the context of Vision Transformers (ViTs), presenting a CLT-based model for ViT activations and investigating both functional replacement and visual interpretability.

\section{Method}
\label{sec:method}

\subsection{Cross-Layer Transcoder (CLT)}

\paragraph{Transformer Preliminaries.}
Let a Vision Transformer (ViT) with $L$ layers operate on $T$ tokens of width $d$. In layer $i$, let $x_i \in \mathbb{R}^{T \times d}$ denote the post-attention (LN2), pre-MLP activations and let the MLP produce $y_i = \text{MLP}_i(x_i) \in \mathbb{R}^{T \times d}$. We target these $y_i$ for linear reconstruction using sparse features extracted from \emph{earlier} layers, enabling both attribution across depth and drop-in replacement of MLPs.

\paragraph{Sparse Feature Encoding (per Layer).} 
Each layer $i$ has a learned linear encoder with the weights $E_i \in \mathbb{R}^{d \times m}$ (typically $m > d$, overcomplete). 
We denote the embedding of token $t \in \{1,\dots,T\}$ as $x_{i,t} \in \mathbb{R}^d$. For each token, we map $x_{i,t}$ to a sparse feature vector
\begin{equation}
  z_{i,t} = \phi(x_{i,t} E_i) \in \mathbb{R}^m
  \label{eq:token-encoding}
\end{equation}
where $\phi(\cdot)$ is a non-linear function operating on the sparse feature space. We support three non-linear functions $\phi(\cdot)$:
\begin{itemize}
  \item \textbf{JumpReLU (learned thresholds)} \cite{rajamanoharan2024jumping}: for each token,
  \begin{equation}
  z_{i,t} = (x_{i,t} E_i) \odot \mathbf{1}\big[x_{i,t} E_i > \tau_i\big]
  \label{eq:jump-relu}
\end{equation}
  where $\tau_i \in \mathbb{R}^m_{\ge 0}$ are learned per-feature thresholds shared across tokens. The thresholds $\tau_i$ are learned jointly with the encoders and decoders using a straight-through estimator.

  \item \textbf{ReLU-Top-$k$} \cite{gao2024scaling}: for each token,
  \begin{equation}
  z_{i,t} = \mathrm{TopK}\big(\max(0, x_{i,t} E_i),\ k\big)
  \label{eq:relu-topk}
\end{equation}
  which keeps the top-$k$ positive features (by value) per token and sets all others to zero.
  \item \textbf{Abs-Top-$k$}, similar to \cite{zhu2025abstopk}: for each token, let
  $u_{i,t} = x_{i,t} E_i$, and define
  \begin{equation}
  z_{i,t} = \operatorname{sign}(u_{i,t}) \odot \mathrm{TopK}\big(|u_{i,t}|,\ k\big)
  \label{eq:abs-topk}
\end{equation}
  which keeps the top-$k$ features by absolute value per token while preserving their sign.
\end{itemize}

\paragraph{Cross-Layer Transport.}
For each destination layer $j$, we reconstruct the MLP output from all sources $i \le j$:
\begin{equation}
  \hat{y}_j = \sum_{i=1}^{j} z_i W^{i \rightarrow j}_{\text{dec}}, \qquad W^{i \rightarrow j}_{\text{dec}} \in \mathbb{R}^{m \times d}
  \label{eq:clt-reconstruction}
\end{equation}

Here $z_i \in \mathbb{R}^{T \times m}$ denotes the matrix of token-wise sparse
codes at layer $i$, whose $t$-th row is the per-token code $z_{i,t}^\top$.
This additive, token-wise reconstruction enforces a strictly causal structure: each target layer $j$ is reconstructed only from activations in source layers $i \le j$, thereby preventing information leakage from future layers. The resulting triangular decoder architecture naturally defines a directed attribution graph across depth, where each edge $i \rightarrow j$ corresponds to a learned contribution from layer $i$ toward reconstructing $y_j$.

\paragraph{Training Objective.}
Given supervision pairs $\{(x_j, y_j)\}_{j=1}^L$, we minimize the total reconstruction loss:
\begin{equation}
  \mathcal{L}
  = \sum_{\ell=1}^{L} \left\| \hat{y}_\ell - y_\ell \right\|_2^2
    + \lambda \sum_{\ell=1}^{L} \mathcal{R}_{\text{sparse}}(z_\ell)
  \label{eq:clt-loss}
\end{equation}

where $\hat{y}_\ell$ is the CLT reconstruction at layer $\ell$ and $z_\ell$ are the feature activations at that layer. 
The sparsity term $\mathcal{R}_{\text{sparse}}(z_\ell)$ depends on the choice of the non-linear function $\phi(\cdot)$:
\begin{itemize}
  \item \textbf{JumpReLU}: Following the CLT setup in \cite{ameisen2025circuittracing}, we use a decoder-norm weighted Tanh sparsity penalty applied per feature. Let $z_\ell \in \mathbb{R}^{B \times T \times m_\ell}$
denote the activations at layer $\ell$, where $B$ is the batch size, $T$ the number of tokens per example, and $m_\ell$ the number of features at that layer. For each feature $j$, we define $W_{\mathrm{dec}, j}^{(\ell)}$ as the concatenation of all decoder vectors that read from feature $j$ at layer $\ell$.
Our sparsity term is then:
\begin{equation}
\small
\mathcal{R}_{\mathrm{sparse}}(z_\ell)
=
\mathbb{E}_{b,t,j}\left[
  \tanh\left(
    c \, \big\|W_{\mathrm{dec}, j}^{(\ell)}\big\|_2 \cdot
    \big|z_{\ell, b, t, j}\big|
  \right)
\right]
\label{eq:sparsity-penalty}
\end{equation}
where $c > 0$ is a hyperparameter controlling the sharpness of the penalty and $\mathbb{E}_{b,t,j}[\cdot]$ denotes the empirical average over all batch, token, and feature indices. This encourages features with large decoder norm
$\|W_{\mathrm{dec}, j}^{(\ell)}\|_2$ to be used more sparingly, while the $\tanh(\cdot)$ keeps the penalty bounded and provides smooth gradients.
  \item \textbf{ReLU Top-$k$} and \textbf{Abs Top-$k$}: No additional regularization is used, as sparsity is enforced directly by the Top-K operator.
\end{itemize}

\subsection{Cross-Layer Attribution}

The CLT yields a faithful decomposition of each target layer's representation as a linear sum of contributions from all preceding layers:
\begin{equation}
\hat{y}_j = \sum_{i \le j} c_{i \rightarrow j}, \qquad
c_{i \rightarrow j} = z_i \, W^{i \rightarrow j}_{\text{dec}},
\label{eq:clt-contribs}
\end{equation}
where $c_{i \rightarrow j} \in \mathbb{R}^{T \times D}$ is the decoded contribution from source layer $i$ toward reconstructing the MLP output at target layer $j$, with $T$ denoting the number of tokens and $D$ the hidden dimension.

A natural question is which source layers are most responsible for shaping the reconstruction at each target layer. To quantify this, we project each layer's contribution onto the full reconstruction, yielding a per-token, per-layer attribution score that measures the fraction of the output representation explained by each source layer:
\begin{equation}
C^{\mathrm{proj}}_{i \rightarrow j}
= \mathbb{E}_{x} \left[
\frac{1}{T} \sum_{t=1}^{T}
\frac{
\langle c_{i \rightarrow j}^{(t)}(x),\; \hat{y}_j^{(t)}(x) \rangle
}{
\| \hat{y}_j^{(t)}(x) \|_2^2
}
\right],
\label{eq:projection-contrib}
\end{equation}
where $c_{i \rightarrow j}^{(t)}, \hat{y}_j^{(t)} \in \mathbb{R}^{D}$ denote the contribution and reconstruction vectors at token $t$, and $\langle \cdot,\cdot\rangle$ is the inner product over the feature dimension. By construction, the per-source scores decompose each layer's output into signed additive contributions from all preceding layers, providing an attribution over depth. Scores close to zero indicate source layers with negligible influence on the target, while negative scores identify layers whose contributions partially oppose the aggregate reconstruction, a form of inter-layer redundancy that the CLT makes explicitly visible. This decomposition enables us to inspect, at any token and target layer, how the representation is assembled across the network's depth.

\section{Replacing MLP blocks with Cross-Layer Transcoders}
\label{sec:mlp-replacement}

Cross-layer Transcoders aim to reconstruct the representations of a target layer from the pre-MLP representations of previous layers. Reconstructing the target layer representations from earlier ones allows us to construct a \emph{Replacement Model}, where the MLP block outputs are replaced by their corresponding CLT approximations. Cross-Layer Transcoders can then be used as a \emph{proxy} to investigate the original model's internal representations. In this section, we first aim to train Cross-Layer Transcoders for Vision Transformers and explore whether we can faithfully reconstruct late-layer post-MLP representations. We then examine how this replacement affects the model's performance on downstream classification tasks.

\begin{figure*}[t!]
    \centering
    \includegraphics[width=0.32\textwidth]{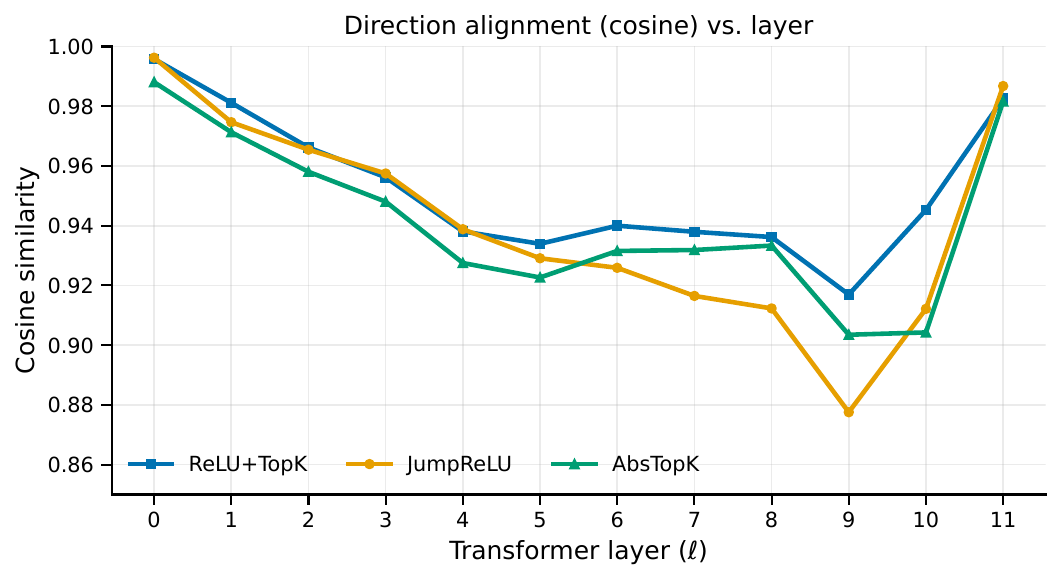}
    \includegraphics[width=0.32\textwidth]{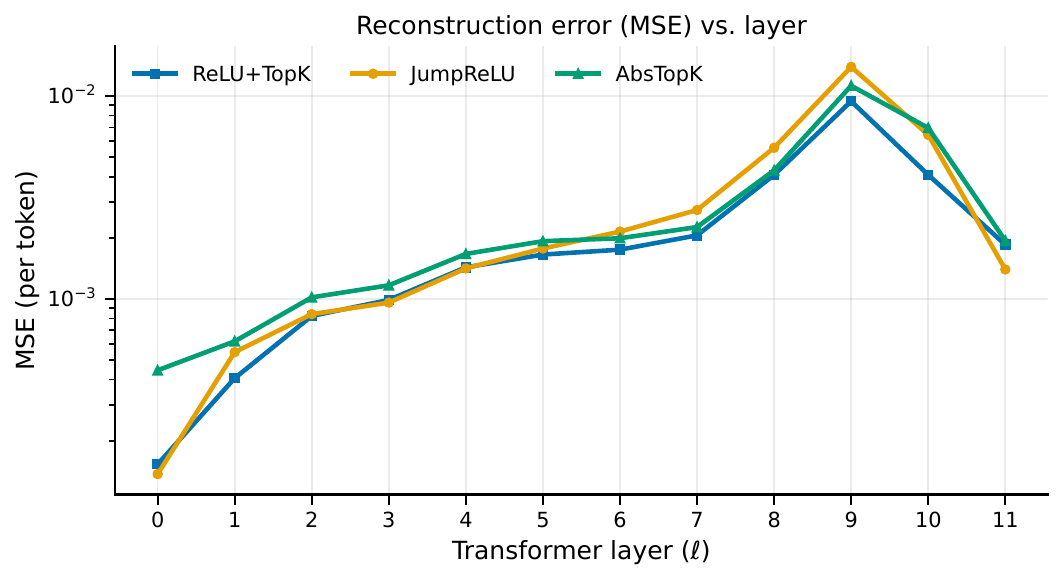}
    \includegraphics[width=0.32\textwidth]{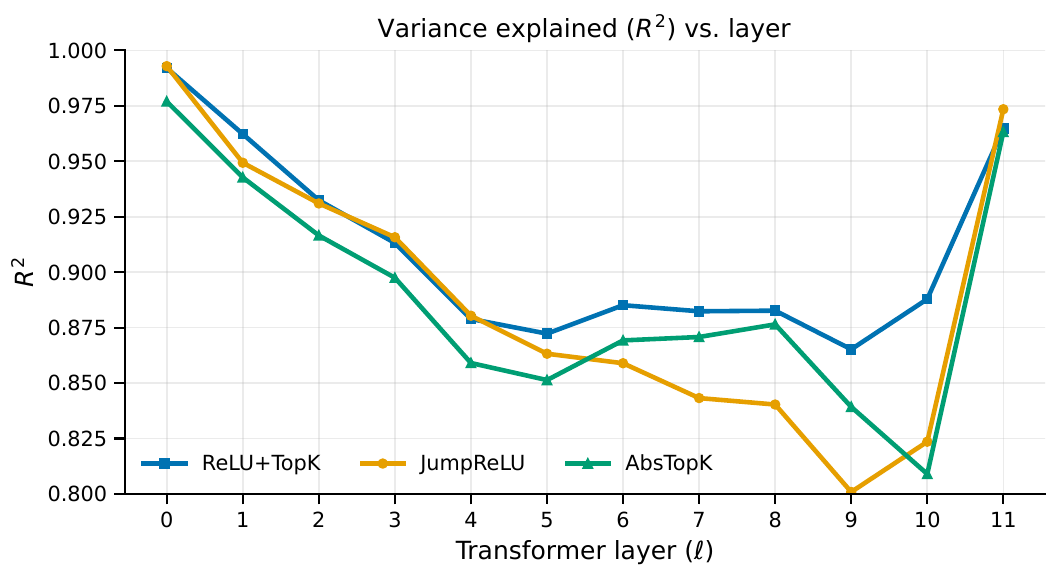}
    \caption{
    Reconstruction performance of Cross-Layer Transcoders (CLTs) across transformer layers on \textsc{CIFAR-100} using ViT-B/32. We compare three sparsity variants, i.e., \textsc{JumpReLU}, \textsc{ReLU-Top-}$k$, and \textsc{Abs-Top-}$k$ (with $k{=}128$) using  (left) cosine similarity, (center) mean squared error (MSE per token, log scale), and (right) variance explained ($R^2$). Metrics are averaged over all tokens in the test set.
    }
    \label{fig:clt_reconstruction_metrics}
\end{figure*}

\subsection{Reconstruction of MLP Representations}
\label{sub:clt-reconstruction}

We train Cross-Layer Transcoders (CLTs) on \textsc{CIFAR-100}, \textsc{COCO}, and \textsc{ImageNet-100} using \textsc{CLIP} ViT-B/32 and ViT-B/16 backbones. For each model--dataset pair, we instantiate three sparsity variants: \textsc{JumpReLU}, \textsc{ReLU-Top-}$k$, and \textsc{Abs-Top-}$k$, with $k{=}128$. CLTs are trained on \emph{all tokens} for $10$ epochs with an expansion factor of $16$, using AdamW with learning rate $2{\times}10^{-4}$. Additional training details are provided in the Supplementary material.

To evaluate how faithfully CLTs approximate post-MLP representations across depth, we compare, for each layer $\ell$, the reconstructed output $\hat{y}_\ell$ to the teacher activation $y_\ell$ at every token and compute Mean Squared Error (MSE), $R^2$, and cosine similarity. For each model–dataset pair, these quantities are first averaged over tokens and layers. Table~\ref{tab:clt-reconstruction} summarizes the reconstruction quality for \textsc{CLIP} ViT-B/32 and ViT-B/16 on \textsc{CIFAR-100}, \textsc{COCO}, and \textsc{ImageNet-100}, comparing the three sparsity variants: \textsc{JumpReLU}, \textsc{ReLU-Top-$k$}, and \textsc{Abs-Top-$k$}. 

Across all settings, CLTs achieve strong reconstruction fidelity with the teacher representations: layer-averaged cosine similarities lie in the $0.92$–$0.97$ range and $R^2$ values are typically between $0.89$ and $0.95$, indicating that most of the variance in post-MLP activations is captured by CLTs. ViT-B/16 is consistently easier to approximate than ViT-B/32 (e.g., on \textsc{CIFAR-100}, the per-layer $R^2$ improves from roughly $0.89$–$0.91$ to $0.92$–$0.94$), and on B/32 backbones \textsc{ReLU-Top-$k$} enjoys the lowest MSE and highest cosine.

\noindent \textbf{Remark}. We hypothesize that the reconstruction gap
between the two backbones stems from patch granularity:
ViT-B/16 uses smaller patches, distributing visual
information across a larger token set and yielding simpler
per-token activations, whereas ViT-B/32 compresses the
same image into fewer, larger patches, producing richer
per-token representations that are inherently harder to
approximate.

To further examine how reconstruction quality varies with depth, Figure~\ref{fig:clt_reconstruction_metrics} plots per-layer MSE, $R^2$, and cosine similarity for ViT-B/32 on \textsc{CIFAR-100}. We observe a clear depth-dependent pattern: very early layers and the final layer are reconstructed almost perfectly (e.g., for \textsc{ReLU-Top-$k$}, $R^2\geq 0.96$ and cosine $\geq 0.98$ at layers 0 and 11), while intermediate blocks incur higher error (with $R^2$ dipping to ${\approx}0.86$ and cosine to ${\approx}0.92$ around layers 7–9). Among sparsifiers, \textsc{ReLU-Top-$k$} is consistently strongest in the mid–late regime, whereas \textsc{JumpReLU} and \textsc{Abs-Top-$k$} exhibit sharper degradation in the deepest middle layers (e.g., $R^2$ near $0.80$–$0.84$ and slightly lower cosine).

\begin{table}[t]
\centering
\caption{Reconstruction performance of CLTs, averaged over all post-MLP layers across datasets and backbones. We report Mean Squared Error (MSE $\downarrow$), $R^2$ ($\uparrow$), and Cosine similarity ($\uparrow$) for different sparsity variants.}
\label{tab:clt-reconstruction}
\scriptsize
\begin{adjustbox}{max width=\linewidth}
\begin{tabular}{lllccc}
\toprule
\textbf{Dataset} & \textbf{Backbone} & \textbf{Sparsity} & \textbf{MSE} $\downarrow$ & \textbf{$R^2$} $\uparrow$ & \textbf{Cosine} $\uparrow$ \\
\midrule
\multirow{6}{*}{\textsc{CIFAR-100}} 
  & \multirow{3}{*}{ViT-B/32} 
    & \textsc{JumpReLU}       & 0.0032& 0.8894& 0.9411\\
  &                         & \textsc{ReLU-Top-}$k$   & \textbf{0.0024}& \textbf{0.9099}& \textbf{0.9525}\\
  &                         & \textsc{Abs-Top-}$k$    & 0.0030& 0.8893& 0.9418\\
  & \multirow{3}{*}{ViT-B/16} 
    & \textsc{JumpReLU}       & 0.0018& 0.9366& 0.9571\\
  &                         & \textsc{ReLU-Top-}$k$   & \textbf{0.0017}& \textbf{0.9369}& \textbf{0.9564}\\
  &                         & \textsc{Abs-Top-}$k$    & 0.0022& 0.9202& 0.9464\\
\midrule
\multirow{6}{*}{\textsc{COCO}} 
  & \multirow{3}{*}{ViT-B/32} 
    & \textsc{JumpReLU}       & 0.0026& 0.8971& 0.9456\\
  &                         & \textsc{ReLU-Top-}$k$   & \textbf{0.0018}& \textbf{0.9119}& \textbf{0.9542}\\
  &                         & \textsc{Abs-Top-}$k$    & 0.0028& 0.8934& 0.9442\\
  & \multirow{3}{*}{ViT-B/16} 
    & \textsc{JumpReLU}       & \textbf{0.0011}& \textbf{0.9541}& \textbf{0.9715}\\
  &                         & \textsc{ReLU-Top-}$k$   & 0.0012& 0.9375& 0.9609\\
  &                         & \textsc{Abs-Top-}$k$    & 0.0026& 0.8945& 0.9336\\
\midrule
\multirow{6}{*}{\textsc{ImageNet-100}} 
  & \multirow{3}{*}{ViT-B/32} 
    & \textsc{JumpReLU}       & 0.0026& 0.8976& 0.9457\\
  &                         & \textsc{ReLU-Top-}$k$   & \textbf{0.0022}& \textbf{0.8995}& \textbf{0.9473}\\
  &                         & \textsc{Abs-Top-}$k$    & 0.0036& 0.8504& 0.9209\\
  & \multirow{3}{*}{ViT-B/16} 
    & \textsc{JumpReLU}       & \textbf{0.0011}& \textbf{0.9505}& \textbf{0.9691}\\
  &                         & \textsc{ReLU-Top-}$k$   & 0.0012& 0.9450& 0.9654\\
  &                         & \textsc{Abs-Top-}$k$    & 0.0024& 0.9081& 0.9421\\
\bottomrule
\end{tabular}
\end{adjustbox}
\end{table}

\subsection{Cascaded Replacement Models}
\label{sub:clt-substitution}

In the previous section, we showed that CLTs can faithfully approximate post-MLP activations across transformer depth. We now ask a functional question: \emph{can CLTs replace MLPs without degrading the ViT’s zero-shot classification performance?} In the baseline ViT, block $\ell$ computes
$y_\ell = \mathrm{MLP}_\ell(x_\ell)$ and updates the
representation via the residual connection
$x_{\ell+1} = x_\ell + y_\ell$. Within a contiguous
replacement range $\ell_1{\to}\ell_2$, we substitute
MLP outputs with CLT reconstructions:
\begin{equation}
y_\ell' =
\begin{cases}
\hat{y}_\ell, & \ell \in [\ell_1, \ell_2] \\[2pt]
y_\ell, & \text{otherwise}
\end{cases},
\qquad
x_{\ell+1}' = x_\ell' + y_\ell'
\label{eq:clt-residual-switch}
\end{equation}
with $x_0' = x_0$ and
\begin{equation}
\hat{y}_\ell = \sum_{i \le \ell} z_i' \, W^{i \to \ell}_{\mathrm{dec}},
\qquad
z_i' = \phi(x_i' \, E_i)
\label{eq:clt-cascade}
\end{equation}
where each modified output $y_\ell'$ propagates forward
through the residual connection, allowing reconstruction
errors to compound across depth. We refer to this procedure as \emph{Cascaded Replacement}, and use the
$\ell_1{\to}\ell_2$ notation notation throughout to denote the contiguous block of substituted layers. In particular, we progressively widen the replacement window toward the final block, evaluating how faithfully CLTs preserve zero-shot accuracy as substitution depth grows.

\begin{table}[t]
  \centering
  \caption{
    Top-1 zero-shot accuracy (\%) for CLIP ViT-B/32 with ReLU-Top-$k$ sparsity ($k{=}128$)
    under cascaded MLP replacement.
  }
  \scriptsize
  \setlength{\tabcolsep}{2.8pt}
  \begin{adjustbox}{width=\columnwidth}
  \begin{tabular}{l l l c cccc}
    \toprule
    Dataset & Model & Tokens & Baseline & 3$\to$11 & 7$\to$11 & 10$\to$11 & 0$\to$11 \\
    \midrule
    \multirow{5}{*}{CIFAR-100}
      & \multicolumn{2}{l}{Baseline} & 61.65 &  &  &  &  \\
      \cmidrule(lr){2-8}
      & Transcoders                    & \multirow{2}{*}{CLS} &  & 61.41& 61.37& 61.29& 61.40\\
      & CLTs &                       &  & 61.69 & 61.86& 61.39& 61.74\\
      \cmidrule(lr){2-8}
      & Transcoders                    & \multirow{2}{*}{All} &  & 56.16& 61.10& 61.48& 53.89\\
      & CLTs &                       &  & 54.84& 62.08& 61.20& 51.12\\
    \midrule
    \multirow{5}{*}{COCO}
      & \multicolumn{2}{l}{Baseline} & 43.12 &  &  &  &  \\
      \cmidrule(lr){2-8}
      & Transcoders                    & \multirow{2}{*}{CLS} &  & 43.10& 42.98& 43.00& 43.22\\
      & CLTs &                       &  & 43.22 & 43.24& 43.10 & 43.36 \\
      \cmidrule(lr){2-8}
      & Transcoders                    & \multirow{2}{*}{All} &  & 41.14& 42.96& 43.00& 39.54\\
      & CLTs &                       &  & 41.34& 43.10& 43.10 & 39.12\\
    \midrule
    \multirow{5}{*}{ImageNet-100}
      & \multicolumn{2}{l}{Baseline} & 80.42 &  &  &  &  \\
      \cmidrule(lr){2-8}
      & Transcoders                    & \multirow{2}{*}{CLS} &  & 80.38& 80.48& 80.84& 80.46\\
      & CLTs &                       &  & 80.78 & 80.72 & 80.72 & 80.86\\
      \cmidrule(lr){2-8}
      & Transcoders                    & \multirow{2}{*}{All} &  & 76.54& 80.02& 80.74& 70.10\\
      & CLTs &                       &  & 75.36& 80.44& 80.58& 68.90\\
    \bottomrule
  \end{tabular}
  \end{adjustbox}
  \label{tab:b32-relu-topk-transcoder-vs-clt}
\end{table}

\subsubsection{Cross-Layer vs.\ Per-Layer Transcoders}
\label{subsec:clt-vs-transcoder}
We first compare Cross-Layer Transcoders to standard per-layer Transcoders trained with the same ReLU-Top-$k$ sparsity on ViT-B/32. Standard Transcoders are a degenerate case of Cross-Layer Transcoders where information flow is restricted to intralayer activations only.
In Table~\ref{tab:b32-relu-topk-transcoder-vs-clt}, we observe that Cross-Layer Transcoders match or slightly outperform Transcoders in the [CLS]-only setting, while exhibiting comparable
behaviour for All-token routing. Both architectures remain close to the baseline
for late-layer substitution (7$\to$11, 10$\to$11), but degrade when cascading
across all layers and all tokens, highlighting the difficulty of fully replacing
early-layer patch computations. Overall, Transcoders and Cross-Layer Transcoders exhibit very similar zero-shot behavior under MLP replacement. However, only Cross-Layer Transcoders expose
explicit cross-layer contributions via their triangular decoder
($c_{i \rightarrow j} = z_i W^{i \rightarrow j}_{\text{dec}}$), which is crucial
for our cross-layer attribution analysis in Section~\ref{sec:interpretability}. For this reason, in the remainder of the paper we focus on Cross-Layer Transcoders.

\subsubsection{Cross-Layer Transcoder Ablations}
\label{subsec:clt-ablations}
In Table~\ref{tab:clt-top1-compact}, we report top-1 zero-shot classification accuracy for CLIP ViT-B/32 and ViT-B/16 on \textsc{CIFAR-100}, \textsc{COCO}, and \textsc{ImageNet-100} under different cascaded replacement ranges ($3{\to}11$, $7{\to}11$, $10{\to}11$, and \textit{All}, where \textit{All} denotes replacing all MLP layers $0{\to}11$). In all cases, CLTs were trained offline against the frozen teacher model (teacher-forcing), but at test time, cascaded substitution uses the CLT outputs as inputs to subsequent layers, allowing us to directly measure how reconstruction errors accumulate along the depth of the network. We also compare the cascading effect of CLT substitution under different token-routing modes, i.e., \texttt{[CLS]} only, patch tokens only, and all tokens, which allows us to disentangle its impact on the global classification token versus spatial tokens. 

To assess functional fidelity, we compare the baseline ViT and its CLT-substituted counterpart on the same test splits used for training CLTs, using standard CLIP zero-shot classification. Following CLIP, we compute cosine similarity between the image embedding (taken from the final \texttt{[CLS]} representation) and a set of text embeddings for class-name prompts, and assign the label with highest similarity. Table~\ref{tab:clt-top1-compact} reports top-1 accuracy for each dataset, backbone, token-routing mode, and layer range, across the three sparsifiers: JR (JumpReLU), RTK (ReLU-Top-$k$), and ATK (Abs-Top-$k$).

\begin{table*}[t]
\centering
\caption{Top-1 classification accuracy (\%) across datasets, backbones, and token types. Columns correspond to layer range and sparsity combinations: JR = JumpReLU, RTK = ReLU-Top-$k$, ATK = Abs-Top-$k$. For reference, baseline top-1 accuracies (in \%) are: CIFAR-100 (B/32: 61.65, B/16: 65.97), COCO (B/32: 43.12, B/16: 43.56), and ImageNet-100 (B/32: 80.42, B/16: 84.34).}
\label{tab:clt-top1-compact}
\scriptsize
\resizebox{\textwidth}{!}{%
\begin{tabular}{lllcccccccccccc}
\toprule
\textbf{Dataset} & \textbf{Backbone} & \textbf{Tokens} & \multicolumn{3}{c}{3$\rightarrow$11} & \multicolumn{3}{c}{7$\rightarrow$11} & \multicolumn{3}{c}{10$\rightarrow$11} & \multicolumn{3}{c}{All} \\
\cmidrule(lr){4-6} \cmidrule(lr){7-9} \cmidrule(lr){10-12} \cmidrule(lr){13-15}
& & & JR & RTK & ATK & JR & RTK & ATK & JR & RTK & ATK & JR & RTK & ATK \\
\midrule

\textsc{CIFAR-100} & \multicolumn{14}{l}{} \\
& ViT-B/32 & CLS  &  61.38 & 61.69 & 61.58 & 61.41 & 61.86 & 61.26 & 61.06 & 61.39 & 61.16 & 61.33 & 61.74 & 61.43 \\
&  & Patches  &  53.65 & 54.20 & 55.83 & 61.51 & 61.69 & 63.18 & 61.49 & 61.25 & 61.56 & 49.63 & 51.12 & 48.68 \\
&  & All  &  54.10 & 54.84 & 55.69 & 61.54 & 62.08 & 62.85 & 61.01 & 61.20 & 61.11 & 49.40 & 51.12 & 48.41 \\
\cmidrule(lr){4-15}
& ViT-B/16 & CLS  &  66.10 & 65.81 & 65.39 & 66.15 & 66.15 & 65.86 & 66.06 & 65.90 & 65.90 & 66.04 & 65.92 & 65.38 \\
&  & Patches  &  63.54 & 60.76 & 63.31 & 66.05 & 65.48 & 67.24 & 65.97 & 65.82 & 66.13 & 62.40 & 58.57 & 56.59 \\
&  & All  &  63.71 & 60.75 & 62.87 & 66.22 & 65.56 & 67.08 & 65.91 & 65.98 & 65.89 & 62.45 & 58.82 & 56.72 \\

\textsc{COCO} & \multicolumn{14}{l}{} \\
& ViT-B/32 & CLS  &  43.08 & 43.22 & 43.12 & 43.26 & 43.24 & 43.20 & 43.04 & 43.10 & 42.90 & 43.12 & 43.36 & 43.00 \\
&  & Patches  &  40.58 & 40.92 & 41.48 & 42.98 & 43.14 & 42.80 & 43.14 & 43.18 & 43.06 & 39.04 & 38.94 & 39.68 \\
&  & All  &  41.00 & 41.34 & 41.46 & 43.14 & 43.10 & 42.76 & 42.78 & 43.10 & 42.92 & 38.60 & 39.12 & 40.06 \\
\cmidrule(lr){4-15}
& ViT-B/16 & CLS  &  43.68 & 43.64 & 42.64 & 43.52 & 43.64 & 43.34 & 43.52 & 43.88 & 43.38 & 43.62 & 43.62 & 42.76 \\
&  & Patches  &  43.00 & 42.34 & 39.46 & 43.26 & 43.36 & 43.30 & 43.62 & 43.64 & 43.72 & 42.72 & 42.00 & 35.46 \\
&  & All  &  42.78 & 42.66 & 38.18 & 43.62 & 43.58 & 43.48 & 43.42 & 43.92 & 43.60 & 43.00 & 42.08 & 34.16 \\

\textsc{ImageNet-100} & \multicolumn{14}{l}{} \\
& ViT-B/32 & CLS  &  80.86 & 80.78 & 80.16 & 80.64 & 80.72 & 80.16 & 80.56 & 80.72 & 80.18 & 80.92 & 80.86 & 80.26 \\
&  & Patches  &  75.58 & 75.26 & 72.46 & 80.34 & 80.18 & 80.64 & 80.52 & 80.46 & 80.46 & 71.96 & 68.74 & 60.26 \\
&  & All  &  75.12 & 75.36 & 71.44 & 80.18 & 80.44 & 79.38 & 80.38 & 80.58 & 80.18 & 71.60 & 68.90 & 60.26 \\
\cmidrule(lr){4-15}
& ViT-B/16 & CLS  &  84.56 & 84.08 & 83.74 & 84.66 & 84.16 & 84.40 & 84.56 & 84.38 & 84.34 & 84.54 & 84.02 & 83.28 \\
&  & Patches  &  83.16 & 82.76 & 79.36 & 83.94 & 84.24 & 83.84 & 84.20 & 84.22 & 84.58 & 83.04 & 81.40 & 73.06 \\
&  & All  &  83.00 & 82.64 & 78.08 & 84.28 & 84.32 & 83.34 & 84.40 & 84.36 & 84.74 & 83.12 & 80.94 & 70.78 \\

\bottomrule
\end{tabular}
}
\end{table*}

\paragraph{\textbf{Results.}}
Across all datasets and backbones, CLT substitution is remarkably stable for \texttt{[CLS]}-only replacement: even when all layers are replaced (column \textit{All}), top-1 accuracies remain essentially at baseline (e.g., \textsc{CIFAR-100} ViT-B/32 CLS: 61.33–61.74 vs.\ 61.65;
\textsc{ImageNet-100} ViT-B/16 CLS: 83.28–84.54 vs.\ 84.34). 
Replacing only the last few layers (10$\rightarrow$11) is also safe for patch and all tokens, with accuracies nearly indistinguishable from the original model. 
In contrast, fully cascading CLTs across all layers and all tokens leads to clear
accuracy degradation, particularly when routing patch tokens. For example,
on \textsc{CIFAR-100} ViT-B/32, patch-token RTK drops from 61.65 (baseline) to
51.12 when all MLPs are replaced (column \textit{All}); on \textsc{COCO} ViT-B/16,
patch-token ATK falls from 43.56 to 35.46; and on \textsc{ImageNet-100} ViT-B/16
with all tokens, ATK drops from 84.34 to 70.78.
 
Overall, ViT-B/16 proves more robust to substitution than ViT-B/32, mirroring the higher reconstruction quality in Table~\ref{tab:clt-reconstruction}. 
The routing breakdown (\texttt{[CLS]} vs.\ patches vs.\ all tokens) thus reveals a consistent pattern: CLTs provide reliable functional approximations for the global classification \texttt{[CLS]} token and for late-layer MLPs, while early-layer patch substitutions accumulate error more aggressively.

\section{Interpreting Vision Transformer Activations via Cross-Layer Sparse Features}
\label{sec:interpretability}

Having established that CLTs can faithfully replace MLP blocks (Section~\ref{sec:mlp-replacement}), we now exploit the structure of the CLT decomposition itself as an interpretability tool. Because each target-layer reconstruction is expressed as an additive sum of decoded contributions from all preceding layers (Eq.~\ref{eq:clt-contribs}), CLTs provide a form of structured attribution unavailable from standard per-layer methods, decomposing each layer's reconstruction into signed, per-source contribution scores that quantify which layers matter most for a given target representation. We first analyze the resulting cross-layer attribution structure (Section~\ref{sec:clt-contrib}) and then validate its faithfulness through projection-based ablation (Section~\ref{sec:faithful-attribution}).

\subsection{Cross-Layer Contribution Scores}
\label{sec:clt-contrib}

To understand how the final-layer representation is assembled across depth, we evaluate the projection-based attribution scores $C^{\mathrm{proj}}_{i \rightarrow j}$ (Eq.~\ref{eq:projection-contrib}) for every target layer $j$, computing the fraction of each
target's reconstruction explained by each source layer
$i \le j$. Scores are computed separately for \texttt{[CLS]} and patch tokens by restricting the token average in Eq.~\ref{eq:projection-contrib} to the corresponding positions, and are then averaged over the validation set. The resulting heatmaps directly reveal how attribution is partitioned across source layers.

\begin{figure}[t]
    \centering
    \begin{subfigure}[t]{0.495\linewidth}
        \centering
        \includegraphics[width=\linewidth]{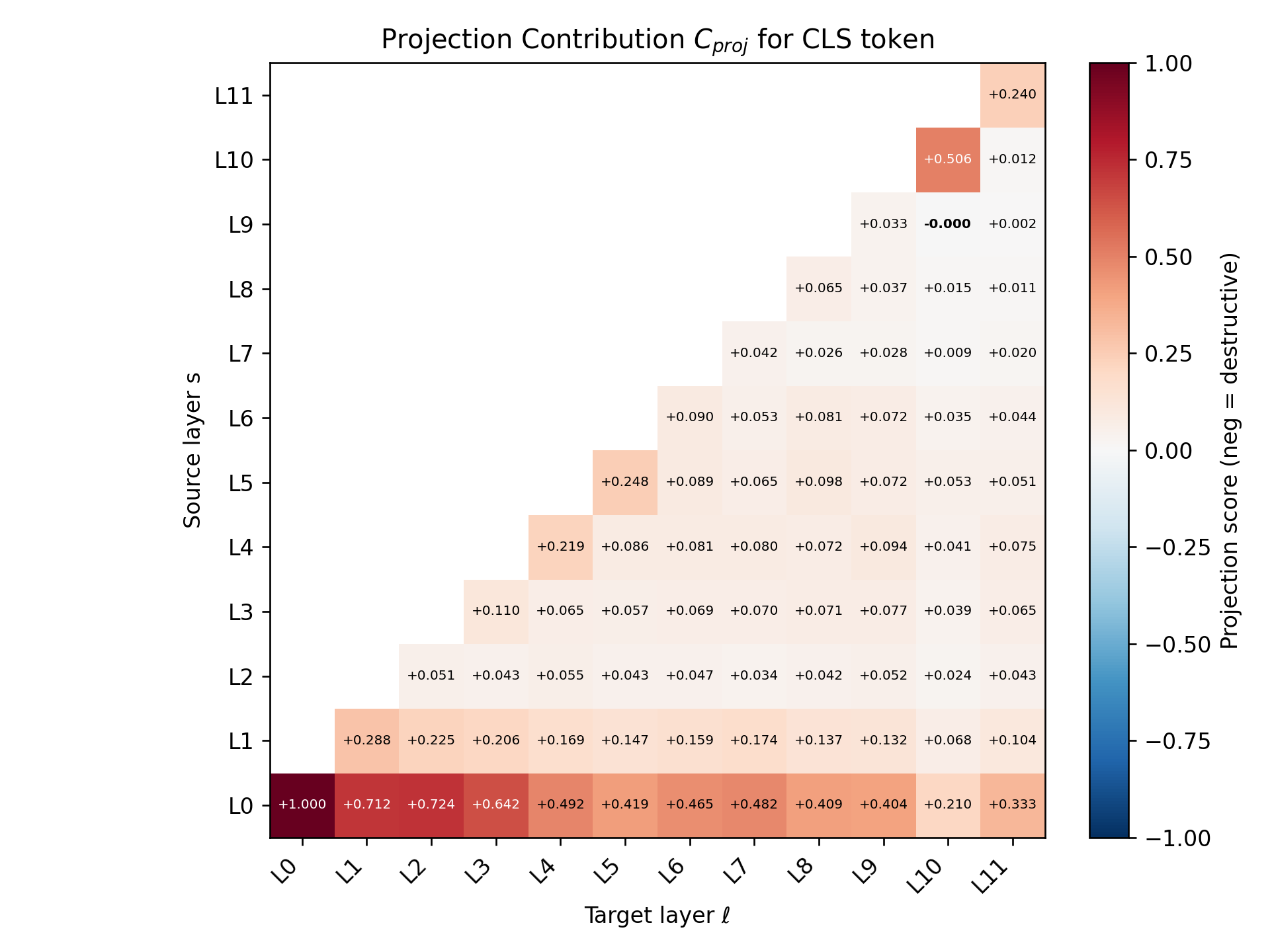}
        \caption{\texttt{[CLS]} token}
        \label{fig:contrib-cls}
    \end{subfigure}
    \hfill
    \begin{subfigure}[t]{0.495\linewidth}
        \centering
        \includegraphics[width=\linewidth]{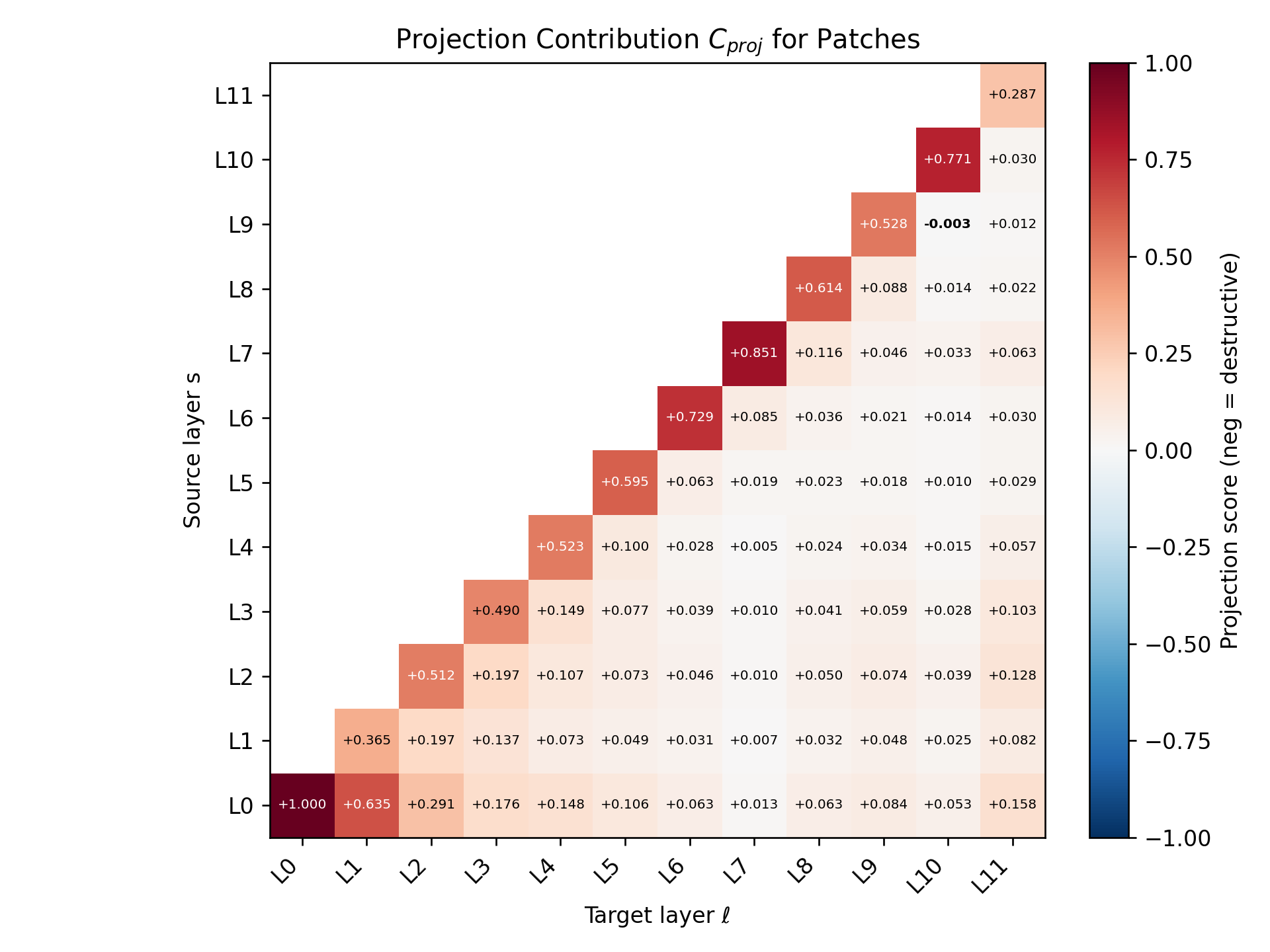}
        \caption{Patch tokens}
        \label{fig:contrib-patches}
    \end{subfigure}
    \caption{
    Cross-layer attribution scores $C^{\mathrm{proj}}_{i \rightarrow j}$
for ViT-B/16 on CIFAR-100, averaged over the validation set. Scores
along each column yield a partition of
attribution across depth. Patch tokens (right) exhibit strong diagonal
dominance, while \texttt{[CLS]} (left) draws attribution broadly across
source layers.
    }
    \label{fig:clt-contribution}
\end{figure}

Figure~\ref{fig:clt-contribution} visualizes $C^{\mathrm{proj}}_{i \rightarrow L}$ for \texttt{[CLS]} (left) and patch tokens (right), revealing two qualitatively distinct regimes. For patch tokens, the attribution matrix is strongly diagonal-dominant: each layer explains the majority of its own post-MLP output, with only modest credit diffusing to immediate neighbors. This locality is consistent with the view that patch-level computations are dominated by within-layer transformations that progressively refine spatial features. The \texttt{[CLS]} token exhibits a markedly different structure. Credit is distributed broadly across depth, with several early layers receiving scores comparable to or exceeding the self-layer term $C^{\mathrm{proj}}_{L \rightarrow L}$. This indicates that the final \texttt{[CLS]} representation does not emerge primarily from the last MLP block but is instead a depth-integrated aggregation of semantic signals accumulated over many preceding layers, consistent with its architectural role as a global summary token.

We also observe that even for patch tokens, the earliest
layer (L0) retains a non-negligible fraction of credit at
every target depth, which may indicate that low-level
features such as edges and textures persist in deeper
representations rather than being fully overwritten.
Additionally, the contrast between the diagonal patch structure and the distributed \texttt{[CLS]} structure offers an intuition for the asymmetric substitution behavior in Table~\ref{tab:clt-top1-compact}: \texttt{[CLS]}-only replacement may succeed because attribution is spread across many source layers, making the reconstruction less sensitive to errors at any single layer, whereas patch tokens depend predominantly on their own layer and are thus more affected by per-layer approximation errors.

\begin{table*}[t]
\centering

\caption{Faithful attribution via 
projection-based ablation on the final layer of ViT-B/32.
Accuracy (\%) and KL divergence from the unmodified logits
are reported for \texttt{[CLS]} and all tokens (\texttt{[CLS]}+patch).}

\resizebox{\textwidth}{!}{%
\begin{tabular}{llc cc cc cc cc}
\toprule
\multirow{2}{*}{\textbf{Model}} & \multirow{2}{*}{\textbf{Dataset}} & \multirow{2}{*}{\textbf{Tokens}} 
  & \multicolumn{2}{c}{\textbf{Baseline}} 
  & \multicolumn{2}{c}{\textbf{Full CLT}} 
  & \multicolumn{2}{c}{\textbf{Drop Top-1}} 
  & \multicolumn{2}{c}{\textbf{Keep Top-4}} \\
\cmidrule(lr){4-5} \cmidrule(lr){6-7} \cmidrule(lr){8-9} \cmidrule(lr){10-11}
& & & Acc$\uparrow$ & KL$\downarrow$ & Acc$\uparrow$ & KL$\downarrow$ & Acc$\uparrow$ & KL$\downarrow$ & Acc$\uparrow$ & KL$\downarrow$ \\
\midrule
ViT-B/32 & cifar100 & all (cls+patch) & 61.65\% & - & 61.31\% & 0.0104 & 59.68\% & 0.0800  & 59.96\% & 0.0714 \\
 &  & cls & 61.65\% & - & 61.31\% & 0.0104 & 59.91\% & 0.0704  & 60.72\% & 0.0582 \\
 & coco & all (cls+patch) & 43.12\% & - & 43.26\% & 0.0103 & 42.98\% & 0.1619  & 43.02\% & 0.0156 \\
 &  & cls & 43.12\% & - & 43.26\% & 0.0103 & 42.78\% & 0.1157  & 43.00\% & 0.0155 \\
 & imagenet100 & all (cls+patch) & 80.42\% & - & 80.54\% & 0.0073 & 74.94\% & 0.2320  & 80.56\% & 0.0109 \\
 &  & cls & 80.42\% & - & 80.54\% & 0.0073 & 74.86\% & 0.2733  & 80.48\% & 0.0106 \\
\bottomrule
\end{tabular}%
}
\label{tab:faithful}
\end{table*}

\subsection{Faithful Attribution}
\label{sec:faithful-attribution}

To validate that the cross-layer attribution scores faithfully reflect each source layer's functional importance, we perform projection-based ablation experiments on the final-layer reconstruction. Specifically, for each input we rank the source layers by their attribution score $C^{\mathrm{proj}}_{i \rightarrow j}$ at the final target layer $j{=}L$ and evaluate three ablation conditions: (i)~\textbf{Full CLT}, which retains all layer contributions and measures the baseline fidelity of the decomposition; (ii)~\textbf{Drop Top-1}, which removes the single highest-scored source layer; and (iii)~\textbf{Keep Top-4}, which retains only the four highest-scored layers and zeros out all remaining contributions. For each condition, we substitute the ablated reconstruction $\tilde{y}_L$ into the CLIP vision encoder in place of the original MLP output and report zero-shot classification accuracy and KL divergence from the unmodified logit distribution. All rankings are computed per instance, ensuring that the ablation respects input-dependent variation in layer importance. We evaluate on both the \texttt{[CLS]} token alone and all tokens.

Results are presented in Table~\ref{tab:faithful}. The Full CLT reconstruction closely preserves baseline accuracy across all datasets, with KL divergence below $0.011$ in every setting, confirming that the additive decomposition is near-lossless. Removing the single highest-ranked source layer (Drop Top-1) produces consistent accuracy drops, most notably on ImageNet-100, where accuracy falls from $80.54\%$ to $74.94\%$ ($-5.6$ \%) with a corresponding KL increase to $0.232$, demonstrating that the top-scoring layer captures a disproportionate share of the classification-relevant signal. Conversely, retaining only the top-4 source layers (Keep Top-4) recovers near-baseline performance, with accuracy within $0.2$ \% and KL divergence under $0.02$ on ImageNet-100. 
This dual evidence of \emph{necessity} (removal of high-scoring layers degrades performance) and \emph{sufficiency} (retention of a small subset preserves it) confirms that the projection-based scores provide a faithful assessment of each layer's contribution to the final representation. The consistency of these trends across datasets and token configurations further supports the generality of the attribution mechanism.

\section{Conclusion and Future Work}

In this paper, we introduced the novel adoption of CLTs as sparse, depth-aware, interpretable proxy models for MLP blocks in Vision Transformers. 
Across three datasets and two CLIP backbones, we have shown that CLTs achieve high reconstruction fidelity while preserving, and in some cases even improving, zero-shot classification accuracy, particularly when substituting the \texttt{[CLS]} token or considering late-layer blocks CLT-based approximations.
Beyond functional replacement, the innate linear decomposition structure of CLTs yields signed, per-source attribution scores that reveal a striking asymmetry: patch tokens are governed predominantly by within-layer transformations, whereas the \texttt{[CLS]} token aggregates semantic signal broadly across depth; an architectural insight that cannot be seen in standard per-layer methods such as conventional transcoders or SAEs. In addition, the novel adoption of CLTs in the vision domain provides a faithful attribution: the final representation is concentrated in a small subset of dominant source layers whose removal degrades performance and whose retention largely preserves it. Together, these results and insights position CLTs as a principled framework for structured, cross-layer interpretability in vision models without compromising zero-shot classification performance.

The depth-resolved sparse features learned by CLTs give rise to three distinct research directions that belong to our future research agenda. 
 First, the layer-wise decomposition provides a principled substrate for hierarchical concept discovery, revealing how visual primitives in early layers give rise to semantic abstractions in deeper ones and tracing their composition across depth through the CLT attribution structure. Second, CLTs currently model only the MLP pathway; incorporating attention heads into the cross-layer decomposition can potentially yield a unified sparse proxy that accounts for both feature transformation and token interaction, enabling end-to-end circuit analysis for Vision Transformers. Finally, we aim to explore CLTs as an interpretable proxy for a broader range of vision tasks and for diverse  Multimodal Large Language Models.

\newpage
\section*{Acknowledgments}
\label{Acknowledgments}

We would like to thank the reviewers and the area chair for their evaluation and feedback. This work was supported in part by NSF grants 2212302, 2212301, 2212303, AFOSR 23RT0630, and NIH 2R01HL127661. 

{
    \small
    \bibliographystyle{ieeenat_fullname}
    \bibliography{main}
}

\clearpage
\setcounter{page}{1}
\maketitlesupplementary
\newcommand{\contribsubfig}[4]{%
  \begin{subfigure}[t]{0.3\linewidth}
    \centering
    \includegraphics[width=\linewidth]{figures/supplementary/#1_#2_#3_#4.png}
    \caption{#3, #4}
    \label{fig:contrib-#1-#2-#3-#4}
  \end{subfigure}%
}

\begin{figure*}[t]
    \centering
    \includegraphics[width=1\textwidth]{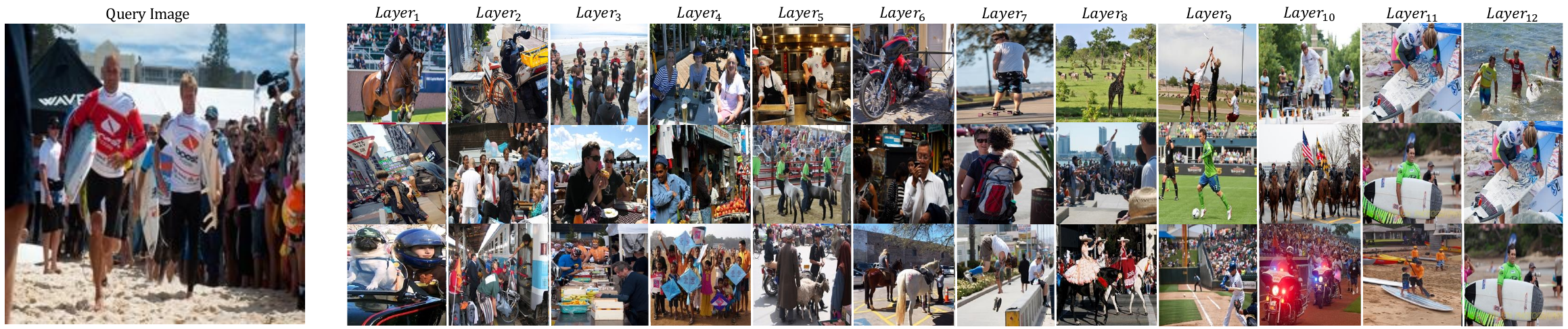}
    \caption{
    Layerwise visual retrieval using CLT sparse codes of CLS for a test image. Each row shows the top-3 retrieved training samples across transformer layers, revealing the semantic evolution of representations.
    }
    \label{fig:clt_retrieval_example}
\end{figure*}

\begin{figure*}[t]
    \centering
    \includegraphics[width=1\textwidth]{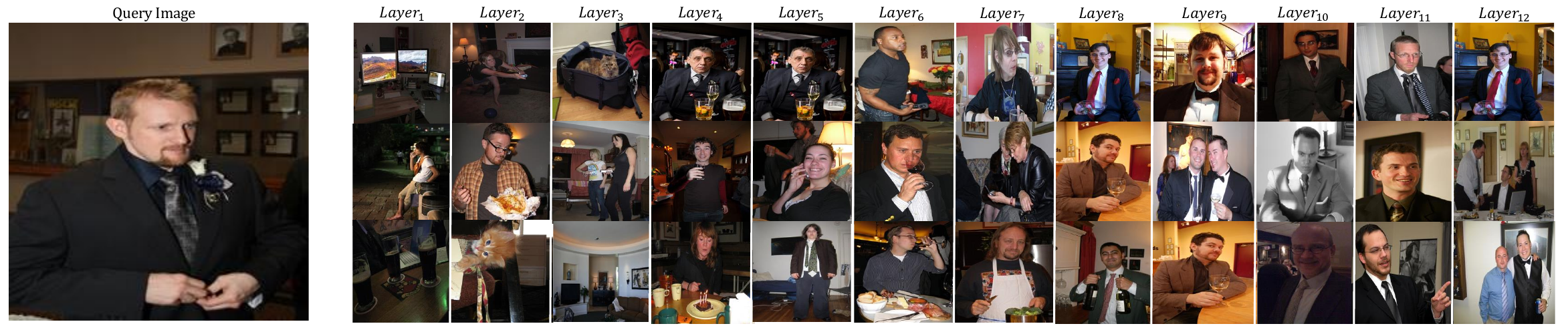}
    \caption{
    Layerwise visual retrieval using CLT sparse codes of Patches for a test image. Each row shows the top-3 retrieved training samples across transformer layers, revealing the semantic evolution of representations.
    }
    \label{fig:patches_clt_retrieval_example_sup}
\end{figure*}

\section{Visually Explainable Cross-Layer Contribution}
\label{sec:clt-retrieval}



The contribution scores $C_{s \to \ell}$ quantify \emph{which} layers matter, but not \emph{what} they represent. To obtain example-based explanations of cross-layer influence, we provide some evidence of a retrieval-based framework that operates directly on the CLT sparse codes. Recall that a target post-MLP representation $\hat{y}_L$ at the final layer can be decomposed as

\begin{equation*}
\hat{y}_L = W^{\text{dec}}_{0 \rightarrow L} z_0 + \dots + W^{\text{dec}}_{L \rightarrow L} z_L
\label{eq:decoder-sum}
\end{equation*}
where each term corresponds to the contribution of a previous layer’s sparse representation $z_i$, transformed by its decoder $W^{\text{dec}}_{i \rightarrow L}$. Instead of inspecting individual neurons $z_{ik}$ in isolation, we treat the full sparse vector $z_i$ as a feature and use it for layer-wise retrieval. 

Concretely, we construct an external corpus from the training images of each dataset. For each image $j$ and layer $i$, we pass the image through the CLT and extract its sparse codes $z_i^{(j)}$. We then aggregate token-level features (e.g., by averaging over spatial tokens) into a global sparse descriptor $z_{i,\mathrm{agg}}^{(j)}$ and index these descriptors into a per-layer FAISS database $D_i$ \cite{douze2025faiss,johnson2019billion}. 
At test time, for a given input image, we compute its sparse activations $\{z_i\}$ and use them to retrieve the top-$K$ most similar training images from each layer's index:
\begin{equation}
\mathcal{N}_i(z_i) = \operatorname*{arg\,topK}_{j} \, \mathrm{sim}\left( z_i, z_{i,\text{agg}}^{(j)} \right), 
\quad \text{for } i \leq \ell
\label{eq:knn-retrieval}
\end{equation}
This mirrors the reconstruction process $\hat{y}^\ell = \sum_{i \le \ell} z_i W^{i \rightarrow \ell}_{\text{dec}}$ by surfacing images that share similar latent activations at each contributing layer. Unlike decoding, this retrieval provides interpretable, layer-specific visual evidence of what each layer's sparse features contribute to the final representation.

Figures~\ref{fig:clt_retrieval_example} and~\ref{fig:patches_clt_retrieval_example_sup} illustrate our retrieval-based framework for a single query image, showing the top-3 retrieved training samples per layer. For the \texttt{[CLS]} token, retrievals exhibit a depth-wise progression from low-level to high-level semantics: shallow layers primarily group images by generic color and layout statistics; mid-level layers emphasize similar configurations of people and objects; and the deepest layers retrieve highly class-consistent examples (e.g., images depicting the same activity, i.e., surfing), exhibiting robustness to viewpoint and background variation. This depth-wise semantic sharpening aligns with our cross-layer contribution analysis and with the strong functional performance of CLS-only substitution: as features become more class-specific with depth, CLT-based surrogates can both reconstruct and visually explain the representations driving the final decision.

\begin{table}[t]
\centering
\caption{CLT surrogate faithfulness metrics.}
\label{tab:faithful_supplementary}
\resizebox{\columnwidth}{!}{%
\begin{tabular}{l|ccc|cc|ccc|cc}
\toprule
& \multicolumn{3}{c|}{\textbf{Pred.\ Distrib.}} & \multicolumn{2}{c|}{\textbf{Top-k Agree}} & \multicolumn{3}{c|}{\textbf{Embedding}} & \multicolumn{2}{c}{\textbf{Prompt Sens.}} \\
\textbf{Layers} & $\Delta$Acc & Flip$\downarrow$ & KL$\downarrow$ & Top-1 & Top-5 & Cos & CKA & Spear. & $r$ & KL$\downarrow$ \\
\midrule
0-11\textsuperscript{CLS}  & +0.07 & 10.4\% & .034 & 89.6 & 88.5 & .984 & .948 & .929 & .995 & .036 \\
7-11\textsuperscript{CLS}  & +0.01 & 9.5\%  & .030 & 90.5 & 89.0 & .985 & .954 & .937 & .992 & .031 \\
10-11\textsuperscript{CLS} & $-$0.07 & 7.2\% & .018 & 92.8 & 91.2 & .991 & .979 & .973 & .994 & .019 \\
11\textsuperscript{CLS}   & $-$0.04 & 5.4\% & .010 & 94.6 & 93.2 & .996 & .991 & .991 & .997 & .010 \\
\midrule
0-11\textsuperscript{Patches}  & $-$11.1 & 35.2\% & .412 & 64.8 & 70.0 & .957 & .817 & .759 & .942 & .383 \\
7-11\textsuperscript{Patches}  & +0.05 & 15.0\% & .079 & 85.0 & 85.8 & .990 & .972 & .957 & .989 & .073 \\
10-11\textsuperscript{Patches} & $-$0.13 & 2.5\% & .002 & 97.5 & 96.8 & .999 & .999 & .998 & .996 & .002 \\
\bottomrule
\end{tabular}%
}
\end{table}

\section{Additional Metrics for CLTs' faithfulness}

To provide further evidence of CLT's faithfulness to the original model, we evaluate distributional alignment (KL, flip rate), top-$k$ agreement, embedding geometry (cos/CKA/Spearman), and prompt sensitivity across 18 templates
for the ViT-B/32 model on CIFAR100.
Table~\ref{tab:faithful_supplementary} shows that in the regimes the CLTs faithfully reconstruct the original model's representations (CLS across layers; late-layer replacement), the surrogate closely matches the teacher beyond accuracy (e.g., KL$<0.035$, CKA$>0.94$, prompt-trend $r>0.98$). These results also validate the faithful CLT late-layer patch replacement (e.g., $10$--$11^{P}$: KL$=0.002$, flip$=2.5\%$, CKA$=0.999$). In contrast, early patch cascades degrade (e.g., $0$--$11^{P}$: KL$=0.412$, flip$=35.2\%$), as identified in Table~\ref{tab:clt-top1-compact}.

\section{Training Details}

\paragraph{Teacher models and datasets.}
All Cross-Layer Transcoders (CLTs) are trained on top of frozen CLIP image encoders with ViT-B/32 and ViT-B/16 backbones. For each backbone, we consider three datasets: CIFAR-100, COCO, and ImageNet-100. For every (dataset, backbone) pair we train three separate CLT variants, one for each sparsifier: JumpReLU, ReLU-Top-$k$, and Abs-Top-$k$ (with $k=128$). CLTs only access the internal activations of the teacher ViT and do not modify or finetune the underlying CLIP parameters.

\paragraph{Supervision and targets.}
Let $x_\ell \in \mathbb{R}^{T \times d}$ denote the post-attention (LN2), pre-MLP activations at layer $\ell$, and let $y_\ell = \mathrm{MLP}_\ell(x_\ell)$ be the corresponding post-MLP outputs. For each image, we run the frozen teacher ViT once and cache $(x_\ell, y_\ell)$ for all layers $\ell = 0, \dots, 11$ and all tokens (both [CLS] and patch tokens). CLTs are trained to reconstruct $y_\ell$ from sparse features computed from $\{x_i\}_{i \le \ell}$, using teacher-forcing at training time; i.e., all CLT inputs come from the unmodified teacher trajectory.

\paragraph{Optimization hyperparameters.}
For all datasets, backbones, and sparsifiers we train CLTs with the AdamW optimizer, learning rate $2 \times 10^{-4}$, and an expansion factor of $16$. Each CLT is trained for $10$ epochs over the corresponding dataset, using all tokens (both [CLS] and patch tokens) in the loss. Hyperparameters are shared across datasets and backbones.

\section{Reconstruction Accuracy across Layers}
\begin{figure*}[t]
    \centering
    \includegraphics[width=0.32\textwidth]{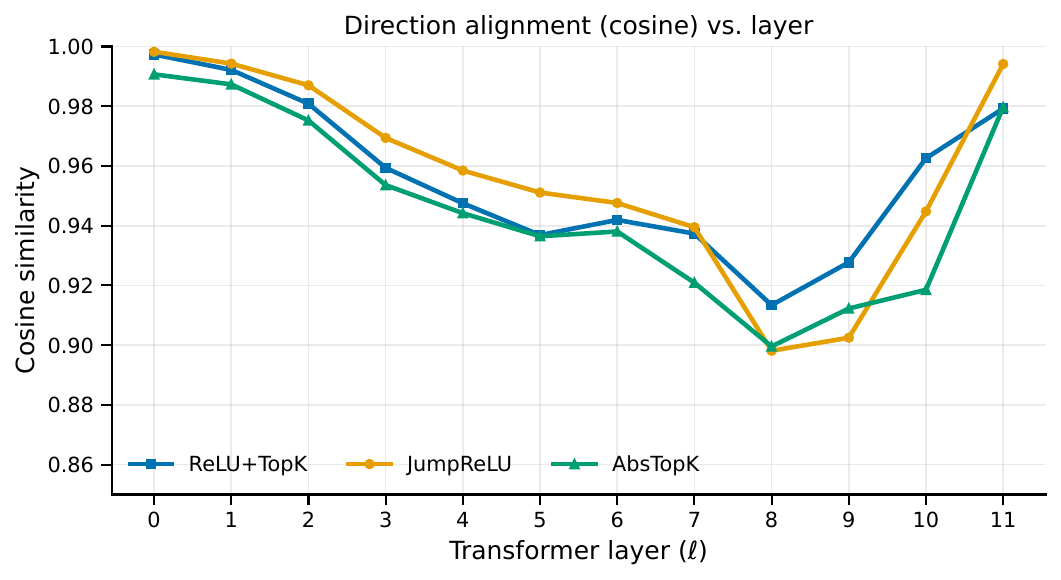}
    \includegraphics[width=0.32\textwidth]{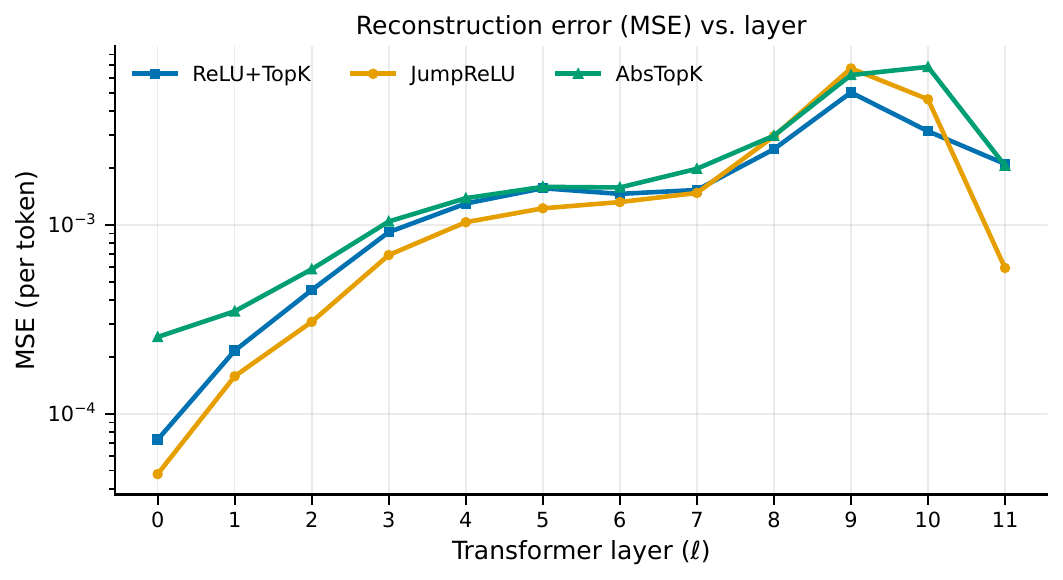}
    \includegraphics[width=0.32\textwidth]{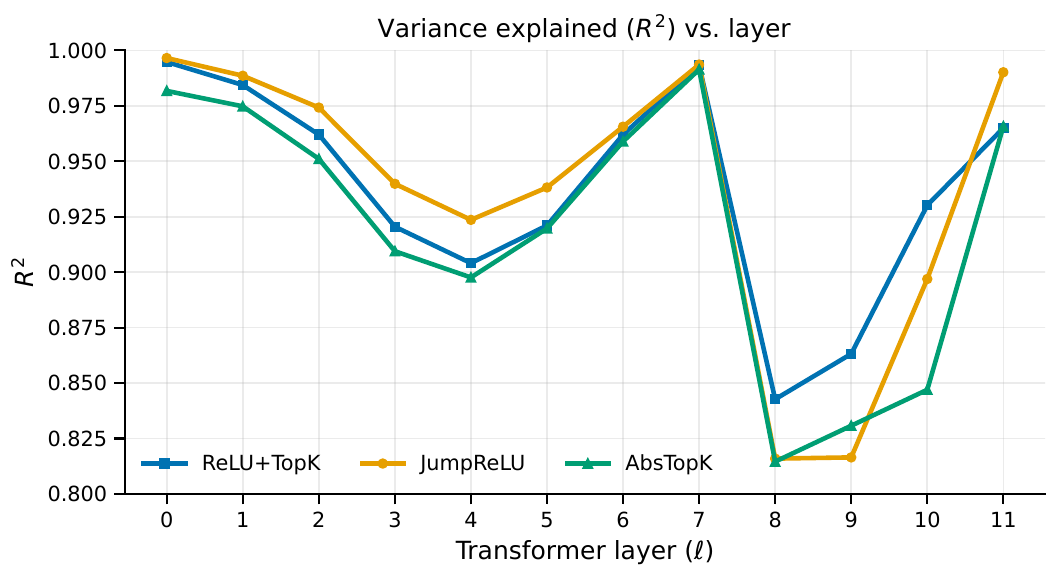}
    \caption{
    Reconstruction performance of CLTs across transformer layers on CIFAR-100 using ViT-B/16. We report cosine similarity (left), MSE per token in log scale (center), and $R^2$ (right) for \textsc{JumpReLU}, \textsc{ReLU-Top-}$k$, and \textsc{Abs-Top-}$k$ sparsity variants ($k{=}128$), averaged across all tokens in the validation set.
    }
    \label{fig:clt_reconstruction_cifar100_vitb16}
\end{figure*}

\begin{figure*}[t]
    \centering
    \includegraphics[width=0.32\textwidth]{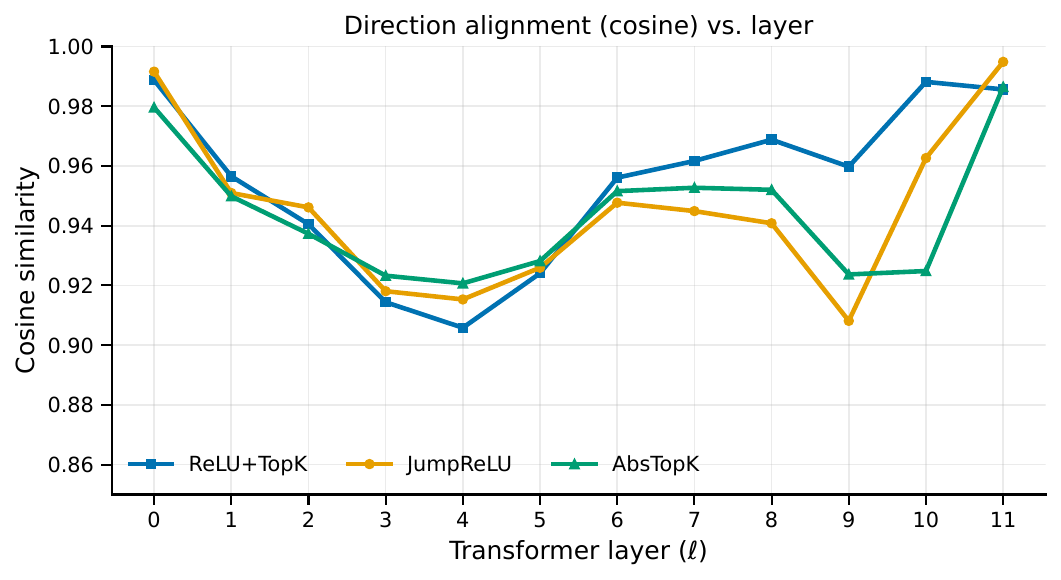}
    \includegraphics[width=0.32\textwidth]{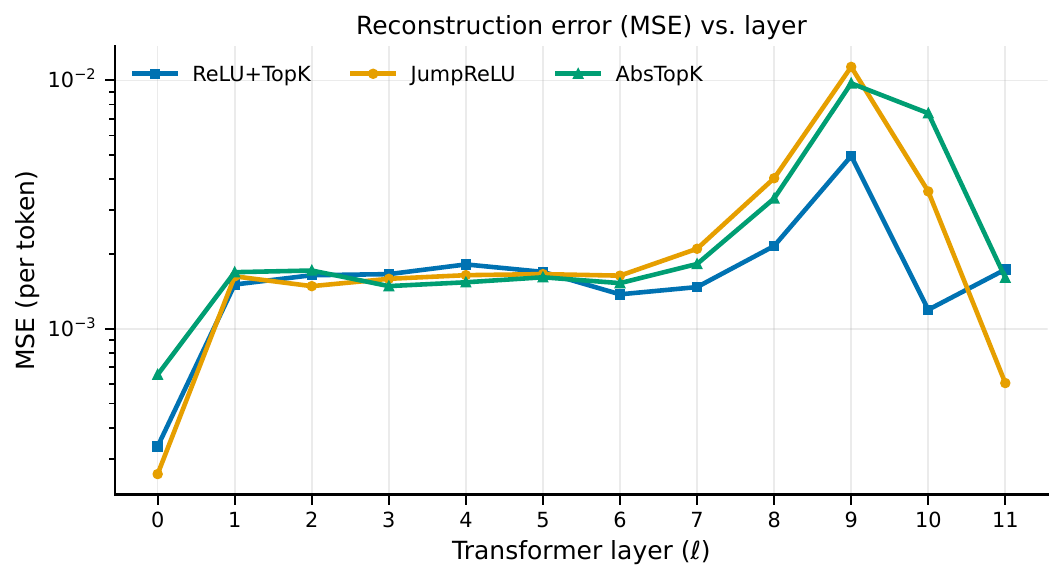}
    \includegraphics[width=0.32\textwidth]{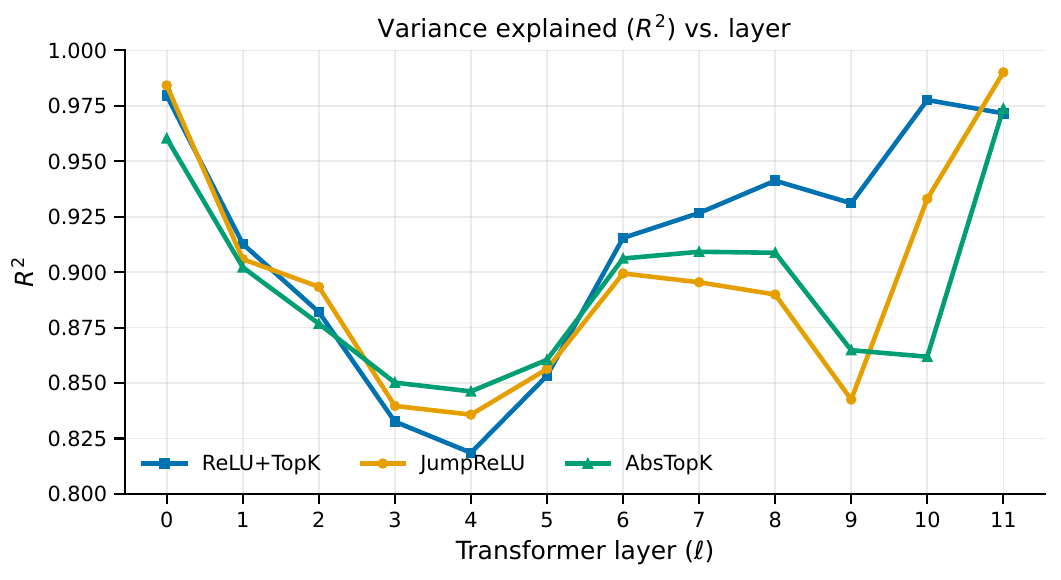}
    \caption{
    Reconstruction performance of CLTs across transformer layers on COCO using ViT-B/32. We show cosine similarity (left), MSE per token (log scale, center), and $R^2$ (right) for the three sparsity variants \textsc{JumpReLU}, \textsc{ReLU-Top-}$k$, and \textsc{Abs-Top-}$k$ ($k{=}128$), averaged across all tokens in the validation set.
    }
    \label{fig:clt_reconstruction_coco_vitb32}
\end{figure*}

\begin{figure*}[t]
    \centering
    \includegraphics[width=0.32\textwidth]{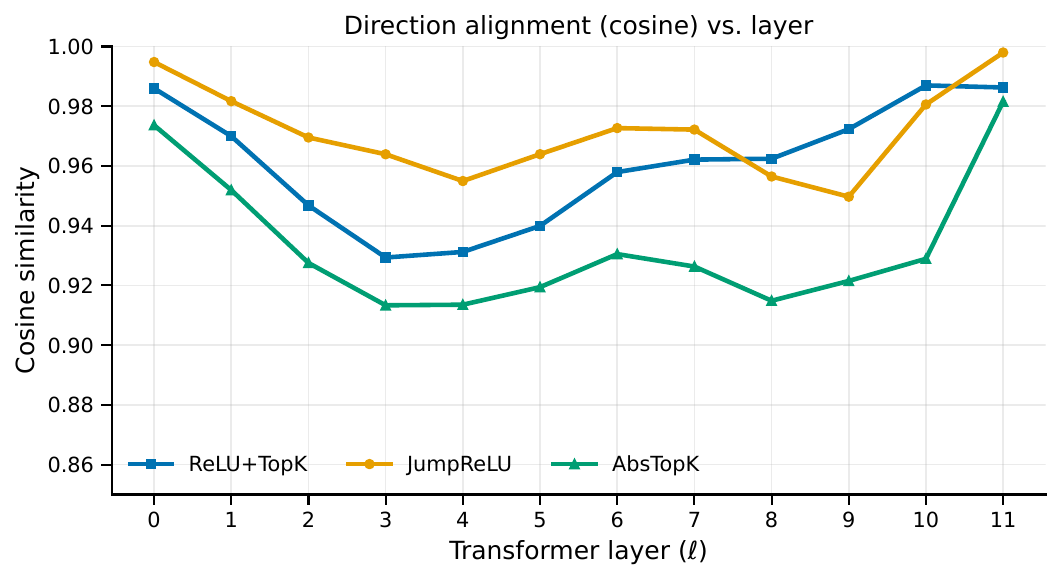}
    \includegraphics[width=0.32\textwidth]{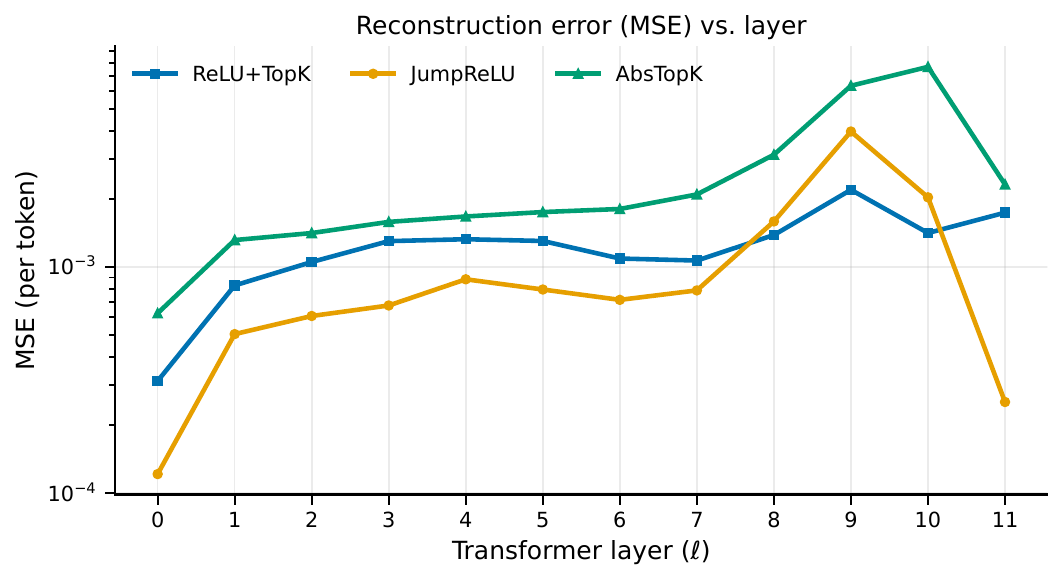}
    \includegraphics[width=0.32\textwidth]{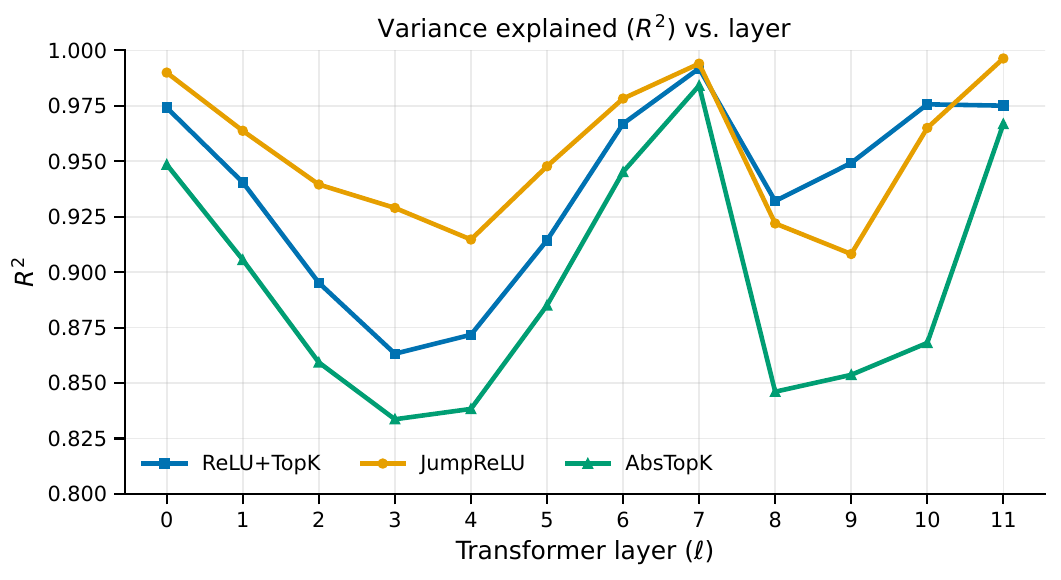}
    \caption{
    Reconstruction performance of CLTs across transformer layers on COCO using ViT-B/16. As in the main paper, we compare \textsc{JumpReLU}, \textsc{ReLU-Top-}$k$, and \textsc{Abs-Top-}$k$ ($k{=}128$) with cosine similarity (left), MSE per token (log scale, center), and $R^2$ (right), averaged over the validation set.
    }
    \label{fig:clt_reconstruction_coco_vitb16}
\end{figure*}

\begin{figure*}[t]
    \centering
    \includegraphics[width=0.32\textwidth]{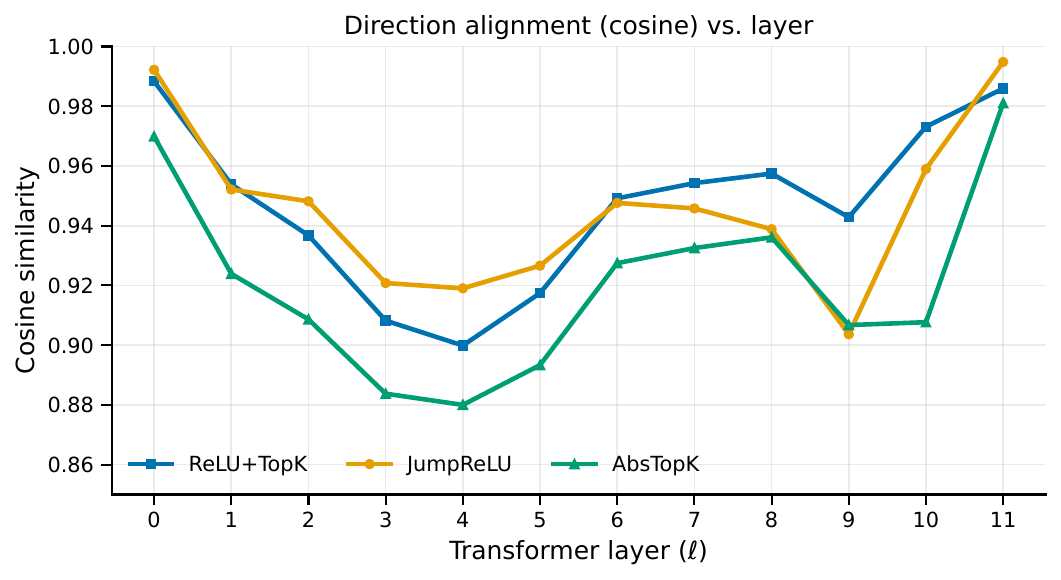}
    \includegraphics[width=0.32\textwidth]{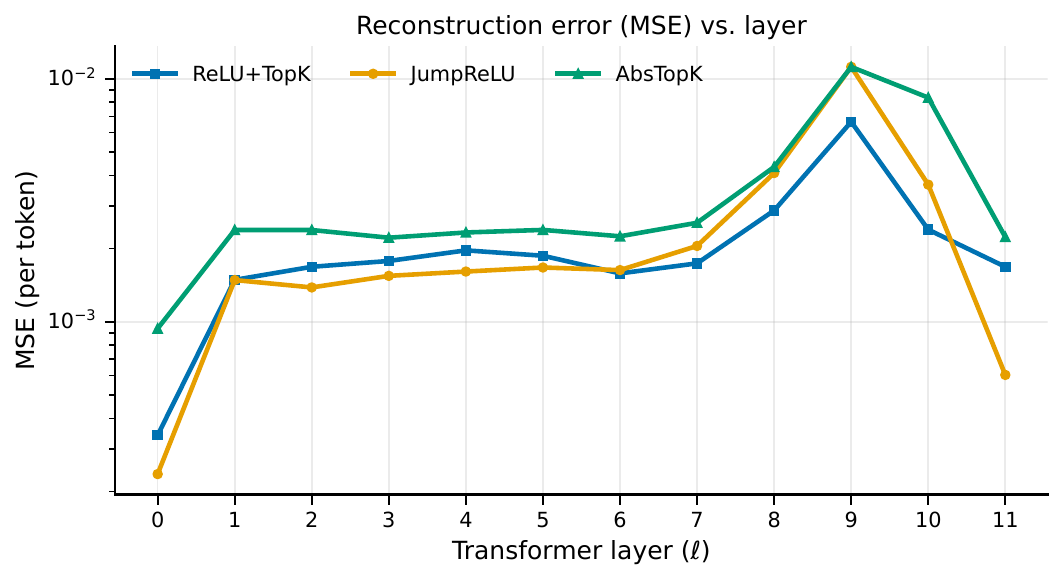}
    \includegraphics[width=0.32\textwidth]{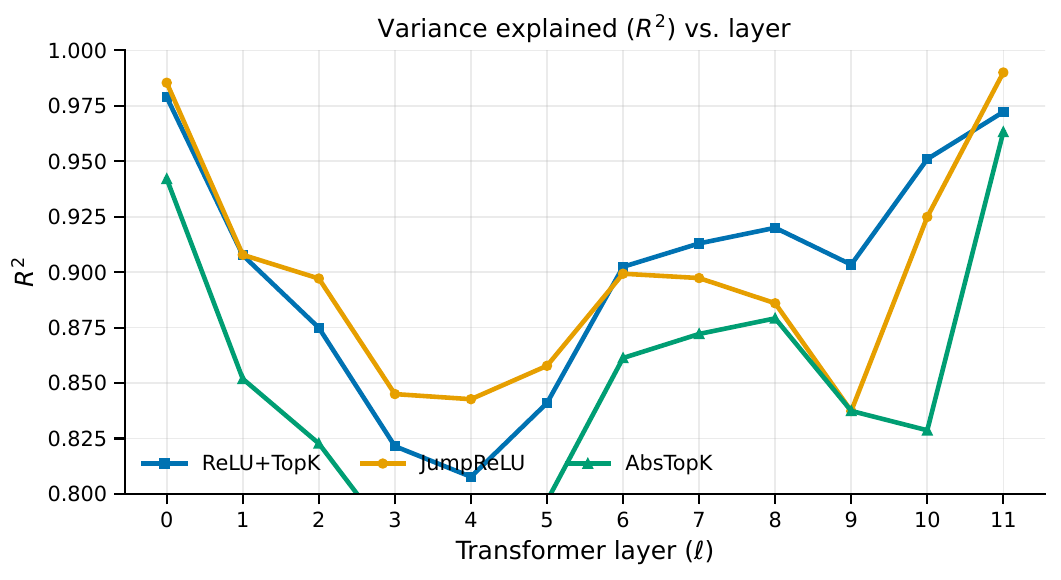}
    \caption{
    Reconstruction performance of CLTs across transformer layers on ImageNet-100 using ViT-B/32. We plot cosine similarity (left), MSE per token in log scale (center), and $R^2$ (right) for the three sparsity variants \textsc{JumpReLU}, \textsc{ReLU-Top-}$k$, and \textsc{Abs-Top-}$k$ ($k{=}128$), averaged across all tokens in the validation set.
    }
    \label{fig:clt_reconstruction_imagenet100_vitb32}
\end{figure*}

\begin{figure*}[t]
    \centering
    \includegraphics[width=0.32\textwidth]{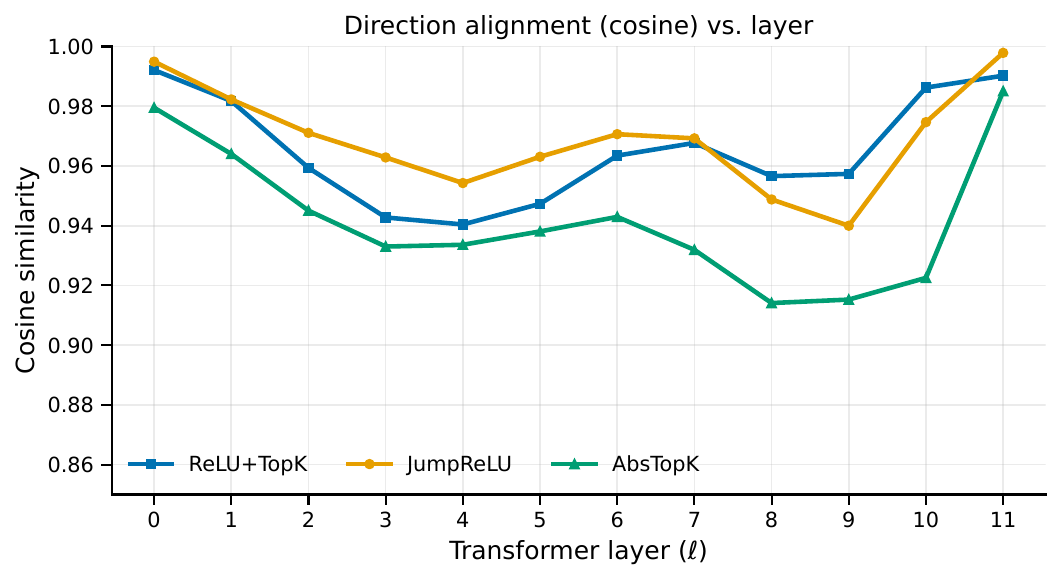}
    \includegraphics[width=0.32\textwidth]{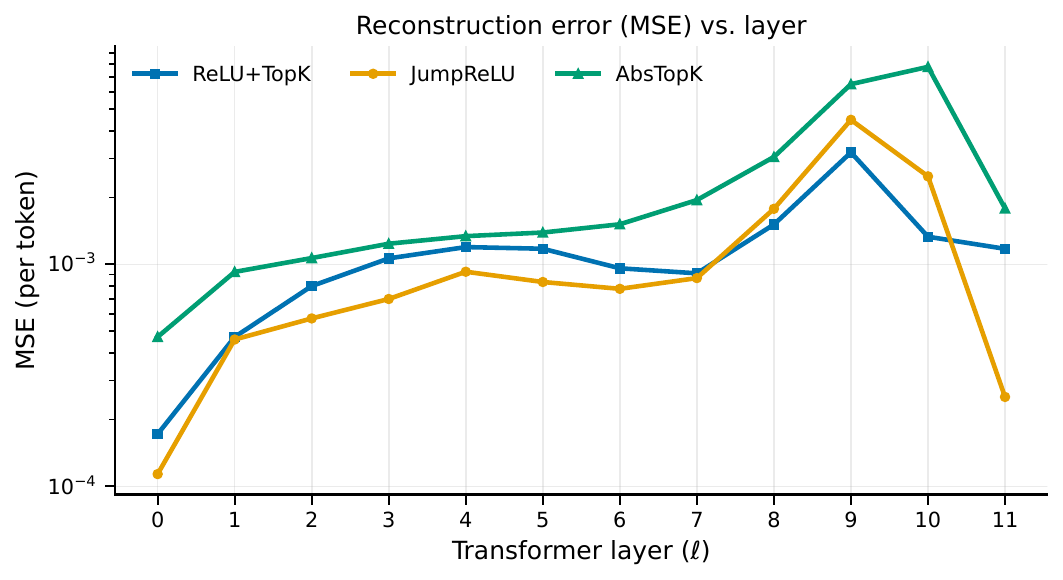}
    \includegraphics[width=0.32\textwidth]{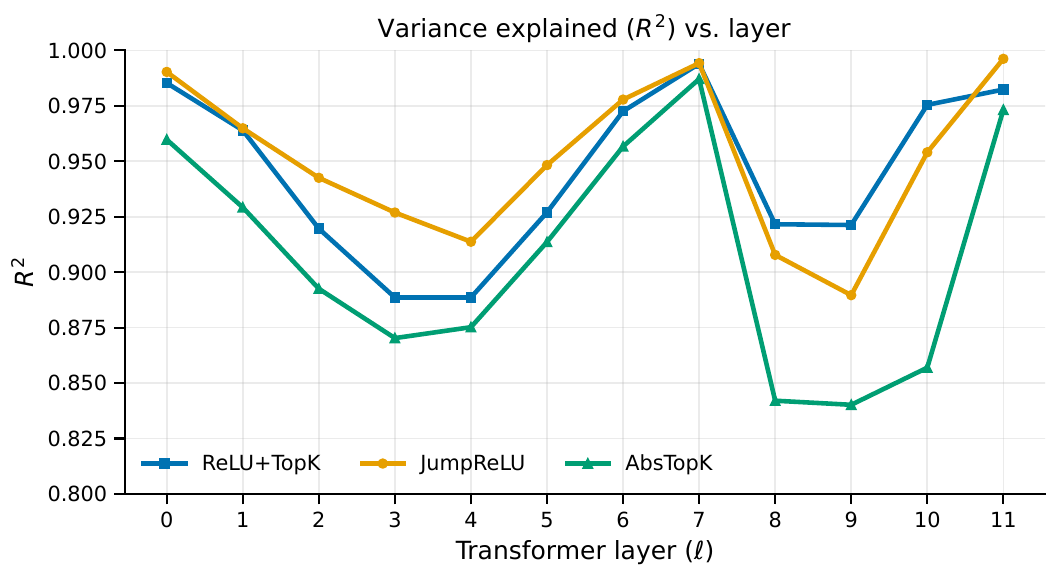}
    \caption{
    Reconstruction performance of CLTs across transformer layers on ImageNet-100 using ViT-B/16. Cosine similarity (left), MSE per token (log scale, center), and $R^2$ (right) are reported for \textsc{JumpReLU}, \textsc{ReLU-Top-}$k$, and \textsc{Abs-Top-}$k$ sparsity ($k{=}128$), averaged across all tokens in the test set.
    }
    \label{fig:clt_reconstruction_imagenet100_vitb16}
\end{figure*}

In Figures \ref{fig:clt_reconstruction_cifar100_vitb16}–\ref{fig:clt_reconstruction_imagenet100_vitb16}, we report the reconstruction quality of Cross-Layer Transcoders (CLTs) across all transformer layers on three datasets (CIFAR-100, COCO, and ImageNet-100) and two CLIP backbones (ViT-B/32 and ViT-B/16). For each configuration, we compare three sparsity variants, i.e., \textsc{JumpReLU}, \textsc{ReLU-Top-}$k$, and \textsc{Abs-Top-}$k$, using cosine similarity, mean squared error (log scale), and variance explained ($R^2$), averaged over all tokens in the test set.

\section{Classification Accuracy under Cascaded CLT Replacement}
We report top-1 classification accuracy (\%) under cascaded CLT replacement across all layers ($s \rightarrow 11$) in Figures \ref{tab:clt-top1-cifar100-vit-b32-cls-ranges-2}–\ref{tab:clt-top1-imagenet100-vit-b16-all-ranges}. For each dataset (CIFAR-100, COCO, ImageNet-100) and ViT backbone (ViT-B/32, ViT-B/16), we evaluate three sparsity mechanisms: \textsc{JumpReLU} (JR), \textsc{ReLU-Top-}$k$ (RTK), and \textsc{Abs-Top-}$k$ (ATK), under three token settings (CLS-only, patch-only, and all tokens). The replacement is performed progressively from early to late layers ($s = 0$ to $s = 11$), and results are compared to the frozen ViT baseline. We observe that across all datasets and backbones, CLS-token replacement achieves near-identical or slightly better accuracy compared to the original model. This shows that CLS tokens are robust to replacement. Regarding patch tokens, accuracy improves significantly as more layers are replaced, especially in the later layers.

\begin{table}[t]
\centering
\caption{Top-1 classification accuracy (\%) on CIFAR-100 for ViT-B/32 with CLS tokens, across layer ranges $s\rightarrow 11$. JR = JumpReLU, RTK = ReLU-Top-$k$, ATK = Abs-Top-$k$. Baseline top-1 accuracies (in \%) are ViT-B/32: 61.65 and ViT-B/16: 65.97.}
\label{tab:clt-top1-cifar100-vit-b32-cls-ranges-2}
\scriptsize
\begin{tabular}{lccc}
\toprule
\textbf{Range} & JR & RTK & ATK \\
\midrule
0$\rightarrow$11 & 61.33 & \textbf{61.74} & 61.43 \\
1$\rightarrow$11 & 61.43 & \textbf{61.76} & 61.56 \\
2$\rightarrow$11 & 61.39 & \textbf{61.72} & 61.61 \\
3$\rightarrow$11 & 61.38 & \textbf{61.69} & 61.58 \\
4$\rightarrow$11 & 61.20 & \textbf{61.76} & 61.66 \\
5$\rightarrow$11 & 61.21 & \textbf{61.59} & 61.48 \\
6$\rightarrow$11 & 61.43 & \textbf{61.58} & 61.39 \\
7$\rightarrow$11 & 61.41 & \textbf{61.86} & 61.26 \\
8$\rightarrow$11 & 61.62 & \textbf{61.90} & 61.23 \\
9$\rightarrow$11 & 61.31 & \textbf{61.85} & 61.10 \\
10$\rightarrow$11 & 61.06 & \textbf{61.39} & 61.16 \\
11$\rightarrow$11 & 61.23 & \textbf{61.31} & 61.03 \\
\bottomrule
\end{tabular}
\end{table}

\begin{table}[t]
\centering
\caption{Top-1 classification accuracy (\%) on CIFAR-100 for ViT-B/32 with Patches tokens, across layer ranges $s\rightarrow 11$. JR = JumpReLU, RTK = ReLU-Top-$k$, ATK = Abs-Top-$k$. Baseline top-1 accuracies (in \%) are ViT-B/32: 61.65 and ViT-B/16: 65.97.}
\label{tab:clt-top1-cifar100-vit-b32-patches-ranges}
\scriptsize
\begin{tabular}{lccc}
\toprule
\textbf{Range} & JR & RTK & ATK \\
\midrule
0$\rightarrow$11 & 49.63 & \textbf{51.12} & 48.68 \\
1$\rightarrow$11 & 49.92 & 51.61 & \textbf{52.40} \\
2$\rightarrow$11 & 51.95 & 52.86 & \textbf{53.46} \\
3$\rightarrow$11 & 53.65 & 54.20 & \textbf{55.83} \\
4$\rightarrow$11 & 56.59 & 57.02 & \textbf{58.86} \\
5$\rightarrow$11 & 58.89 & 59.52 & \textbf{61.38} \\
6$\rightarrow$11 & 60.39 & 61.16 & \textbf{63.06} \\
7$\rightarrow$11 & 61.51 & 61.69 & \textbf{63.18} \\
8$\rightarrow$11 & 61.83 & 62.04 & \textbf{63.53} \\
9$\rightarrow$11 & 61.82 & 61.73 & \textbf{62.57} \\
10$\rightarrow$11 & 61.49 & 61.25 & \textbf{61.56} \\
11$\rightarrow$11 & \textbf{61.65} & \textbf{61.65} & \textbf{61.65} \\
\bottomrule
\end{tabular}
\end{table}

\begin{table}[t]
\centering
\caption{Top-1 classification accuracy (\%) on CIFAR-100 for ViT-B/32 with All tokens, across layer ranges $s\rightarrow 11$. JR = JumpReLU, RTK = ReLU-Top-$k$, ATK = Abs-Top-$k$. Baseline top-1 accuracies (in \%) are ViT-B/32: 61.65 and ViT-B/16: 65.97.}
\label{tab:clt-top1-cifar100-vit-b32-all-ranges}
\scriptsize
\begin{tabular}{lccc}
\toprule
\textbf{Range} & JR & RTK & ATK \\
\midrule
0$\rightarrow$11 & 49.40 & \textbf{51.12} & 48.41 \\
1$\rightarrow$11 & 49.90 & 51.91 & \textbf{51.97} \\
2$\rightarrow$11 & 51.84 & \textbf{53.51} & 53.16 \\
3$\rightarrow$11 & 54.10 & 54.84 & \textbf{55.69} \\
4$\rightarrow$11 & 56.62 & 57.32 & \textbf{58.23} \\
5$\rightarrow$11 & 59.21 & 59.96 & \textbf{61.19} \\
6$\rightarrow$11 & 60.54 & 61.60 & \textbf{62.95} \\
7$\rightarrow$11 & 61.54 & 62.08 & \textbf{62.85} \\
8$\rightarrow$11 & 61.69 & 62.32 & \textbf{63.01} \\
9$\rightarrow$11 & 61.63 & 61.83 & \textbf{62.07} \\
10$\rightarrow$11 & 61.01 & \textbf{61.20} & 61.11 \\
11$\rightarrow$11 & 61.23 & \textbf{61.31} & 61.03 \\
\bottomrule
\end{tabular}
\end{table}

\begin{table}[t]
\centering
\caption{Top-1 classification accuracy (\%) on CIFAR-100 for ViT-B/16 with CLS tokens, across layer ranges $s\rightarrow 11$. JR = JumpReLU, RTK = ReLU-Top-$k$, ATK = Abs-Top-$k$. Baseline top-1 accuracies (in \%) are ViT-B/32: 61.65 and ViT-B/16: 65.97.}
\label{tab:clt-top1-cifar100-vit-b16-cls-ranges}
\scriptsize
\begin{tabular}{lccc}
\toprule
\textbf{Range} & JR & RTK & ATK \\
\midrule
0$\rightarrow$11 & \textbf{66.04} & 65.92 & 65.38 \\
1$\rightarrow$11 & \textbf{66.05} & 65.82 & 65.62 \\
2$\rightarrow$11 & \textbf{66.02} & 65.78 & 65.73 \\
3$\rightarrow$11 & \textbf{66.10} & 65.81 & 65.39 \\
4$\rightarrow$11 & \textbf{66.05} & 65.87 & 65.61 \\
5$\rightarrow$11 & \textbf{66.15} & 65.80 & 65.65 \\
6$\rightarrow$11 & \textbf{66.16} & 66.08 & 65.71 \\
7$\rightarrow$11 & \textbf{66.15} & \textbf{66.15} & 65.86 \\
8$\rightarrow$11 & \textbf{66.12} & 65.98 & 66.02 \\
9$\rightarrow$11 & \textbf{66.12} & 65.93 & 66.01 \\
10$\rightarrow$11 & \textbf{66.06} & 65.90 & 65.90 \\
11$\rightarrow$11 & 65.87 & \textbf{66.00} & 65.65 \\
\bottomrule
\end{tabular}
\end{table}

\begin{table}[t]
\centering
\caption{Top-1 classification accuracy (\%) on CIFAR-100 for ViT-B/16 with Patches tokens, across layer ranges $s\rightarrow 11$. JR = JumpReLU, RTK = ReLU-Top-$k$, ATK = Abs-Top-$k$. Baseline top-1 accuracies (in \%) are ViT-B/32: 61.65 and ViT-B/16: 65.97.}
\label{tab:clt-top1-cifar100-vit-b16-patches-ranges}
\scriptsize
\begin{tabular}{lccc}
\toprule
\textbf{Range} & JR & RTK & ATK \\
\midrule
0$\rightarrow$11 & \textbf{62.40} & 58.57 & 56.59 \\
1$\rightarrow$11 & \textbf{62.58} & 59.40 & 60.68 \\
2$\rightarrow$11 & \textbf{63.15} & 60.07 & 61.97 \\
3$\rightarrow$11 & \textbf{63.54} & 60.76 & 63.31 \\
4$\rightarrow$11 & 64.29 & 62.32 & \textbf{64.85} \\
5$\rightarrow$11 & 65.24 & 63.50 & \textbf{66.11} \\
6$\rightarrow$11 & 65.72 & 64.76 & \textbf{66.85} \\
7$\rightarrow$11 & 66.05 & 65.48 & \textbf{67.24} \\
8$\rightarrow$11 & 65.68 & 65.79 & \textbf{66.79} \\
9$\rightarrow$11 & 65.65 & 65.50 & \textbf{66.06} \\
10$\rightarrow$11 & 65.97 & 65.82 & \textbf{66.13} \\
11$\rightarrow$11 & \textbf{65.97} & \textbf{65.97} & \textbf{65.97} \\
\bottomrule
\end{tabular}
\end{table}

\begin{table}[t]
\centering
\caption{Top-1 classification accuracy (\%) on CIFAR-100 for ViT-B/16 with All tokens, across layer ranges $s\rightarrow 11$. JR = JumpReLU, RTK = ReLU-Top-$k$, ATK = Abs-Top-$k$. Baseline top-1 accuracies (in \%) are ViT-B/32: 61.65 and ViT-B/16: 65.97.}
\label{tab:clt-top1-cifar100-vit-b16-all-ranges}
\scriptsize
\begin{tabular}{lccc}
\toprule
\textbf{Range} & JR & RTK & ATK \\
\midrule
0$\rightarrow$11 & \textbf{62.45} & 58.82 & 56.72 \\
1$\rightarrow$11 & \textbf{62.70} & 59.23 & 60.30 \\
2$\rightarrow$11 & \textbf{62.95} & 59.74 & 61.54 \\
3$\rightarrow$11 & \textbf{63.71} & 60.75 & 62.87 \\
4$\rightarrow$11 & 64.14 & 62.16 & \textbf{64.57} \\
5$\rightarrow$11 & 65.20 & 63.60 & \textbf{65.86} \\
6$\rightarrow$11 & 66.08 & 64.73 & \textbf{66.31} \\
7$\rightarrow$11 & 66.22 & 65.56 & \textbf{67.08} \\
8$\rightarrow$11 & 65.72 & 65.84 & \textbf{66.65} \\
9$\rightarrow$11 & 65.77 & 65.51 & \textbf{66.05} \\
10$\rightarrow$11 & 65.91 & \textbf{65.98} & 65.89 \\
11$\rightarrow$11 & 65.87 & \textbf{66.00} & 65.65 \\
\bottomrule
\end{tabular}
\end{table}

\begin{table}[t]
\centering
\caption{Top-1 classification accuracy (\%) on COCO for ViT-B/32 with CLS tokens, across layer ranges $s\rightarrow 11$. JR = JumpReLU, RTK = ReLU-Top-$k$, ATK = Abs-Top-$k$. Baseline top-1 accuracies (in \%) are ViT-B/32: 43.12 and ViT-B/16: 43.56.}
\label{tab:clt-top1-coco-vit-b32-cls-ranges}
\scriptsize
\begin{tabular}{lccc}
\toprule
\textbf{Range} & JR & RTK & ATK \\
\midrule
0$\rightarrow$11 & 43.12 & \textbf{43.36} & 43.00 \\
1$\rightarrow$11 & 43.14 & \textbf{43.32} & 43.08 \\
2$\rightarrow$11 & 43.10 & \textbf{43.26} & 42.96 \\
3$\rightarrow$11 & 43.08 & \textbf{43.22} & 43.12 \\
4$\rightarrow$11 & 43.18 & \textbf{43.22} & 43.06 \\
5$\rightarrow$11 & 43.22 & \textbf{43.30} & 43.06 \\
6$\rightarrow$11 & \textbf{43.30} & 43.24 & 43.22 \\
7$\rightarrow$11 & \textbf{43.26} & 43.24 & 43.20 \\
8$\rightarrow$11 & 43.32 & \textbf{43.40} & 43.26 \\
9$\rightarrow$11 & 43.16 & \textbf{43.34} & 42.92 \\
10$\rightarrow$11 & 43.04 & \textbf{43.10} & 42.90 \\
11$\rightarrow$11 & 43.24 & \textbf{43.26} & 43.00 \\
\bottomrule
\end{tabular}
\end{table}

\begin{table}[t]
\centering
\caption{Top-1 classification accuracy (\%) on COCO for ViT-B/32 with Patches tokens, across layer ranges $s\rightarrow 11$. JR = JumpReLU, RTK = ReLU-Top-$k$, ATK = Abs-Top-$k$. Baseline top-1 accuracies (in \%) are ViT-B/32: 43.12 and ViT-B/16: 43.56.}
\label{tab:clt-top1-coco-vit-b32-patches-ranges}
\scriptsize
\begin{tabular}{lccc}
\toprule
\textbf{Range} & JR & RTK & ATK \\
\midrule
0$\rightarrow$11 & 39.04 & 38.94 & \textbf{39.68} \\
1$\rightarrow$11 & 39.38 & 39.44 & \textbf{40.02} \\
2$\rightarrow$11 & 39.54 & 40.24 & \textbf{40.42} \\
3$\rightarrow$11 & 40.58 & 40.92 & \textbf{41.48} \\
4$\rightarrow$11 & 41.38 & 41.88 & \textbf{42.10} \\
5$\rightarrow$11 & 42.06 & \textbf{42.28} & \textbf{42.28} \\
6$\rightarrow$11 & 42.48 & 42.58 & \textbf{42.60} \\
7$\rightarrow$11 & 42.98 & \textbf{43.14} & 42.80 \\
8$\rightarrow$11 & \textbf{43.14} & 43.02 & 42.92 \\
9$\rightarrow$11 & \textbf{43.14} & 42.92 & 42.92 \\
10$\rightarrow$11 & 43.14 & \textbf{43.18} & 43.06 \\
11$\rightarrow$11 & \textbf{43.12} & \textbf{43.12} & \textbf{43.12} \\
\bottomrule
\end{tabular}
\end{table}

\begin{table}[t]
\centering
\caption{Top-1 classification accuracy (\%) on COCO for ViT-B/32 with All tokens, across layer ranges $s\rightarrow 11$. JR = JumpReLU, RTK = ReLU-Top-$k$, ATK = Abs-Top-$k$. Baseline top-1 accuracies (in \%) are ViT-B/32: 43.12 and ViT-B/16: 43.56.}
\label{tab:clt-top1-coco-vit-b32-all-ranges}
\scriptsize
\begin{tabular}{lccc}
\toprule
\textbf{Range} & JR & RTK & ATK \\
\midrule
0$\rightarrow$11 & 38.60 & 39.12 & \textbf{40.06} \\
1$\rightarrow$11 & 39.02 & 39.62 & \textbf{40.54} \\
2$\rightarrow$11 & 39.98 & 40.10 & \textbf{40.44} \\
3$\rightarrow$11 & 41.00 & 41.34 & \textbf{41.46} \\
4$\rightarrow$11 & 41.62 & 41.68 & \textbf{41.92} \\
5$\rightarrow$11 & 41.88 & \textbf{42.28} & 42.26 \\
6$\rightarrow$11 & \textbf{42.88} & 42.60 & 42.82 \\
7$\rightarrow$11 & \textbf{43.14} & 43.10 & 42.76 \\
8$\rightarrow$11 & \textbf{43.26} & \textbf{43.26} & 42.86 \\
9$\rightarrow$11 & \textbf{43.28} & 43.20 & 42.74 \\
10$\rightarrow$11 & 42.78 & \textbf{43.10} & 42.92 \\
11$\rightarrow$11 & 43.24 & \textbf{43.26} & 43.00 \\
\bottomrule
\end{tabular}
\end{table}

\begin{table}[t]
\centering
\caption{Top-1 classification accuracy (\%) on COCO for ViT-B/16 with CLS tokens, across layer ranges $s\rightarrow 11$. JR = JumpReLU, RTK = ReLU-Top-$k$, ATK = Abs-Top-$k$. Baseline top-1 accuracies (in \%) are ViT-B/32: 43.12 and ViT-B/16: 43.56.}
\label{tab:clt-top1-coco-vit-b16-cls-ranges}
\scriptsize
\begin{tabular}{lccc}
\toprule
\textbf{Range} & JR & RTK & ATK \\
\midrule
0$\rightarrow$11 & \textbf{43.62} & \textbf{43.62} & 42.76 \\
1$\rightarrow$11 & 43.56 & \textbf{43.62} & 42.40 \\
2$\rightarrow$11 & 43.60 & \textbf{43.64} & 42.34 \\
3$\rightarrow$11 & \textbf{43.68} & 43.64 & 42.64 \\
4$\rightarrow$11 & 43.66 & \textbf{43.72} & 42.64 \\
5$\rightarrow$11 & 43.58 & \textbf{43.80} & 42.98 \\
6$\rightarrow$11 & 43.56 & \textbf{43.66} & 43.50 \\
7$\rightarrow$11 & 43.52 & \textbf{43.64} & 43.34 \\
8$\rightarrow$11 & \textbf{43.72} & 43.66 & 43.68 \\
9$\rightarrow$11 & 43.82 & \textbf{43.84} & 43.24 \\
10$\rightarrow$11 & 43.52 & \textbf{43.88} & 43.38 \\
11$\rightarrow$11 & 43.50 & \textbf{43.62} & 43.28 \\
\bottomrule
\end{tabular}
\end{table}

\begin{table}[t]
\centering
\caption{Top-1 classification accuracy (\%) on COCO for ViT-B/16 with Patches tokens, across layer ranges $s\rightarrow 11$. JR = JumpReLU, RTK = ReLU-Top-$k$, ATK = Abs-Top-$k$. Baseline top-1 accuracies (in \%) are ViT-B/32: 43.12 and ViT-B/16: 43.56.}
\label{tab:clt-top1-coco-vit-b16-patches-ranges}
\scriptsize
\begin{tabular}{lccc}
\toprule
\textbf{Range} & JR & RTK & ATK \\
\midrule
0$\rightarrow$11 & \textbf{42.72} & 42.00 & 35.46 \\
1$\rightarrow$11 & \textbf{42.84} & 41.98 & 37.28 \\
2$\rightarrow$11 & \textbf{42.68} & 42.20 & 38.36 \\
3$\rightarrow$11 & \textbf{43.00} & 42.34 & 39.46 \\
4$\rightarrow$11 & \textbf{43.00} & 42.96 & 40.86 \\
5$\rightarrow$11 & 42.98 & \textbf{43.00} & 42.58 \\
6$\rightarrow$11 & \textbf{43.38} & 43.16 & 42.98 \\
7$\rightarrow$11 & 43.26 & \textbf{43.36} & 43.30 \\
8$\rightarrow$11 & 43.44 & 43.56 & \textbf{43.86} \\
9$\rightarrow$11 & 43.62 & 43.70 & \textbf{43.76} \\
10$\rightarrow$11 & 43.62 & 43.64 & \textbf{43.72} \\
11$\rightarrow$11 & \textbf{43.50} & \textbf{43.50} & \textbf{43.50} \\
\bottomrule
\end{tabular}
\end{table}

\begin{table}[t]
\centering
\caption{Top-1 classification accuracy (\%) on COCO for ViT-B/16 with All tokens, across layer ranges $s\rightarrow 11$. JR = JumpReLU, RTK = ReLU-Top-$k$, ATK = Abs-Top-$k$. Baseline top-1 accuracies (in \%) are ViT-B/32: 43.12 and ViT-B/16: 43.56.}
\label{tab:clt-top1-coco-vit-b16-all-ranges}
\scriptsize
\begin{tabular}{lccc}
\toprule
\textbf{Range} & JR & RTK & ATK \\
\midrule
0$\rightarrow$11 & \textbf{43.00} & 42.08 & 34.16 \\
1$\rightarrow$11 & \textbf{42.92} & 42.08 & 36.06 \\
2$\rightarrow$11 & \textbf{42.74} & 42.20 & 36.82 \\
3$\rightarrow$11 & \textbf{42.78} & 42.66 & 38.18 \\
4$\rightarrow$11 & 42.84 & \textbf{43.10} & 39.80 \\
5$\rightarrow$11 & 43.04 & \textbf{43.12} & 42.22 \\
6$\rightarrow$11 & 43.44 & \textbf{43.56} & 43.04 \\
7$\rightarrow$11 & \textbf{43.62} & 43.58 & 43.48 \\
8$\rightarrow$11 & \textbf{43.72} & 43.70 & 43.66 \\
9$\rightarrow$11 & 43.66 & \textbf{43.96} & 43.56 \\
10$\rightarrow$11 & 43.42 & \textbf{43.92} & 43.60 \\
11$\rightarrow$11 & 43.50 & \textbf{43.62} & 43.28 \\
\bottomrule
\end{tabular}
\end{table}

\begin{table}[t]
\centering
\caption{Top-1 classification accuracy (\%) on ImageNet-100 for ViT-B/32 with CLS tokens, across layer ranges $s\rightarrow 11$. JR = JumpReLU, RTK = ReLU-Top-$k$, ATK = Abs-Top-$k$. Baseline top-1 accuracies (in \%) are ViT-B/32: 80.42 and ViT-B/16: 84.34.}
\label{tab:clt-top1-imagenet100-vit-b32-cls-ranges}
\scriptsize
\begin{tabular}{lccc}
\toprule
\textbf{Range} & JR & RTK & ATK \\
\midrule
0$\rightarrow$11 & \textbf{80.92} & 80.86 & 80.26 \\
1$\rightarrow$11 & \textbf{80.84} & 80.82 & 80.32 \\
2$\rightarrow$11 & 80.82 & \textbf{80.86} & 80.24 \\
3$\rightarrow$11 & \textbf{80.86} & 80.78 & 80.16 \\
4$\rightarrow$11 & 80.80 & \textbf{80.86} & 80.24 \\
5$\rightarrow$11 & 80.74 & \textbf{80.86} & 80.44 \\
6$\rightarrow$11 & 80.68 & \textbf{80.84} & 79.92 \\
7$\rightarrow$11 & 80.64 & \textbf{80.72} & 80.16 \\
8$\rightarrow$11 & 80.66 & \textbf{80.68} & 80.06 \\
9$\rightarrow$11 & 80.46 & \textbf{80.60} & 80.10 \\
10$\rightarrow$11 & 80.56 & \textbf{80.72} & 80.18 \\
11$\rightarrow$11 & \textbf{80.54} & \textbf{80.54} & 80.36 \\
\bottomrule
\end{tabular}
\end{table}

\begin{table}[t]
\centering
\caption{Top-1 classification accuracy (\%) on ImageNet-100 for ViT-B/32 with Patches tokens, across layer ranges $s\rightarrow 11$. JR = JumpReLU, RTK = ReLU-Top-$k$, ATK = Abs-Top-$k$. Baseline top-1 accuracies (in \%) are ViT-B/32: 80.42 and ViT-B/16: 84.34.}
\label{tab:clt-top1-imagenet100-vit-b32-patches-ranges}
\scriptsize
\begin{tabular}{lccc}
\toprule
\textbf{Range} & JR & RTK & ATK \\
\midrule
0$\rightarrow$11 & \textbf{71.96} & 68.74 & 60.26 \\
1$\rightarrow$11 & \textbf{72.34} & 70.10 & 64.80 \\
2$\rightarrow$11 & \textbf{73.10} & 72.24 & 67.54 \\
3$\rightarrow$11 & \textbf{75.58} & 75.26 & 72.46 \\
4$\rightarrow$11 & \textbf{77.86} & 77.18 & 75.88 \\
5$\rightarrow$11 & 78.92 & \textbf{79.10} & 77.54 \\
6$\rightarrow$11 & \textbf{79.80} & 79.78 & 79.56 \\
7$\rightarrow$11 & 80.34 & 80.18 & \textbf{80.64} \\
8$\rightarrow$11 & 80.60 & 80.38 & \textbf{80.86} \\
9$\rightarrow$11 & 80.62 & 80.50 & \textbf{80.76} \\
10$\rightarrow$11 & \textbf{80.52} & 80.46 & 80.46 \\
11$\rightarrow$11 & \textbf{80.42} & \textbf{80.42} & \textbf{80.42} \\
\bottomrule
\end{tabular}
\end{table}

\begin{table}[t]
\centering
\caption{Top-1 classification accuracy (\%) on ImageNet-100 for ViT-B/32 with All tokens, across layer ranges $s\rightarrow 11$. JR = JumpReLU, RTK = ReLU-Top-$k$, ATK = Abs-Top-$k$. Baseline top-1 accuracies (in \%) are ViT-B/32: 80.42 and ViT-B/16: 84.34.}
\label{tab:clt-top1-imagenet100-vit-b32-all-ranges}
\scriptsize
\begin{tabular}{lccc}
\toprule
\textbf{Range} & JR & RTK & ATK \\
\midrule
0$\rightarrow$11 & \textbf{71.60} & 68.90 & 60.26 \\
1$\rightarrow$11 & \textbf{71.92} & 70.10 & 63.82 \\
2$\rightarrow$11 & \textbf{72.88} & 72.32 & 66.72 \\
3$\rightarrow$11 & 75.12 & \textbf{75.36} & 71.44 \\
4$\rightarrow$11 & \textbf{77.56} & 77.10 & 74.68 \\
5$\rightarrow$11 & 78.74 & \textbf{78.84} & 77.30 \\
6$\rightarrow$11 & 79.50 & \textbf{79.86} & 78.82 \\
7$\rightarrow$11 & 80.18 & \textbf{80.44} & 79.38 \\
8$\rightarrow$11 & \textbf{80.52} & 80.24 & 80.32 \\
9$\rightarrow$11 & \textbf{80.72} & 80.48 & 80.46 \\
10$\rightarrow$11 & 80.38 & \textbf{80.58} & 80.18 \\
11$\rightarrow$11 & \textbf{80.54} & \textbf{80.54} & 80.36 \\
\bottomrule
\end{tabular}
\end{table}

\begin{table}[t]
\centering
\caption{Top-1 classification accuracy (\%) on ImageNet-100 for ViT-B/16 with CLS tokens, across layer ranges $s\rightarrow 11$. JR = JumpReLU, RTK = ReLU-Top-$k$, ATK = Abs-Top-$k$. Baseline top-1 accuracies (in \%) are ViT-B/32: 80.42 and ViT-B/16: 84.34.}
\label{tab:clt-top1-imagenet100-vit-b16-cls-ranges}
\scriptsize
\begin{tabular}{lccc}
\toprule
\textbf{Range} & JR & RTK & ATK \\
\midrule
0$\rightarrow$11 & \textbf{84.54} & 84.02 & 83.28 \\
1$\rightarrow$11 & \textbf{84.50} & 84.04 & 83.52 \\
2$\rightarrow$11 & \textbf{84.44} & 83.98 & 83.50 \\
3$\rightarrow$11 & \textbf{84.56} & 84.08 & 83.74 \\
4$\rightarrow$11 & \textbf{84.44} & 84.04 & 83.82 \\
5$\rightarrow$11 & \textbf{84.42} & 84.10 & 84.20 \\
6$\rightarrow$11 & \textbf{84.52} & 84.04 & 84.24 \\
7$\rightarrow$11 & \textbf{84.66} & 84.16 & 84.40 \\
8$\rightarrow$11 & \textbf{84.60} & 84.30 & 84.46 \\
9$\rightarrow$11 & \textbf{84.62} & 84.30 & 84.42 \\
10$\rightarrow$11 & \textbf{84.56} & 84.38 & 84.34 \\
11$\rightarrow$11 & \textbf{84.46} & 84.36 & 84.36 \\
\bottomrule
\end{tabular}
\end{table}

\begin{table}[t]
\centering
\caption{Top-1 classification accuracy (\%) on ImageNet-100 for ViT-B/16 with Patches tokens, across layer ranges $s\rightarrow 11$. JR = JumpReLU, RTK = ReLU-Top-$k$, ATK = Abs-Top-$k$. Baseline top-1 accuracies (in \%) are ViT-B/32: 80.42 and ViT-B/16: 84.34.}
\label{tab:clt-top1-imagenet100-vit-b16-patches-ranges}
\scriptsize
\begin{tabular}{lccc}
\toprule
\textbf{Range} & JR & RTK & ATK \\
\midrule
0$\rightarrow$11 & \textbf{83.04} & 81.40 & 73.06 \\
1$\rightarrow$11 & \textbf{83.00} & 81.44 & 76.74 \\
2$\rightarrow$11 & \textbf{83.18} & 81.88 & 77.54 \\
3$\rightarrow$11 & \textbf{83.16} & 82.76 & 79.36 \\
4$\rightarrow$11 & \textbf{83.56} & 83.34 & 80.98 \\
5$\rightarrow$11 & 83.78 & \textbf{83.88} & 82.50 \\
6$\rightarrow$11 & \textbf{84.20} & 84.12 & 83.26 \\
7$\rightarrow$11 & 83.94 & \textbf{84.24} & 83.84 \\
8$\rightarrow$11 & 84.02 & 84.24 & \textbf{84.26} \\
9$\rightarrow$11 & 84.20 & 84.22 & \textbf{84.72} \\
10$\rightarrow$11 & 84.20 & 84.22 & \textbf{84.58} \\
11$\rightarrow$11 & \textbf{84.34} & \textbf{84.34} & \textbf{84.34} \\
\bottomrule
\end{tabular}
\end{table}

\begin{table}[t]
\centering
\caption{Top-1 classification accuracy (\%) on ImageNet-100 for ViT-B/16 with All tokens, across layer ranges $s\rightarrow 11$. JR = JumpReLU, RTK = ReLU-Top-$k$, ATK = Abs-Top-$k$. Baseline top-1 accuracies (in \%) are ViT-B/32: 80.42 and ViT-B/16: 84.34.}
\label{tab:clt-top1-imagenet100-vit-b16-all-ranges}
\scriptsize
\begin{tabular}{lccc}
\toprule
\textbf{Range} & JR & RTK & ATK \\
\midrule
0$\rightarrow$11 & \textbf{83.12} & 80.94 & 70.78 \\
1$\rightarrow$11 & \textbf{83.08} & 81.58 & 74.70 \\
2$\rightarrow$11 & \textbf{83.28} & 81.90 & 76.30 \\
3$\rightarrow$11 & \textbf{83.00} & 82.64 & 78.08 \\
4$\rightarrow$11 & \textbf{83.62} & 83.08 & 80.12 \\
5$\rightarrow$11 & \textbf{84.16} & 83.86 & 81.76 \\
6$\rightarrow$11 & \textbf{84.46} & 83.92 & 82.56 \\
7$\rightarrow$11 & 84.28 & \textbf{84.32} & 83.34 \\
8$\rightarrow$11 & \textbf{84.38} & 84.10 & 84.08 \\
9$\rightarrow$11 & 84.20 & 84.22 & \textbf{84.74} \\
10$\rightarrow$11 & 84.40 & 84.36 & \textbf{84.74} \\
11$\rightarrow$11 & \textbf{84.46} & 84.36 & 84.36 \\
\bottomrule
\end{tabular}
\end{table}

\section{Cross-Layer Contribution Scores}
\begin{figure*}[t]
  \centering
  \begin{subfigure}[t]{0.32\linewidth}
    \centering
    \includegraphics[width=\linewidth]{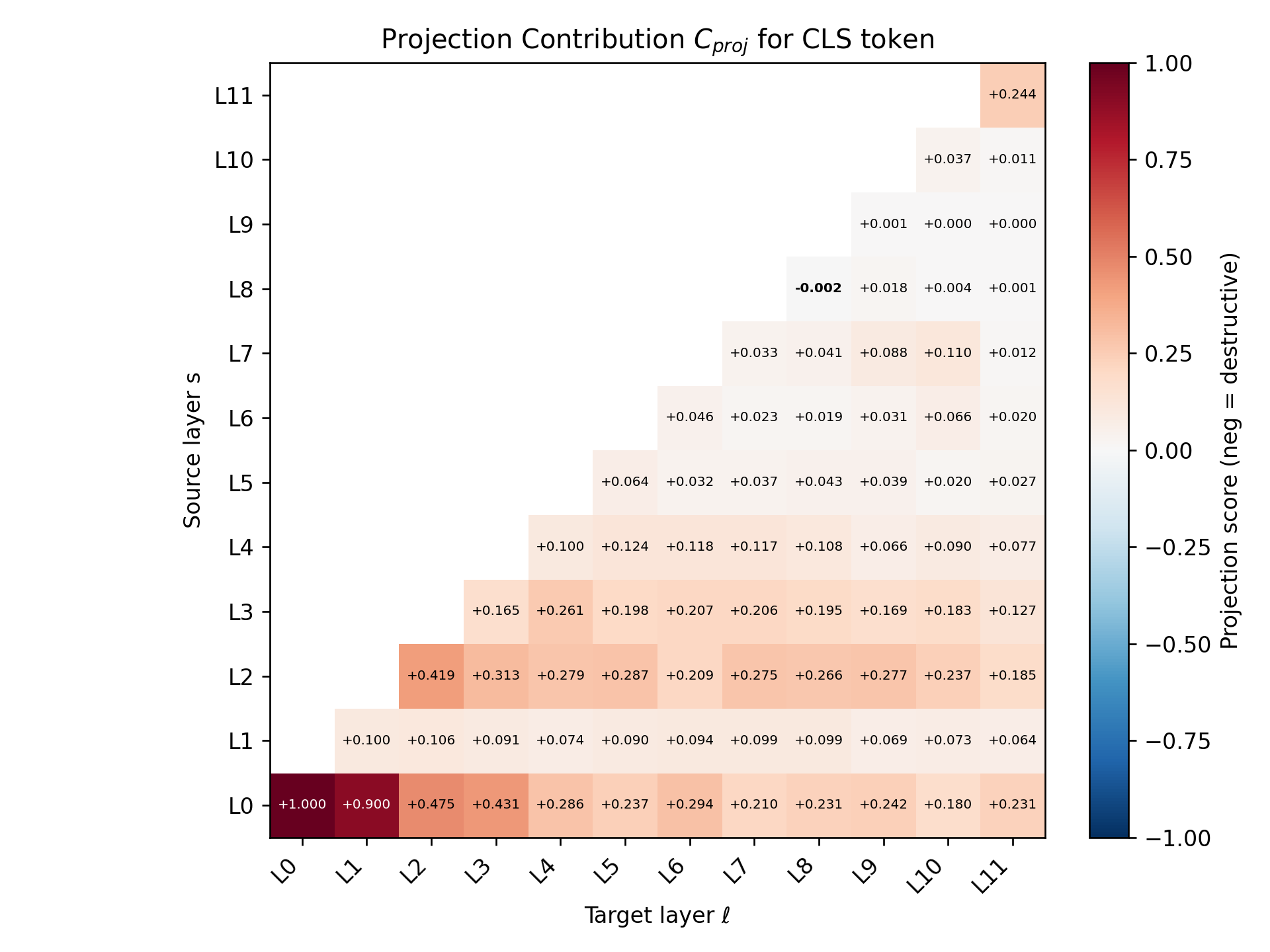}
    \caption{JumpReLU, CLS}
    \label{fig:contrib-v32-cifar100-jr-cls}
  \end{subfigure}
  \hfill
  \begin{subfigure}[t]{0.32\linewidth}
    \centering
    \includegraphics[width=\linewidth]{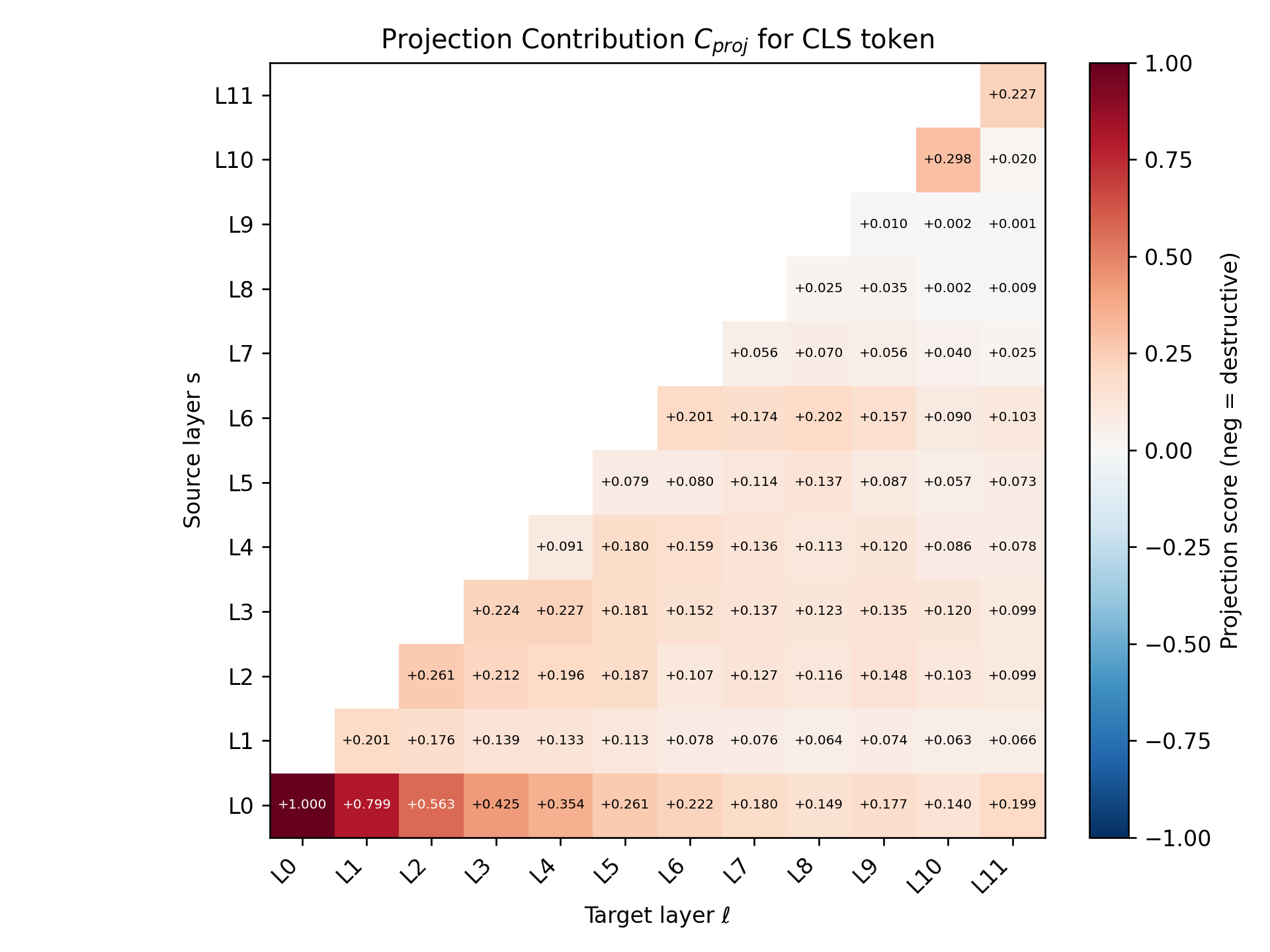}
    \caption{ReLU-Top-$k$, CLS}
    \label{fig:contrib-v32-cifar100-rtk-cls}
  \end{subfigure}
  \hfill
  \begin{subfigure}[t]{0.32\linewidth}
    \centering
    \includegraphics[width=\linewidth]{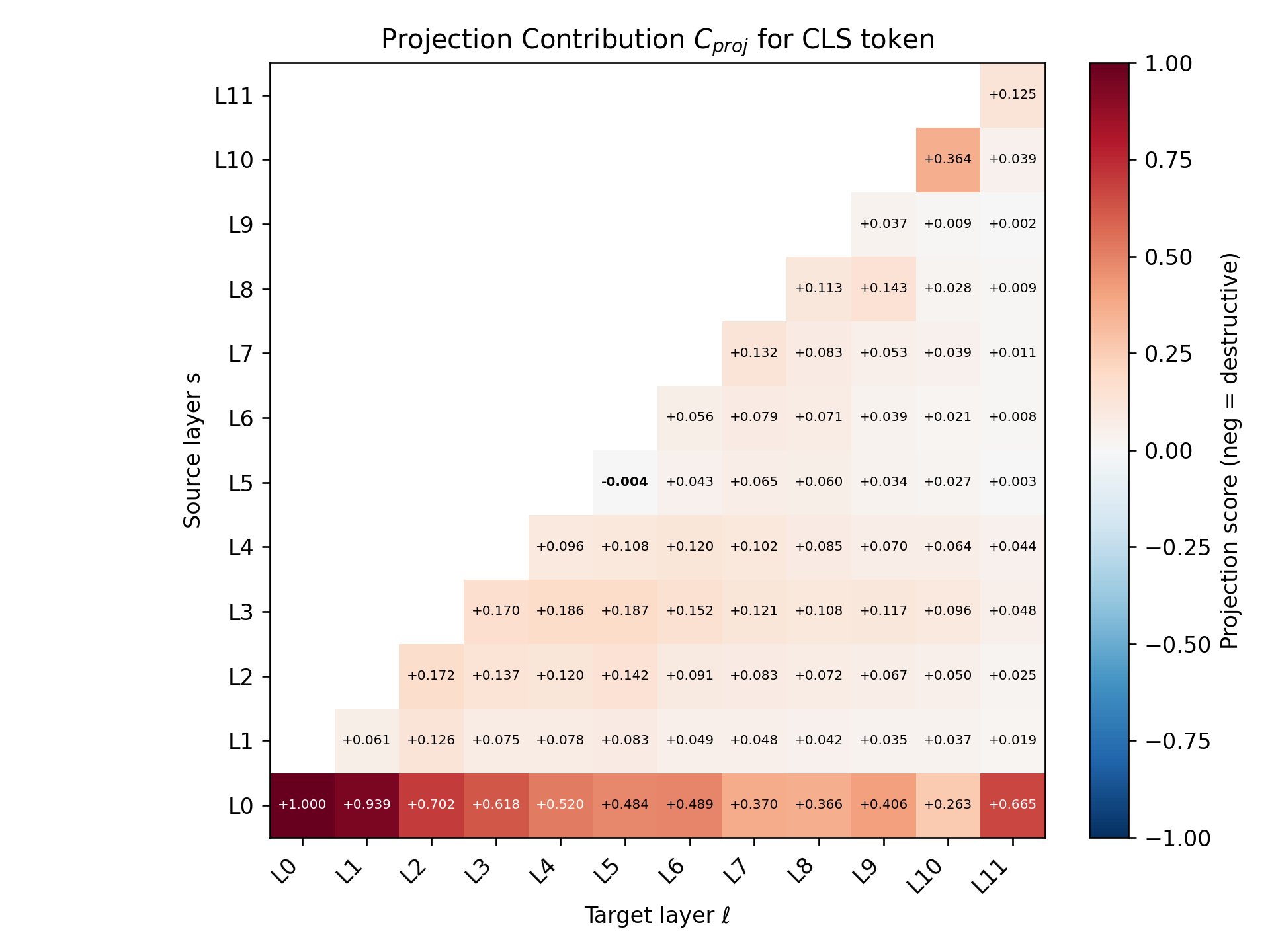}
    \caption{Abs-Top-$k$, CLS}
    \label{fig:contrib-v32-cifar100-atk-cls}
  \end{subfigure}

  \vspace{0.6em}

  \begin{subfigure}[t]{0.32\linewidth}
    \centering
    \includegraphics[width=\linewidth]{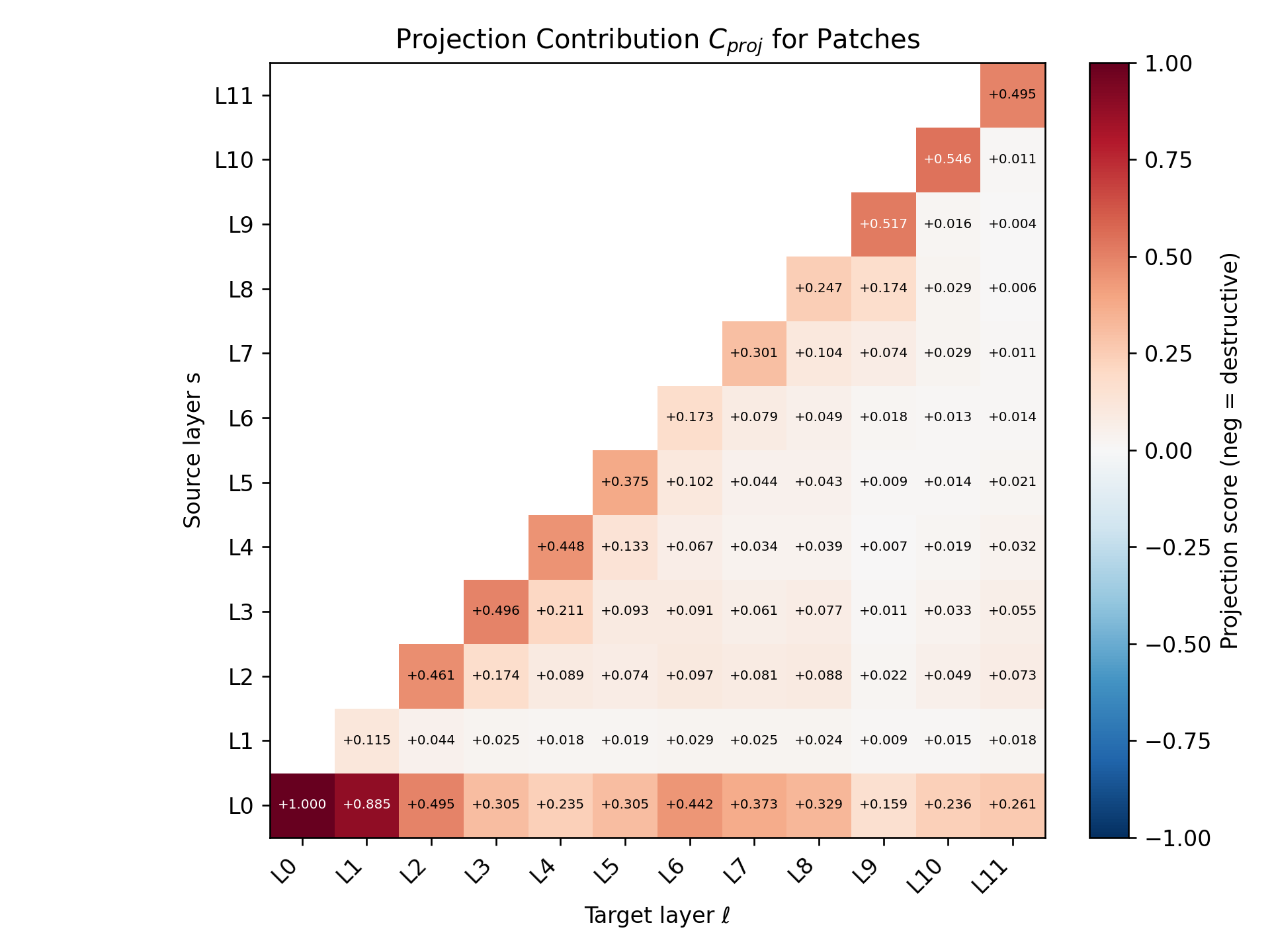}
    \caption{JumpReLU, patches}
    \label{fig:contrib-v32-cifar100-jr-patches}
  \end{subfigure}
  \hfill
  \begin{subfigure}[t]{0.32\linewidth}
    \centering
    \includegraphics[width=\linewidth]{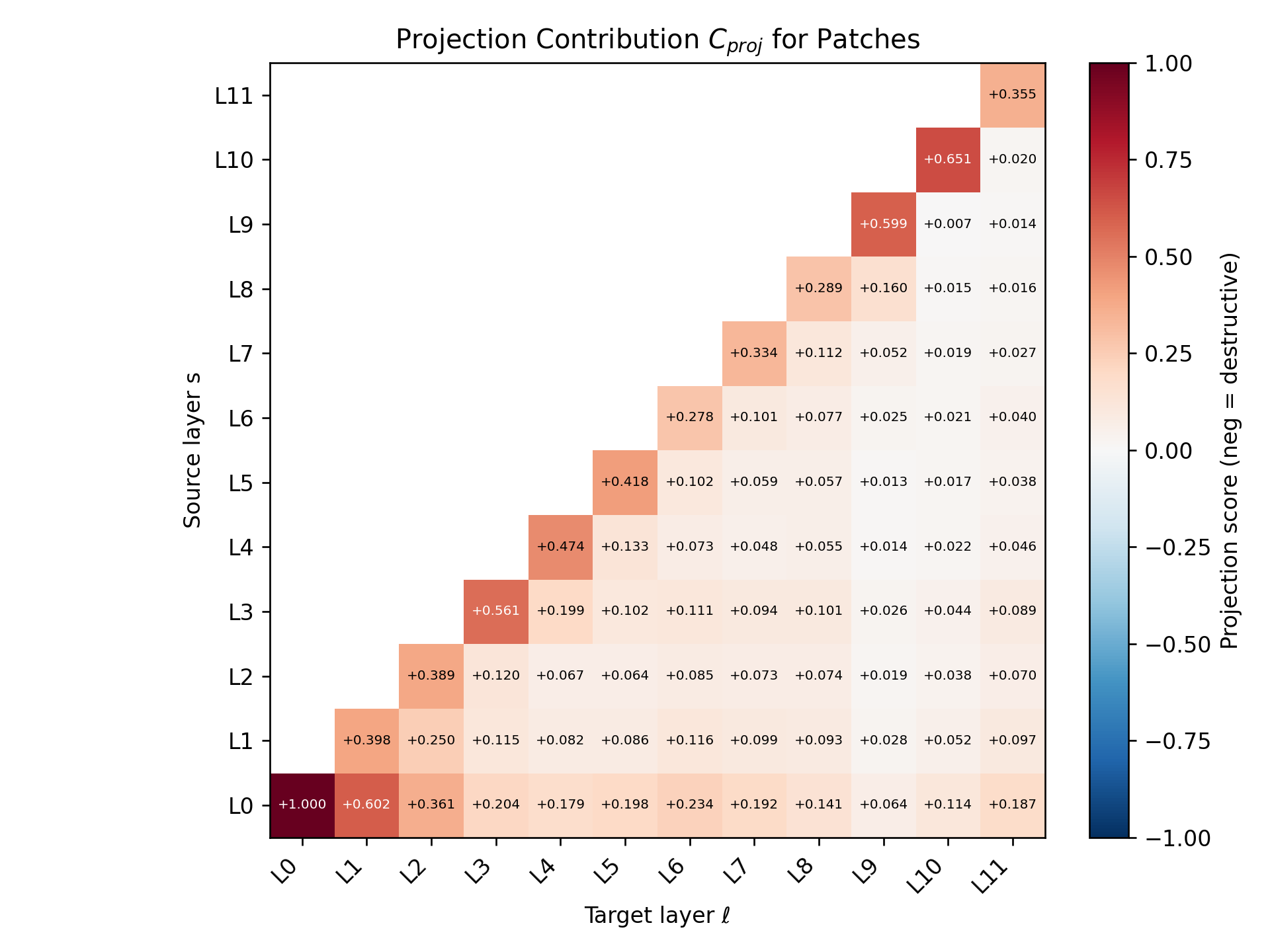}
    \caption{ReLU-Top-$k$, patches}
    \label{fig:contrib-v32-cifar100-rtk-patches}
  \end{subfigure}
  \hfill
  \begin{subfigure}[t]{0.32\linewidth}
    \centering
    \includegraphics[width=\linewidth]{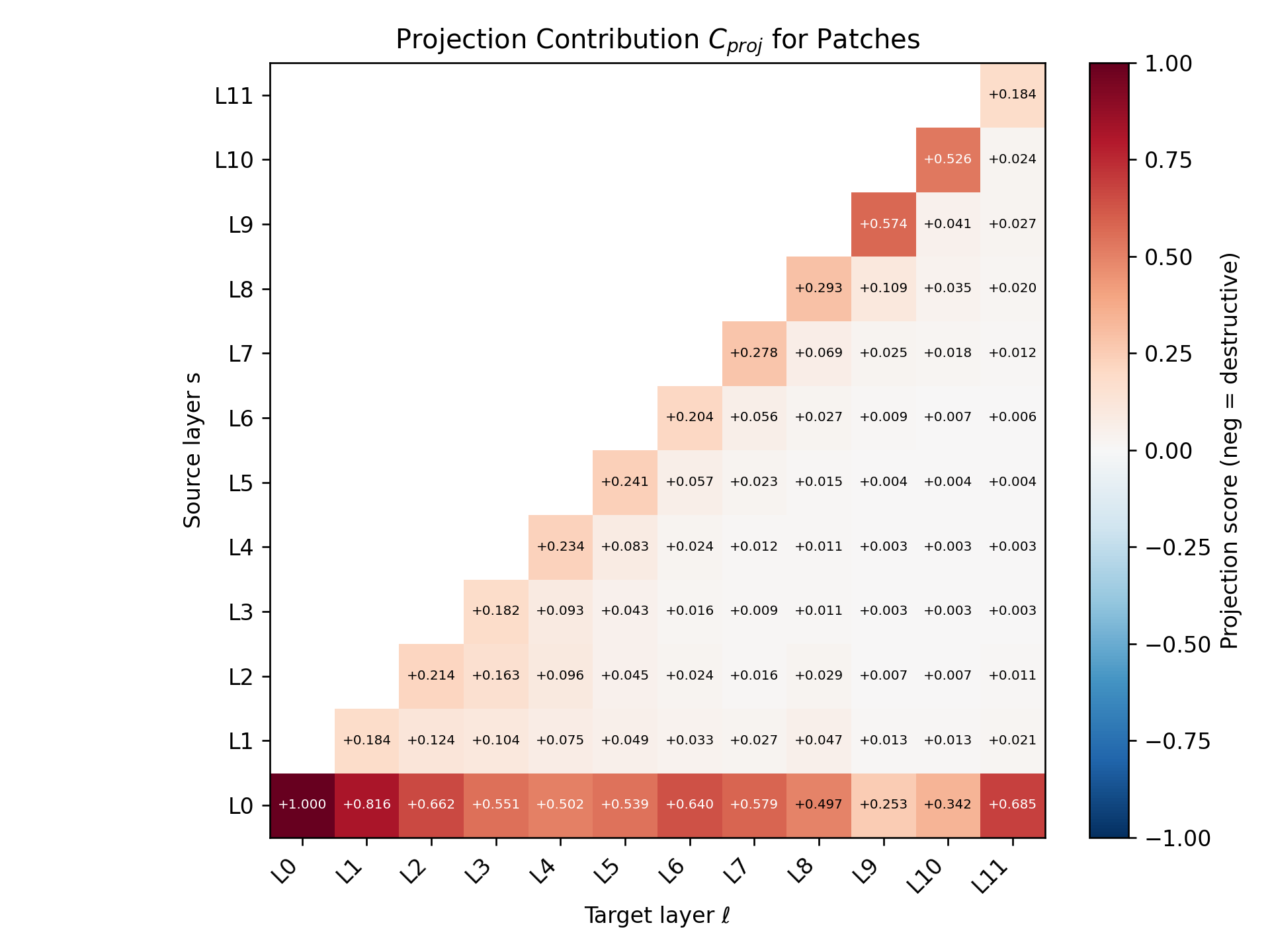}
    \caption{Abs-Top-$k$, patches}
    \label{fig:contrib-v32-cifar100-atk-patches}
  \end{subfigure}

  \caption{
  Cross-layer contribution scores $C_{s \rightarrow \ell}$ on CIFAR-100
  with ViT-B/32. Columns vary the sparsifier (JumpReLU, ReLU-Top-$k$, Abs-Top-$k$)
  and rows show CLS (top) and patch tokens (bottom). Each heatmap visualizes the
  proportional contribution of source layer $s$ to the reconstructed activation at
  target layer $\ell$, averaged over the validation set.
  }
  \label{fig:contrib-v32-cifar100-grid}
\end{figure*}

\begin{figure*}[t]
  \centering
  \begin{subfigure}[t]{0.32\linewidth}
    \centering
    \includegraphics[width=\linewidth]{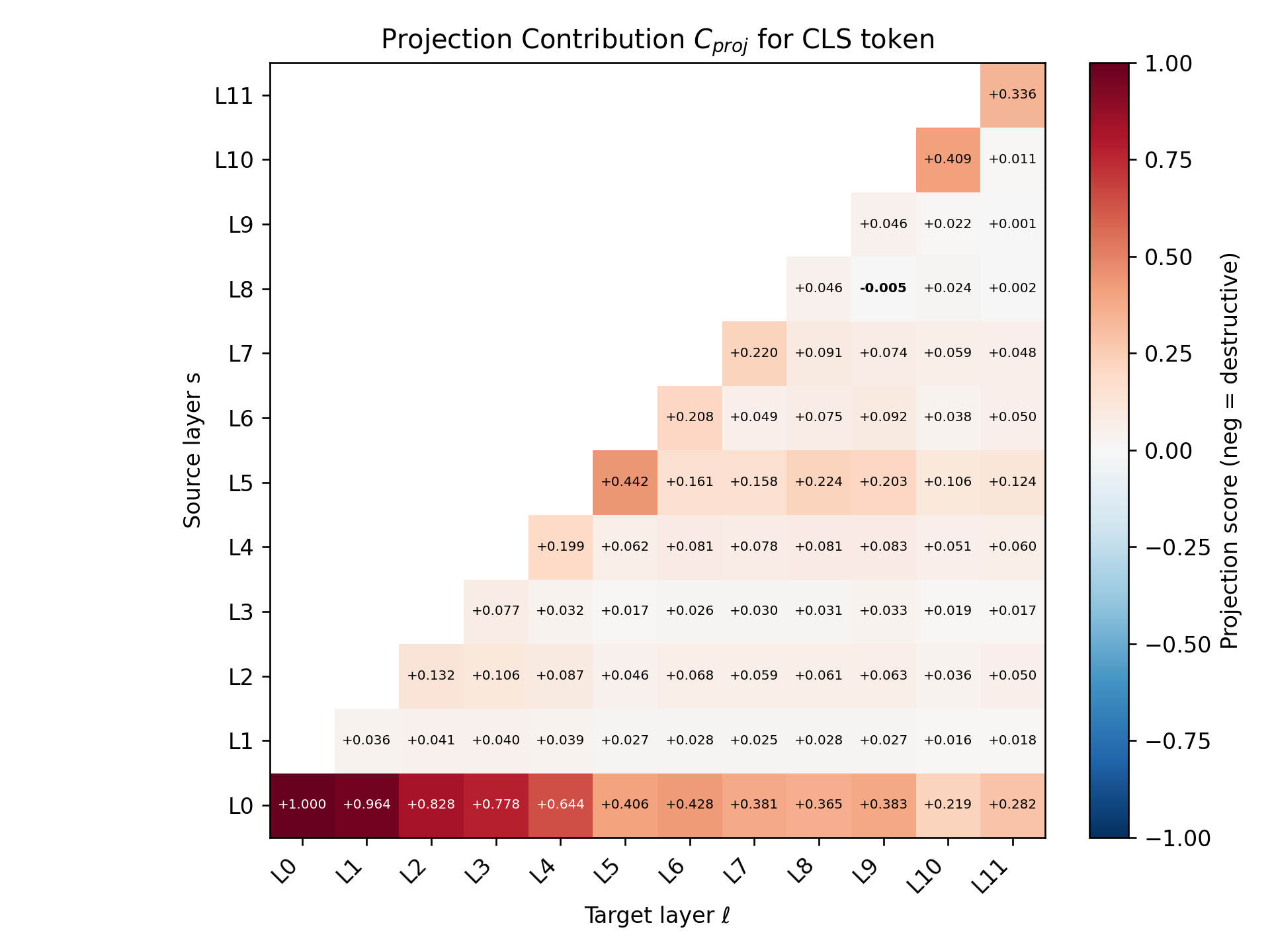}
    \caption{JumpReLU, CLS}
    \label{fig:contrib-v16-cifar100-jr-cls}
  \end{subfigure}
  \hfill
  \begin{subfigure}[t]{0.32\linewidth}
    \centering
    \includegraphics[width=\linewidth]{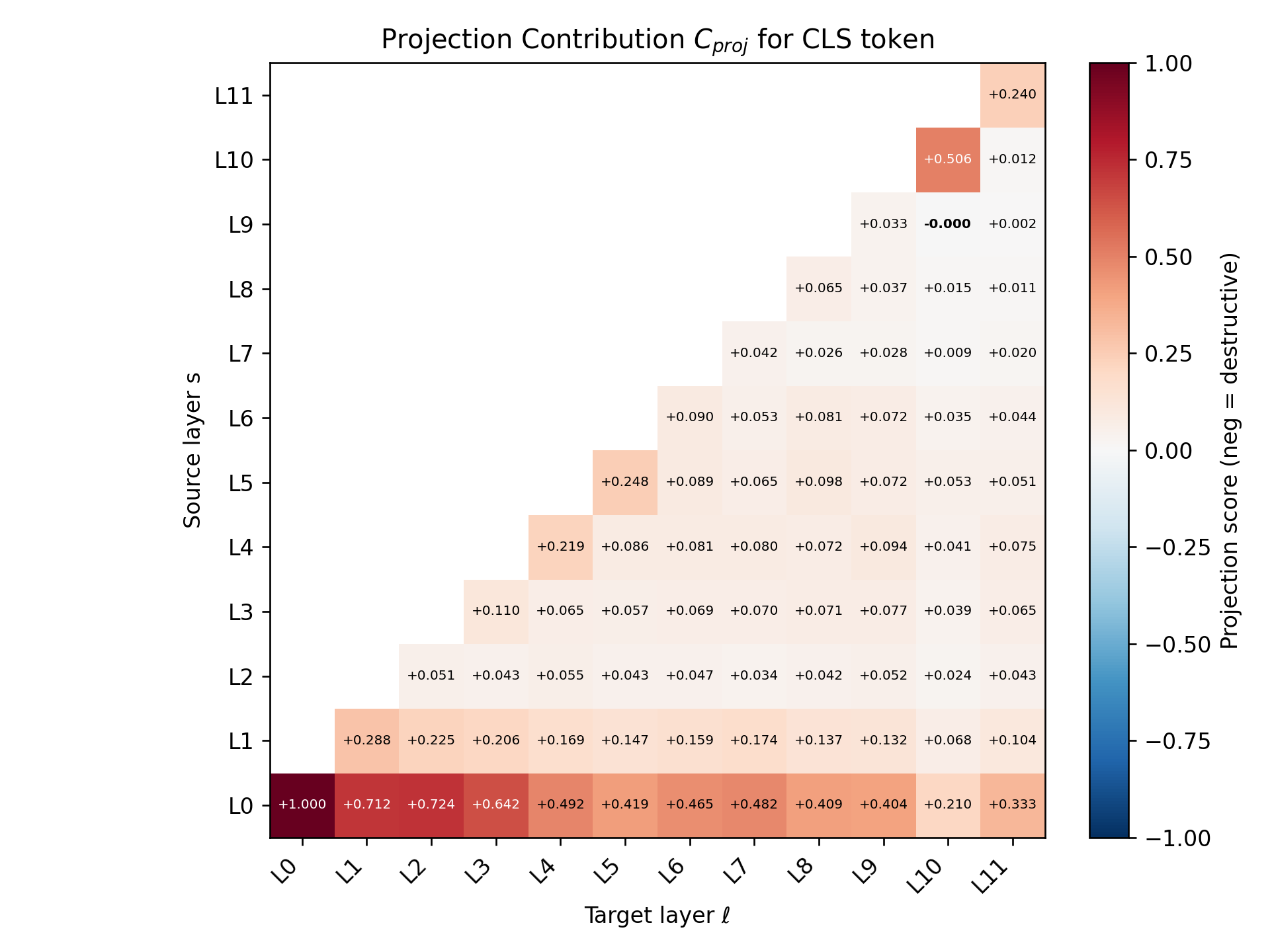}
    \caption{ReLU-Top-$k$, CLS}
    \label{fig:contrib-v16-cifar100-rtk-cls}
  \end{subfigure}
  \hfill
  \begin{subfigure}[t]{0.32\linewidth}
    \centering
    \includegraphics[width=\linewidth]{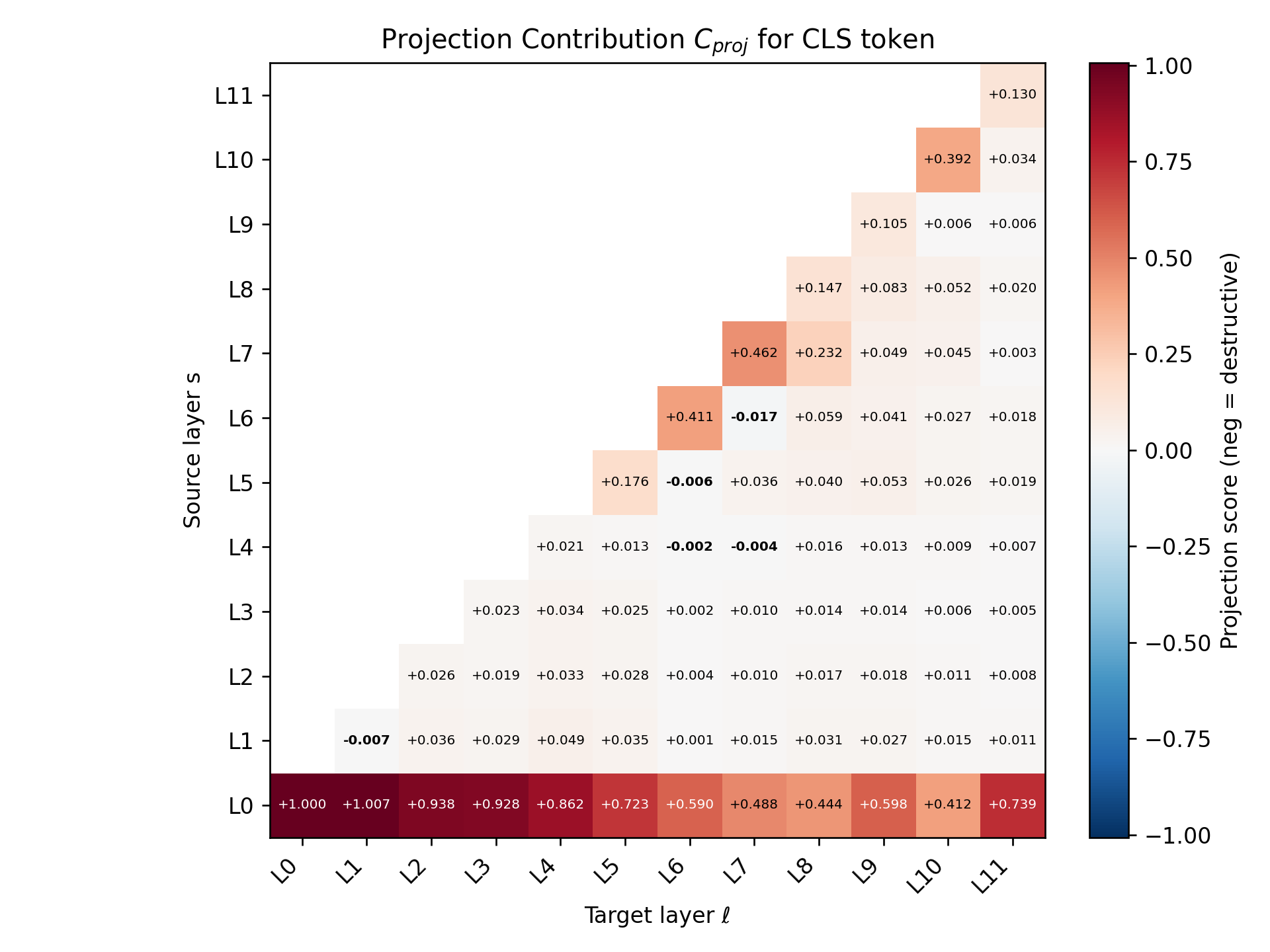}
    \caption{Abs-Top-$k$, CLS}
    \label{fig:contrib-v16-cifar100-atk-cls}
  \end{subfigure}

  \vspace{0.6em}

  \begin{subfigure}[t]{0.32\linewidth}
    \centering
    \includegraphics[width=\linewidth]{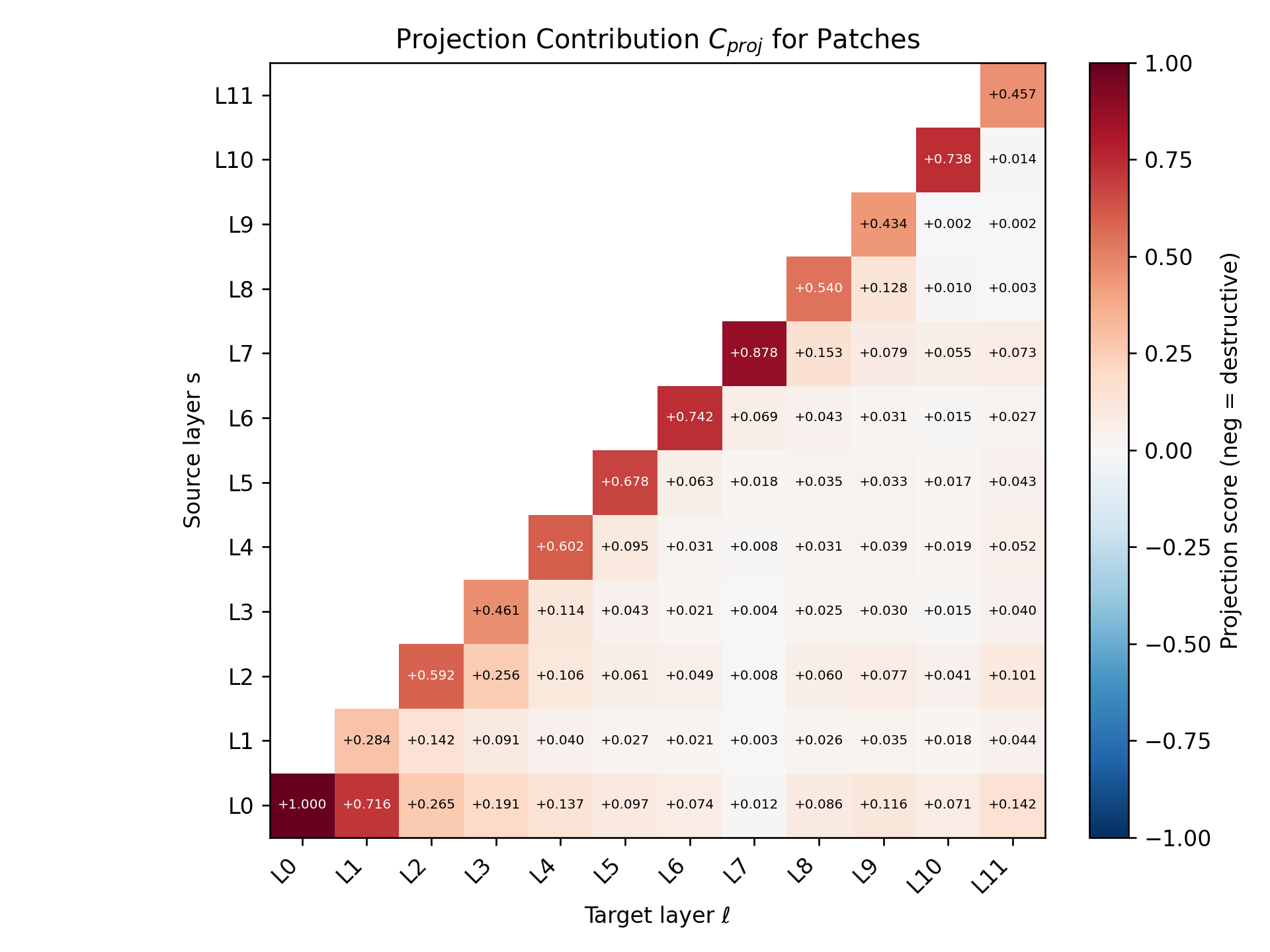}
    \caption{JumpReLU, patches}
    \label{fig:contrib-v16-cifar100-jr-patches}
  \end{subfigure}
  \hfill
  \begin{subfigure}[t]{0.32\linewidth}
    \centering
    \includegraphics[width=\linewidth]{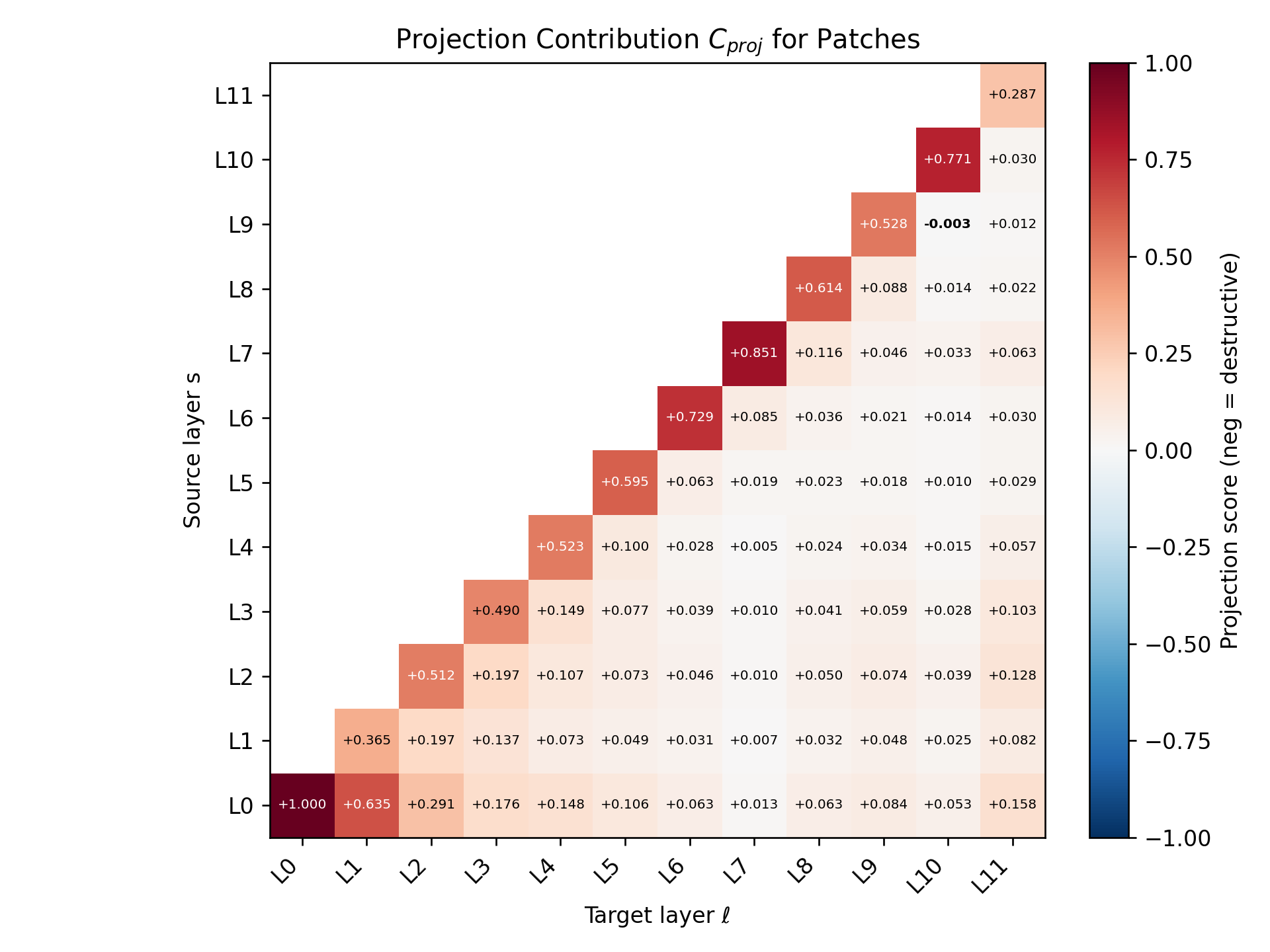}
    \caption{ReLU-Top-$k$, patches}
    \label{fig:contrib-v16-cifar100-rtk-patches}
  \end{subfigure}
  \hfill
  \begin{subfigure}[t]{0.32\linewidth}
    \centering
    \includegraphics[width=\linewidth]{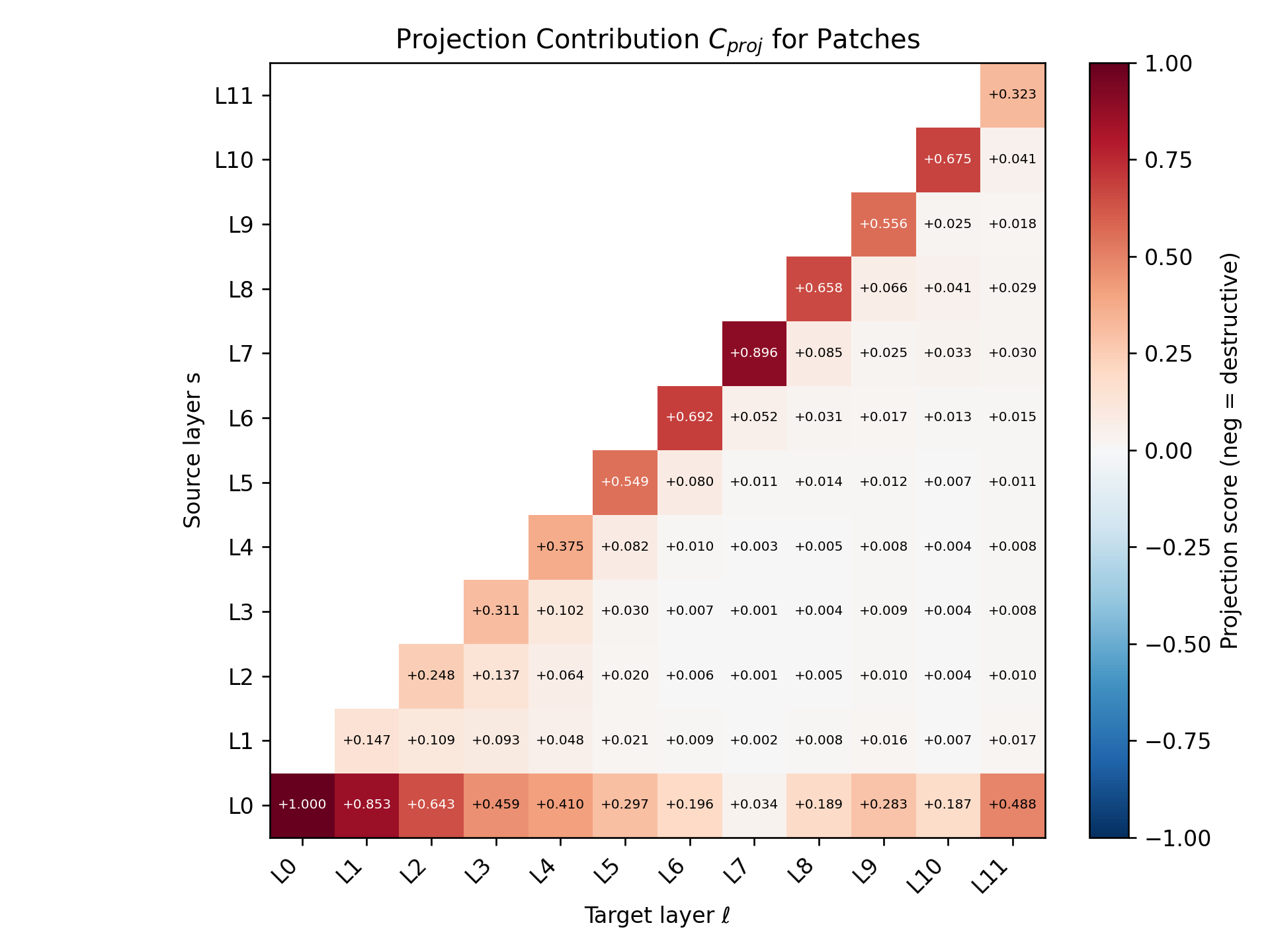}
    \caption{Abs-Top-$k$, patches}
    \label{fig:contrib-v16-cifar100-atk-patches}
  \end{subfigure}
\caption{
  Cross-layer contribution scores $C_{s \rightarrow \ell}$ on CIFAR-100
  with ViT-B/16. Columns vary the sparsifier (JumpReLU, ReLU-Top-$k$, Abs-Top-$k$)
  and rows show CLS (top) and patch tokens (bottom). Each heatmap visualizes the
  proportional contribution of source layer $s$ to the reconstructed activation at
  target layer $\ell$, averaged over the validation set.
  }
  \label{fig:contrib-v16-cifar100-grid}
\end{figure*}

\begin{figure*}[t]
  \centering
  \begin{subfigure}[t]{0.32\linewidth}
    \centering
    \includegraphics[width=\linewidth]{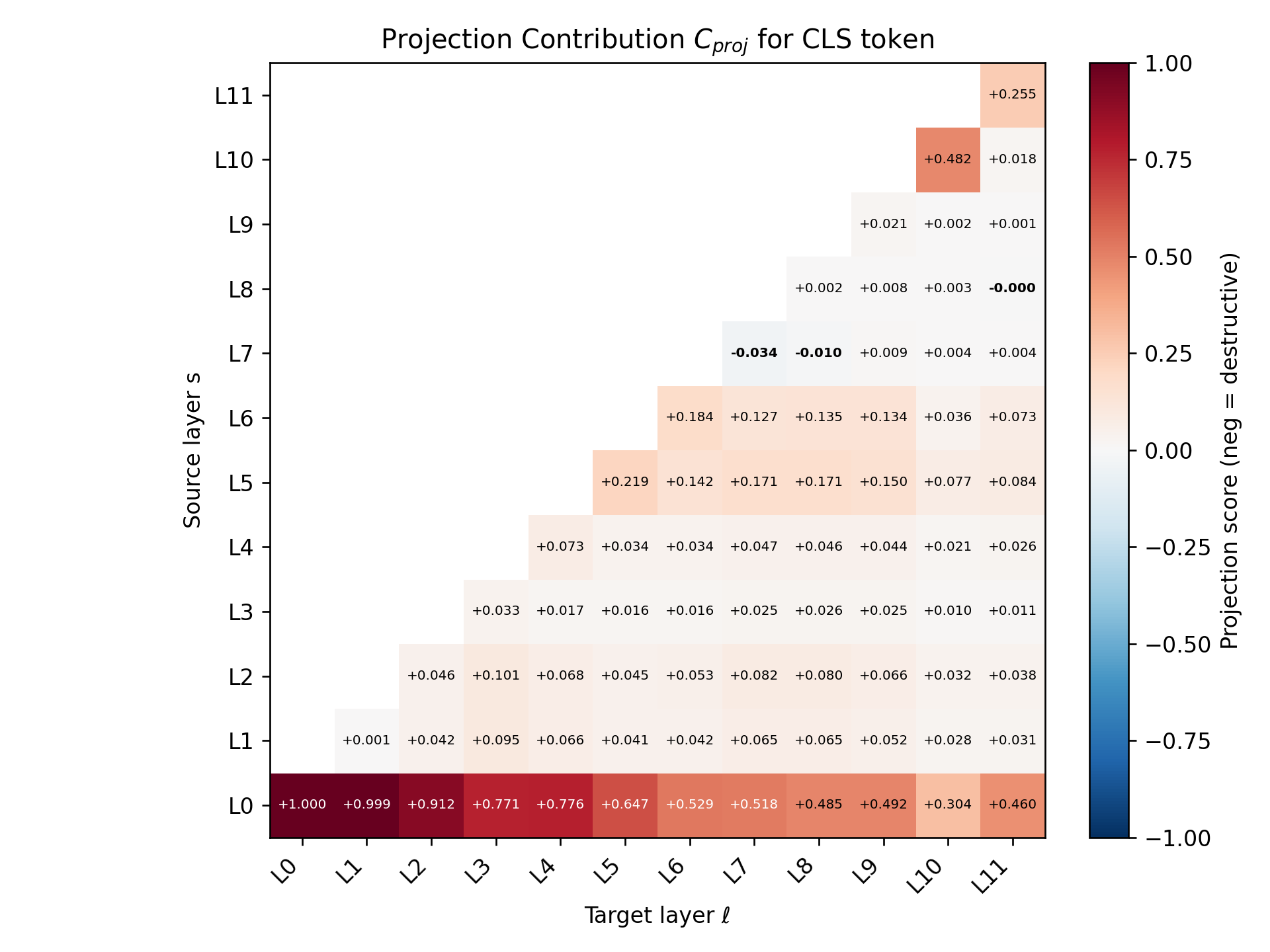}
    \caption{JumpReLU, CLS}
    \label{fig:contrib-v32-coco-jr-cls}
  \end{subfigure}
  \hfill
  \begin{subfigure}[t]{0.32\linewidth}
    \centering
    \includegraphics[width=\linewidth]{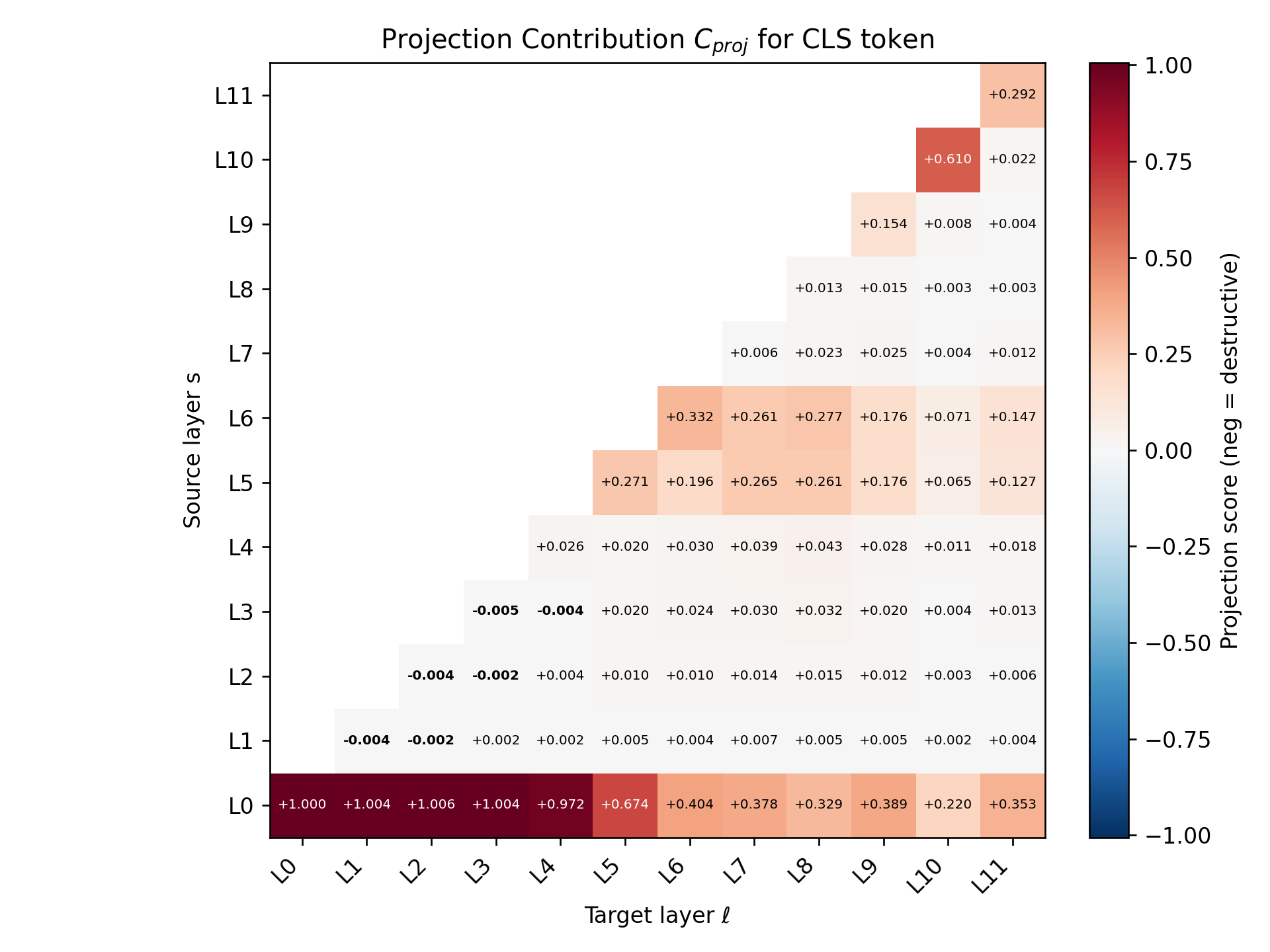}
    \caption{ReLU-Top-$k$, CLS}
    \label{fig:contrib-v32-coco-rtk-cls}
  \end{subfigure}
  \hfill
  \begin{subfigure}[t]{0.32\linewidth}
    \centering
    \includegraphics[width=\linewidth]{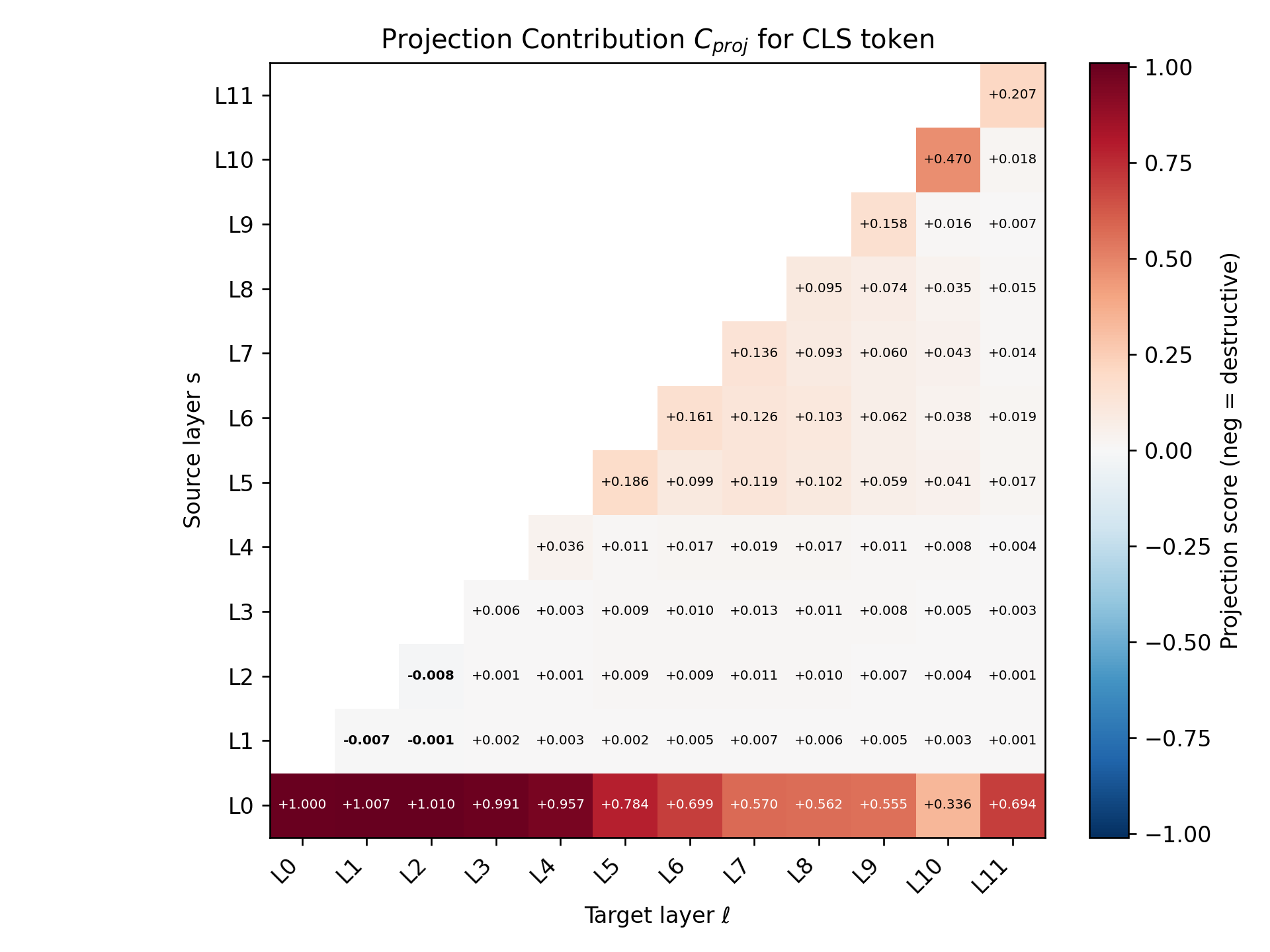}
    \caption{Abs-Top-$k$, CLS}
    \label{fig:contrib-v32-coco-atk-cls}
  \end{subfigure}

  \vspace{0.6em}

  \begin{subfigure}[t]{0.32\linewidth}
    \centering
    \includegraphics[width=\linewidth]{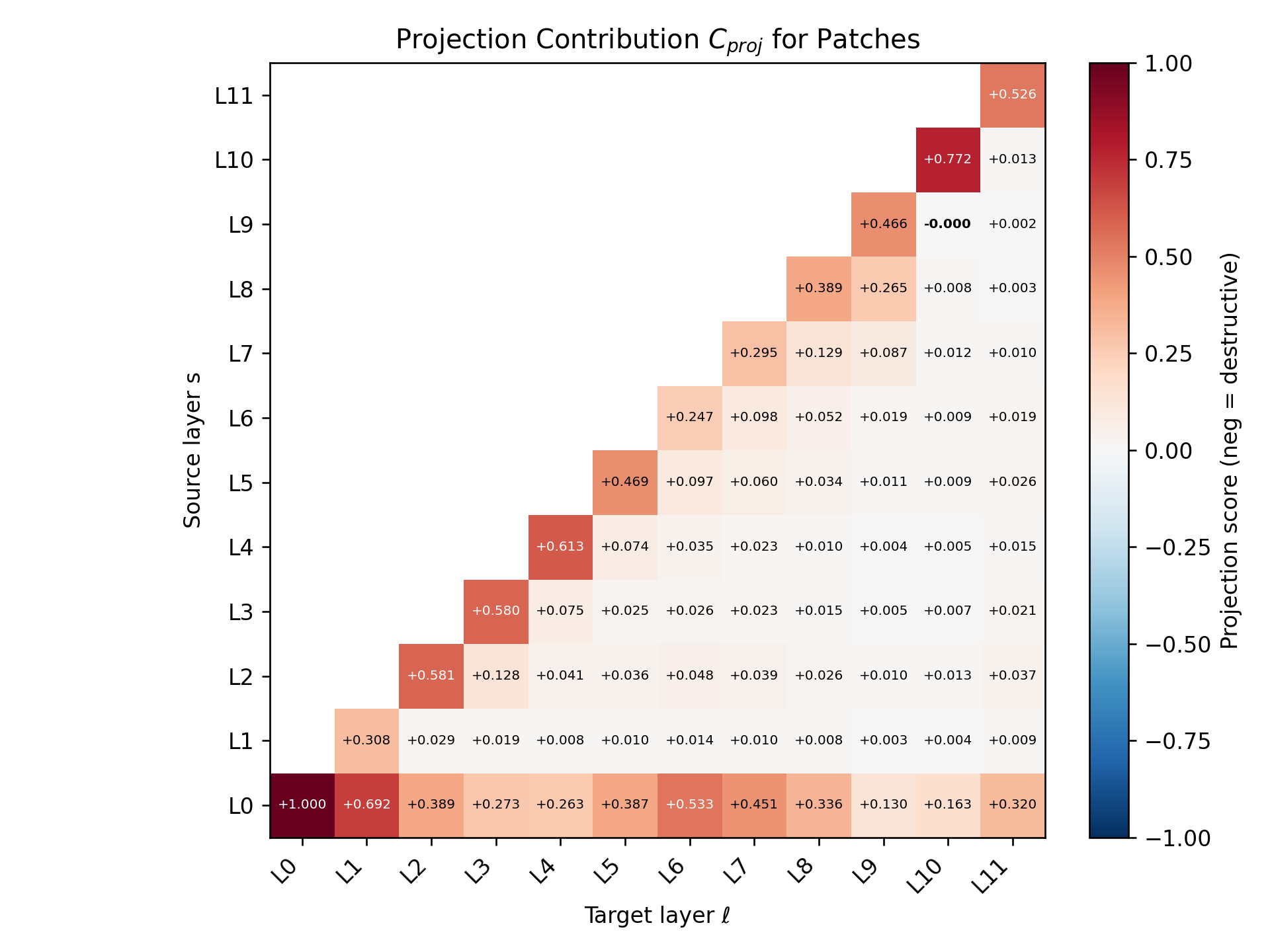}
    \caption{JumpReLU, patches}
    \label{fig:contrib-v32-coco-jr-patches}
  \end{subfigure}
  \hfill
  \begin{subfigure}[t]{0.32\linewidth}
    \centering
    \includegraphics[width=\linewidth]{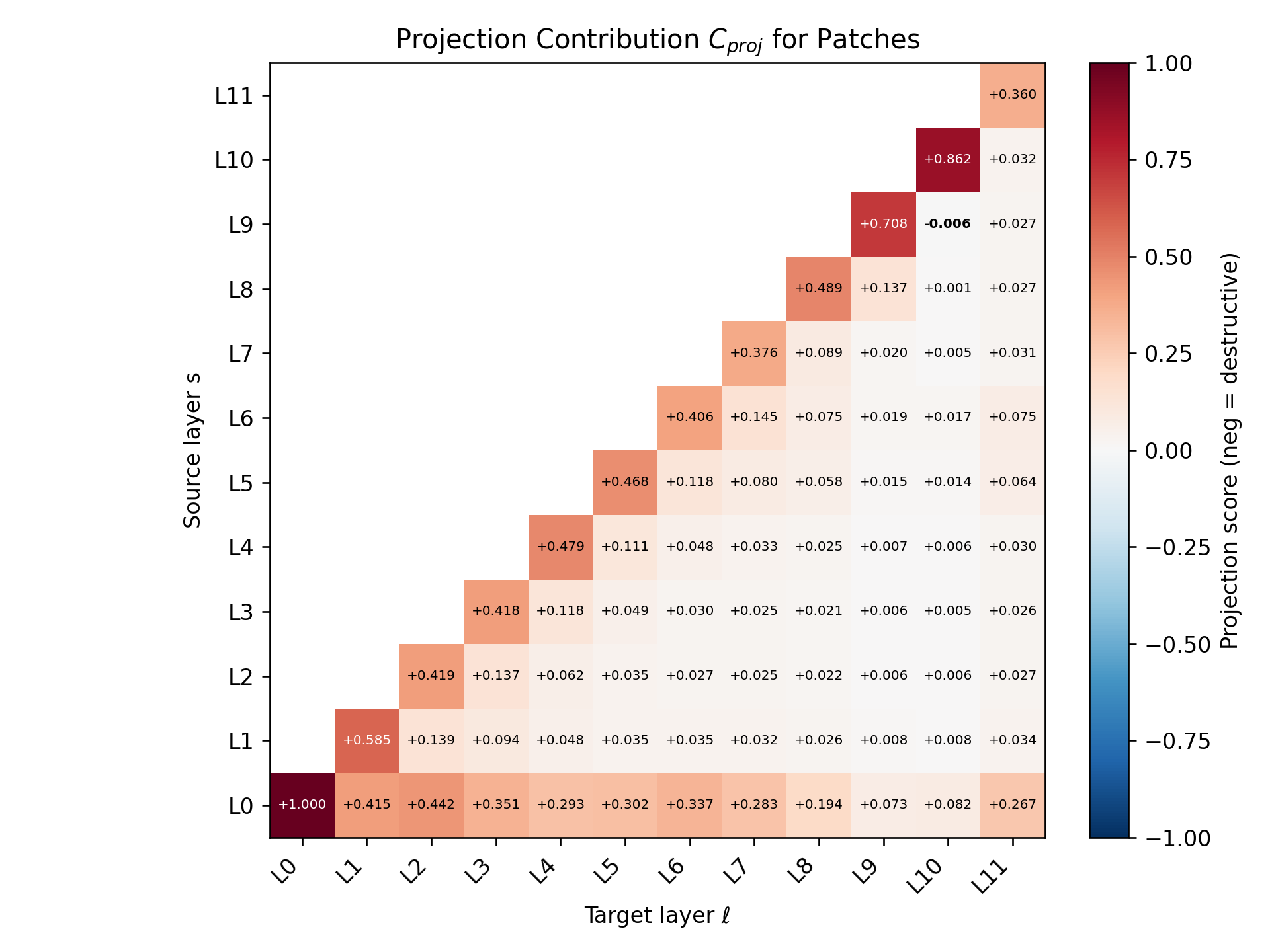}
    \caption{ReLU-Top-$k$, patches}
    \label{fig:contrib-v32-coco-rtk-patches}
  \end{subfigure}
  \hfill
  \begin{subfigure}[t]{0.32\linewidth}
    \centering
    \includegraphics[width=\linewidth]{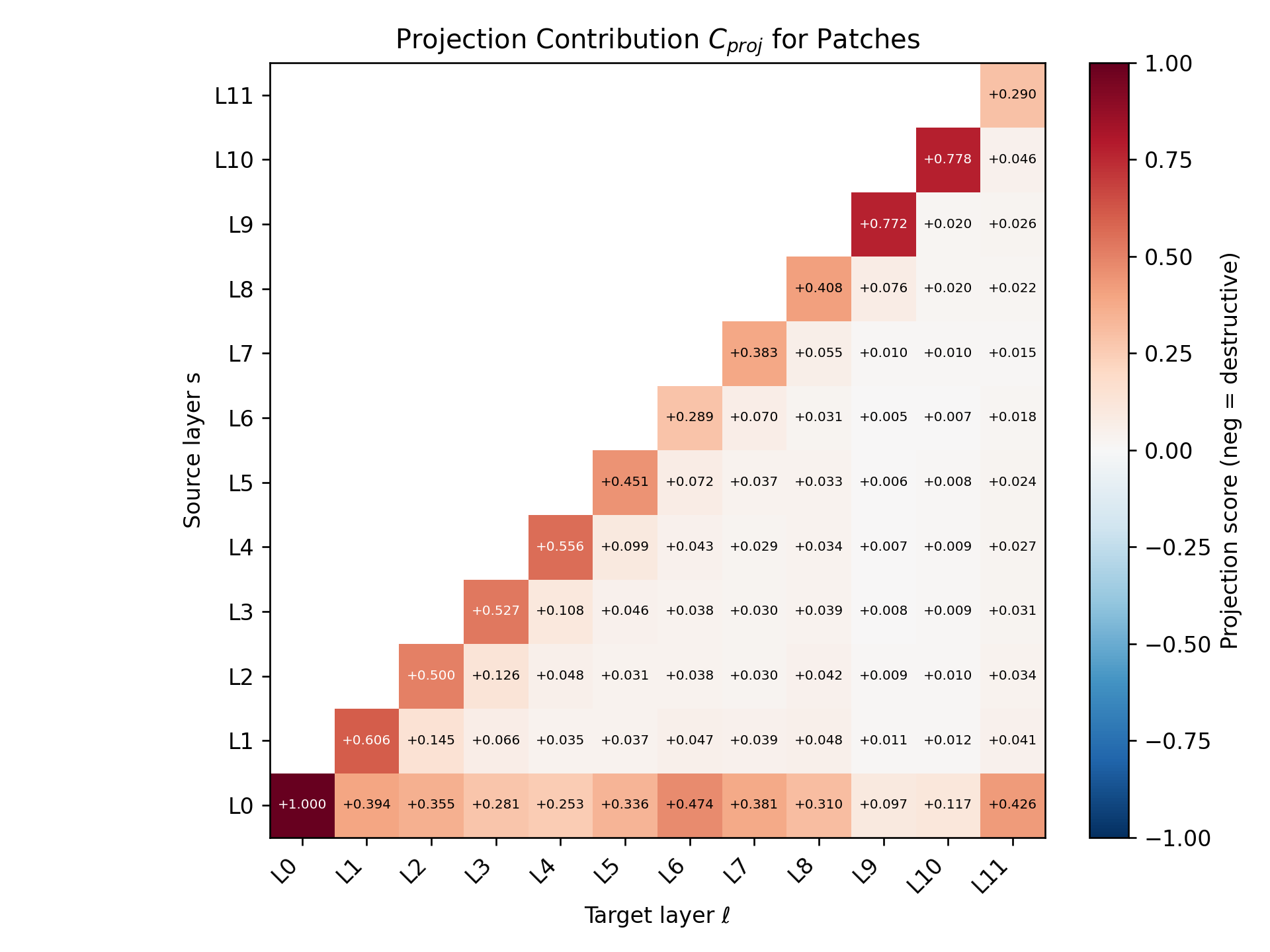}
    \caption{Abs-Top-$k$, patches}
    \label{fig:contrib-v32-coco-atk-patches}
  \end{subfigure}
\caption{
  Cross-layer contribution scores $C_{s \rightarrow \ell}$ on COCO
  with ViT-B/32. Columns vary the sparsifier (JumpReLU, ReLU-Top-$k$, Abs-Top-$k$)
  and rows show CLS (top) and patch tokens (bottom). Each heatmap visualizes the
  proportional contribution of source layer $s$ to the reconstructed activation at
  target layer $\ell$, averaged over the validation set.
  }
  \label{fig:contrib-v32-coco-grid}
\end{figure*}

\begin{figure*}[t]
  \centering
  \begin{subfigure}[t]{0.32\linewidth}
    \centering
    \includegraphics[width=\linewidth]{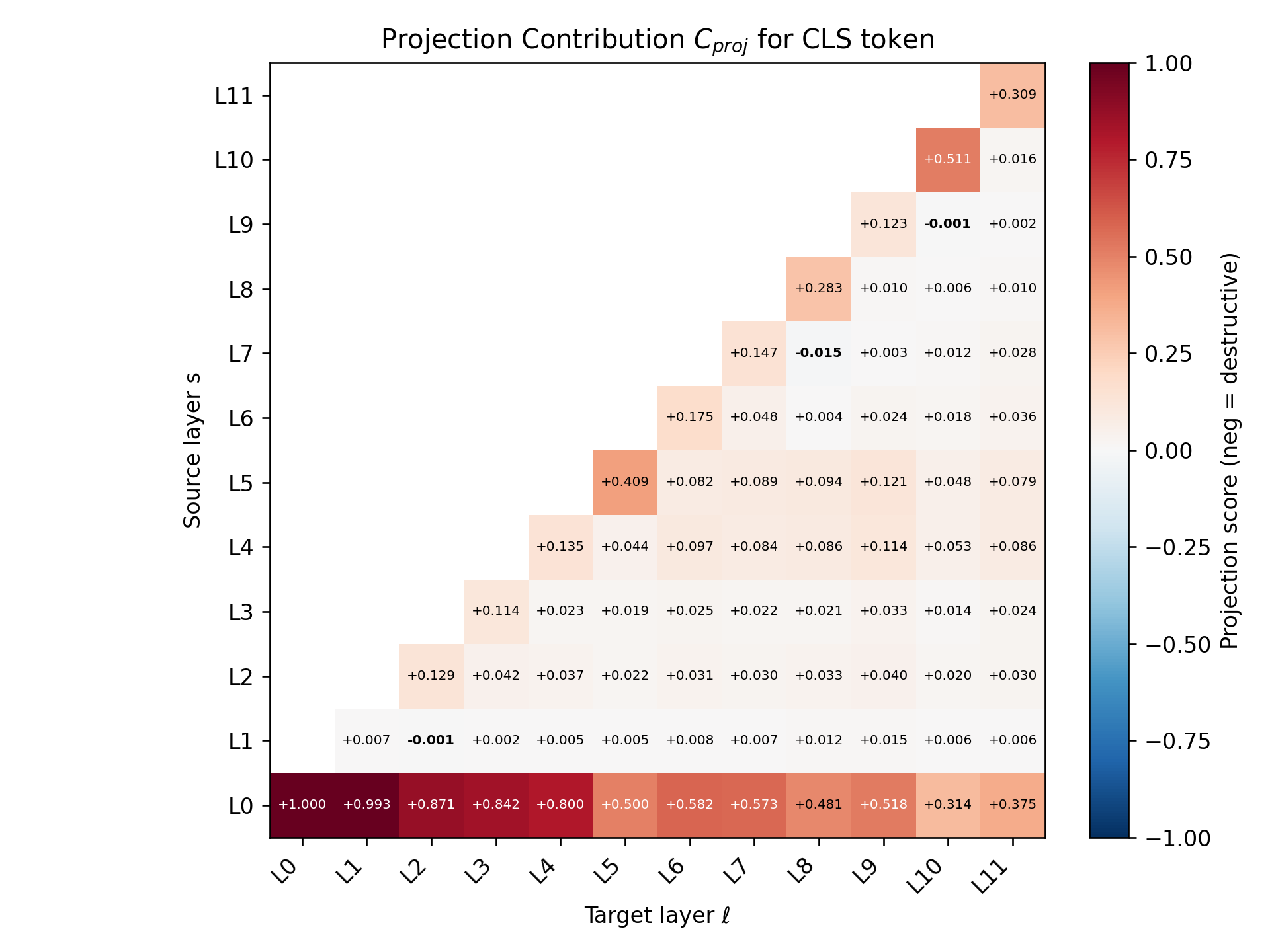}
    \caption{JumpReLU, CLS}
    \label{fig:contrib-v16-coco-jr-cls}
  \end{subfigure}
  \hfill
  \begin{subfigure}[t]{0.32\linewidth}
    \centering
    \includegraphics[width=\linewidth]{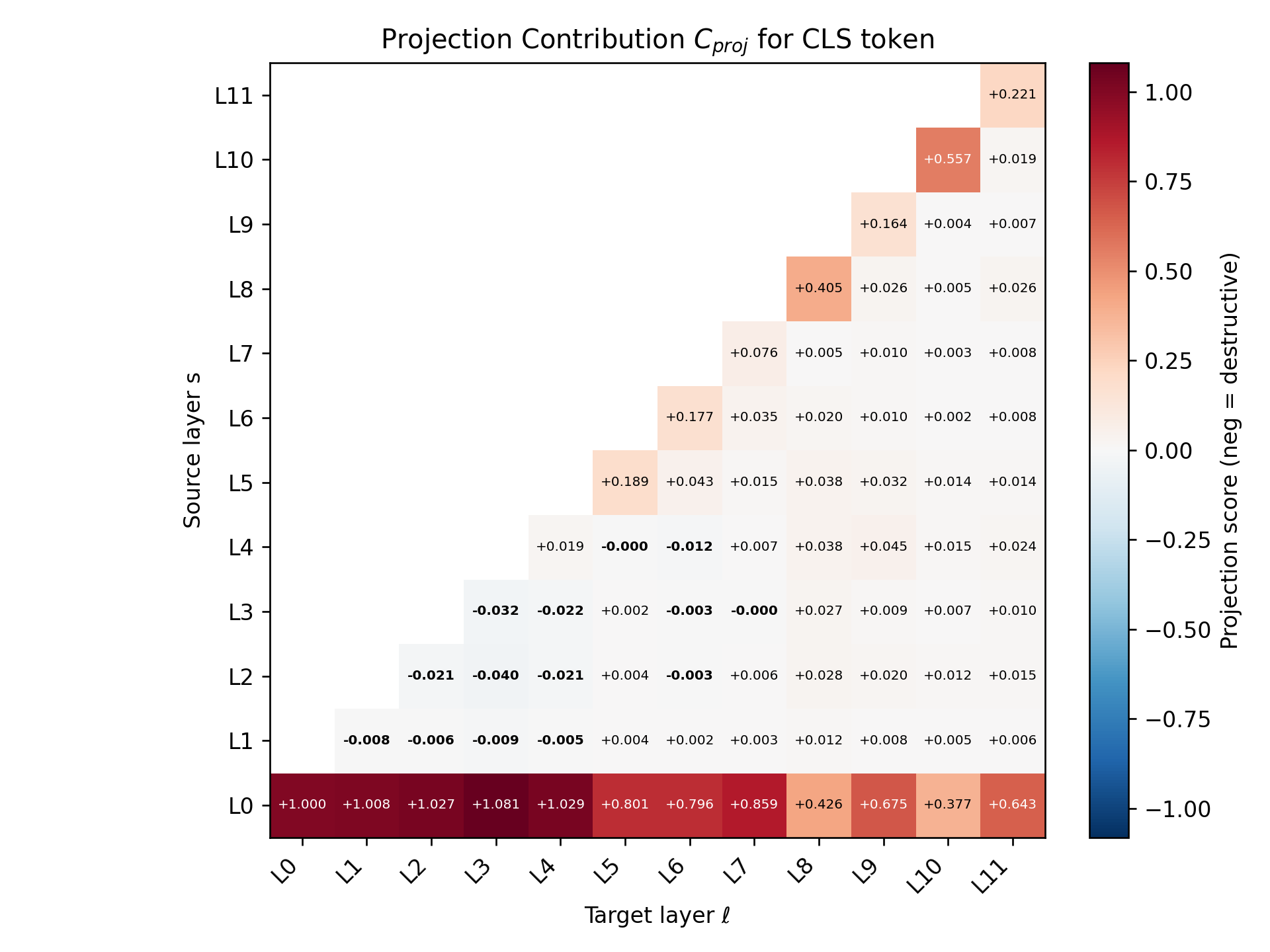}
    \caption{ReLU-Top-$k$, CLS}
    \label{fig:contrib-v16-coco-rtk-cls}
  \end{subfigure}
  \hfill
  \begin{subfigure}[t]{0.32\linewidth}
    \centering
    \includegraphics[width=\linewidth]{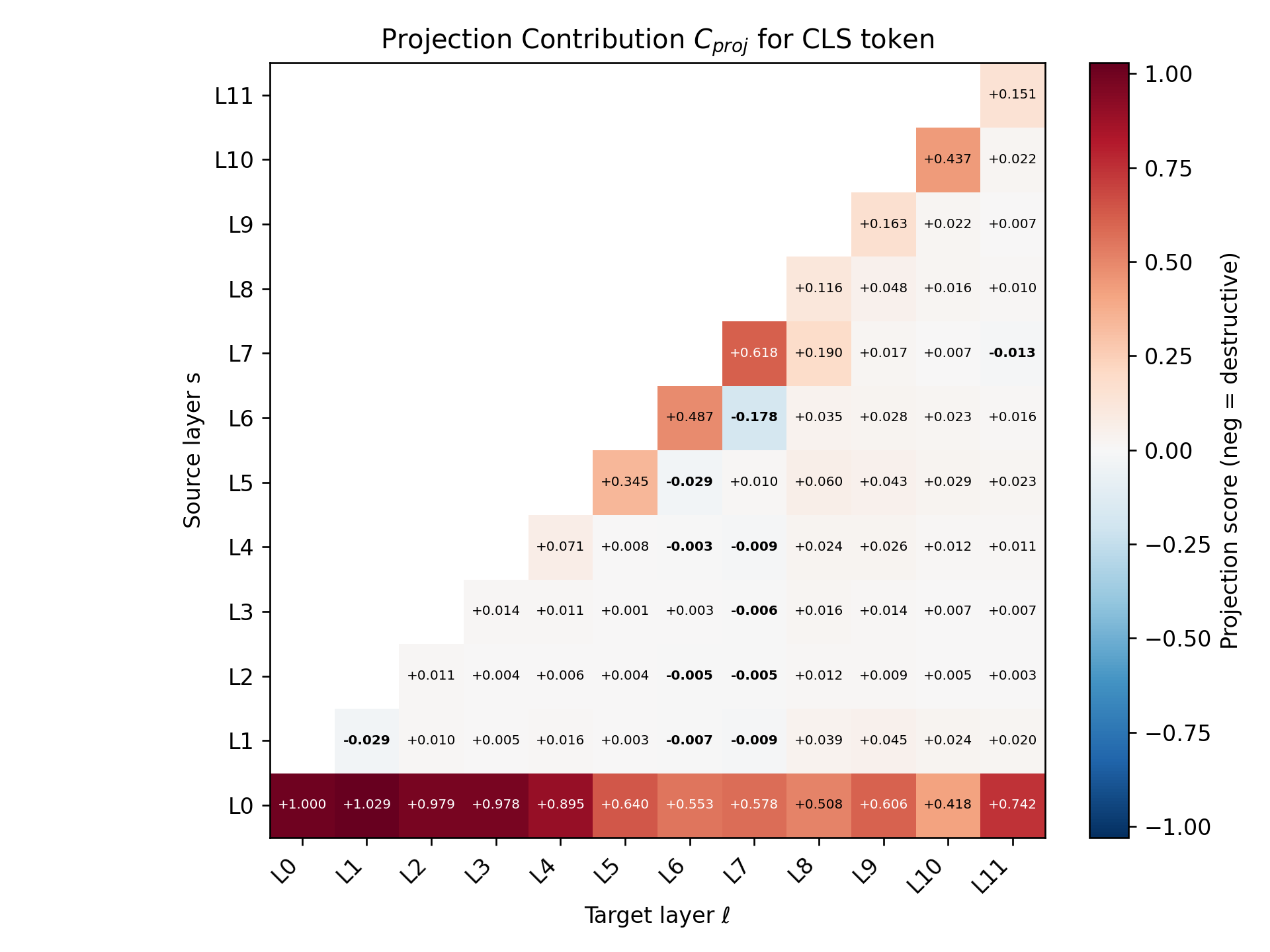}
    \caption{Abs-Top-$k$, CLS}
    \label{fig:contrib-v16-coco-atk-cls}
  \end{subfigure}

  \vspace{0.6em}

  \begin{subfigure}[t]{0.32\linewidth}
    \centering
    \includegraphics[width=\linewidth]{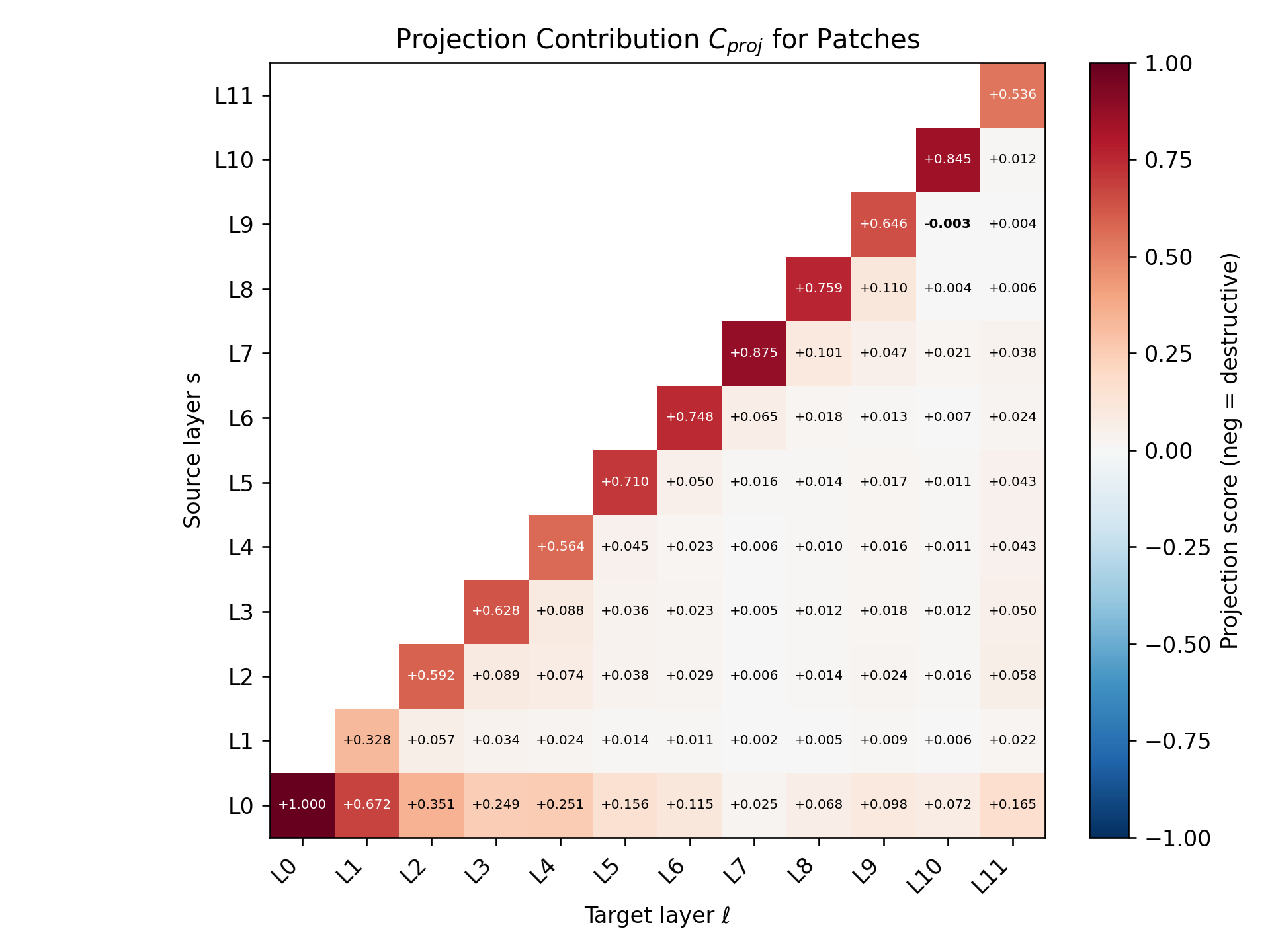}
    \caption{JumpReLU, patches}
    \label{fig:contrib-v16-coco-jr-patches}
  \end{subfigure}
  \hfill
  \begin{subfigure}[t]{0.32\linewidth}
    \centering
    \includegraphics[width=\linewidth]{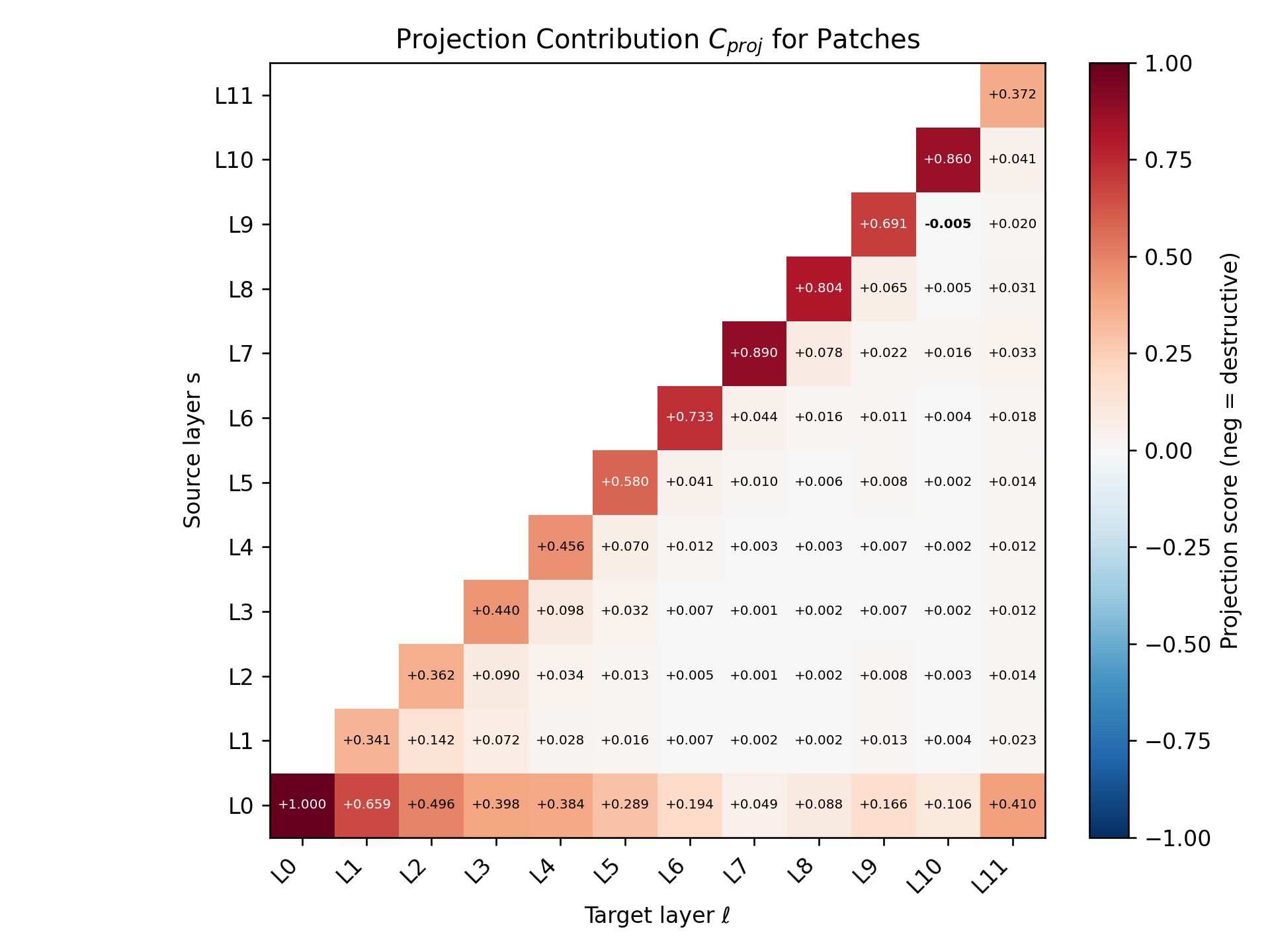}
    \caption{ReLU-Top-$k$, patches}
    \label{fig:contrib-v16-coco-rtk-patches}
  \end{subfigure}
  \hfill
  \begin{subfigure}[t]{0.32\linewidth}
    \centering
    \includegraphics[width=\linewidth]{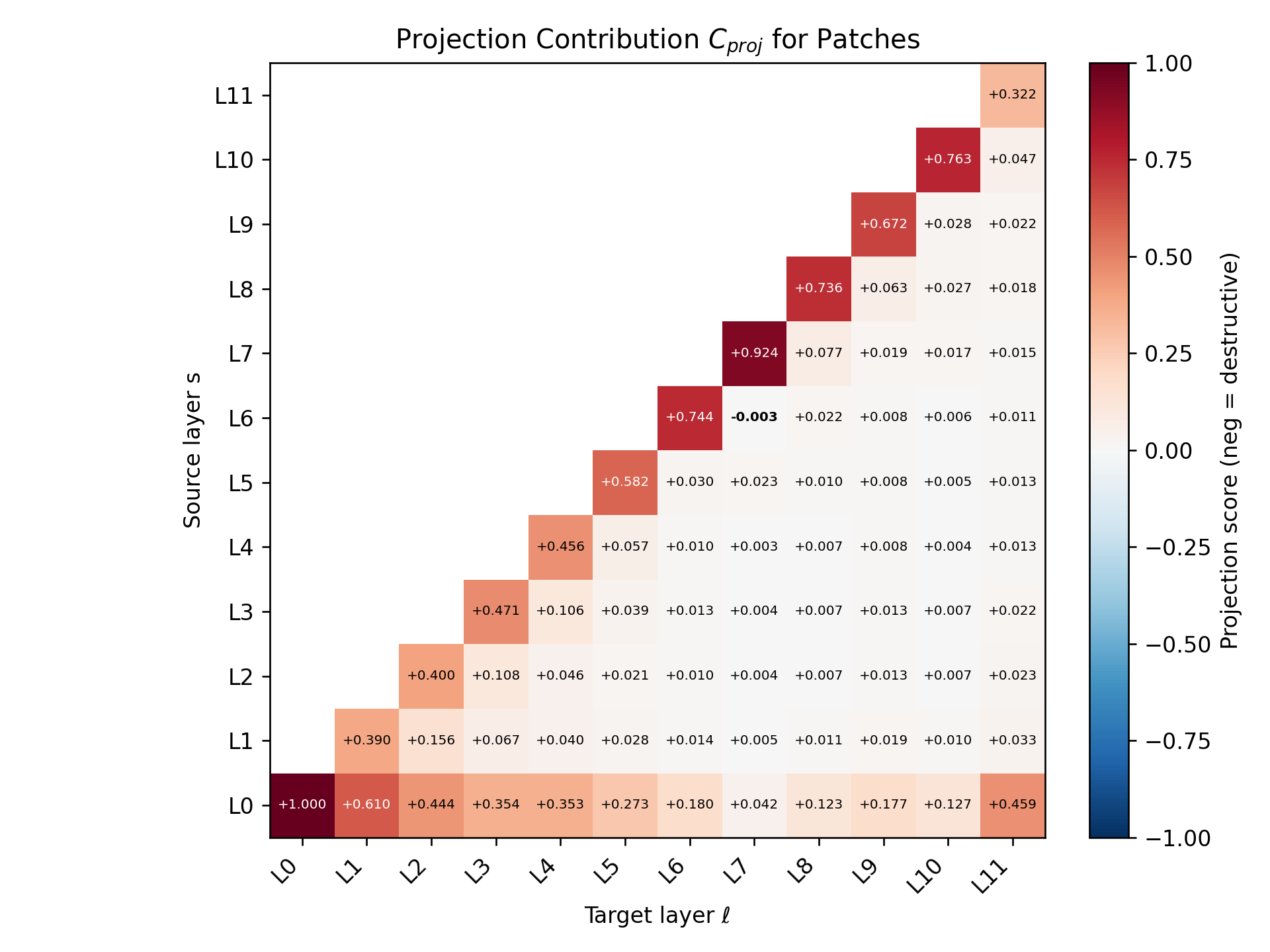}
    \caption{Abs-Top-$k$, patches}
    \label{fig:contrib-v16-coco-atk-patches}
  \end{subfigure}
\caption{
  Cross-layer contribution scores $C_{s \rightarrow \ell}$ on COCO
  with ViT-B/16. Columns vary the sparsifier (JumpReLU, ReLU-Top-$k$, Abs-Top-$k$)
  and rows show CLS (top) and patch tokens (bottom). Each heatmap visualizes the
  proportional contribution of source layer $s$ to the reconstructed activation at
  target layer $\ell$, averaged over the validation set.
  }
  \label{fig:contrib-v16-coco-grid}
\end{figure*}

\begin{figure*}[t]
  \centering
  \begin{subfigure}[t]{0.32\linewidth}
    \centering
    \includegraphics[width=\linewidth]{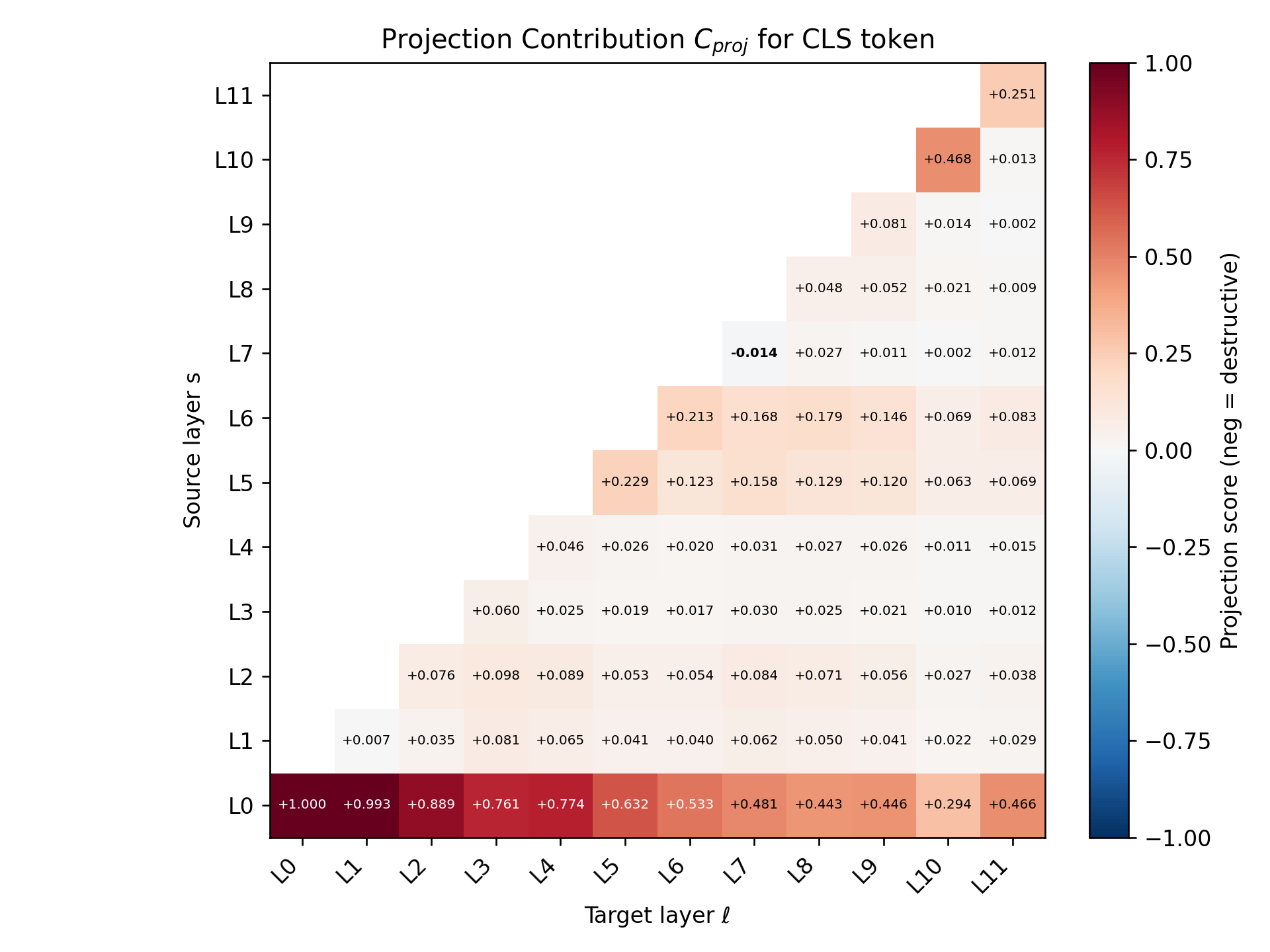}
    \caption{JumpReLU, CLS}
    \label{fig:contrib-v32-imnet-jr-cls}
  \end{subfigure}
  \hfill
  \begin{subfigure}[t]{0.32\linewidth}
    \centering
    \includegraphics[width=\linewidth]{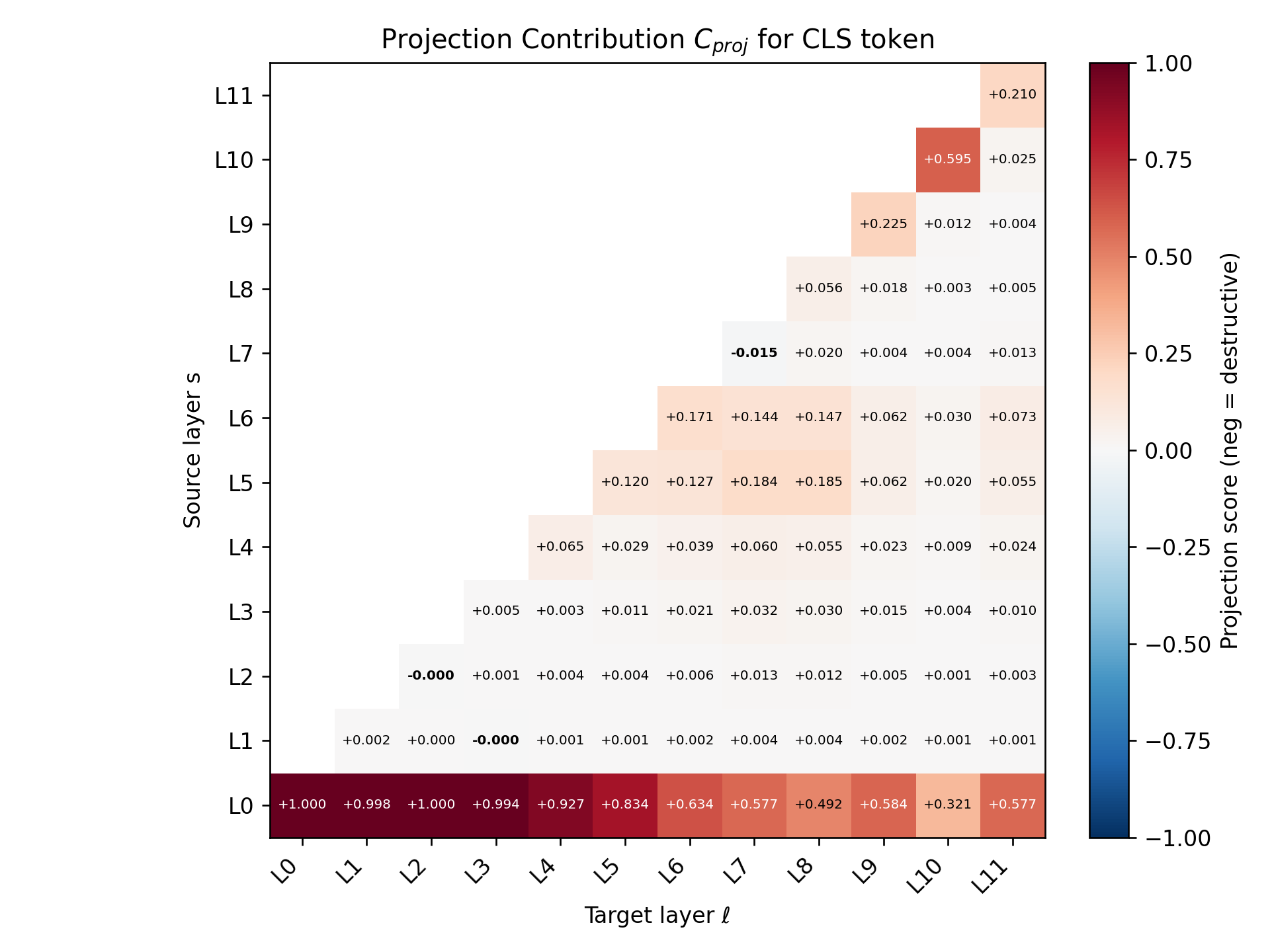}
    \caption{ReLU-Top-$k$, CLS}
    \label{fig:contrib-v32-imnet-rtk-cls}
  \end{subfigure}
  \hfill
  \begin{subfigure}[t]{0.32\linewidth}
    \centering
    \includegraphics[width=\linewidth]{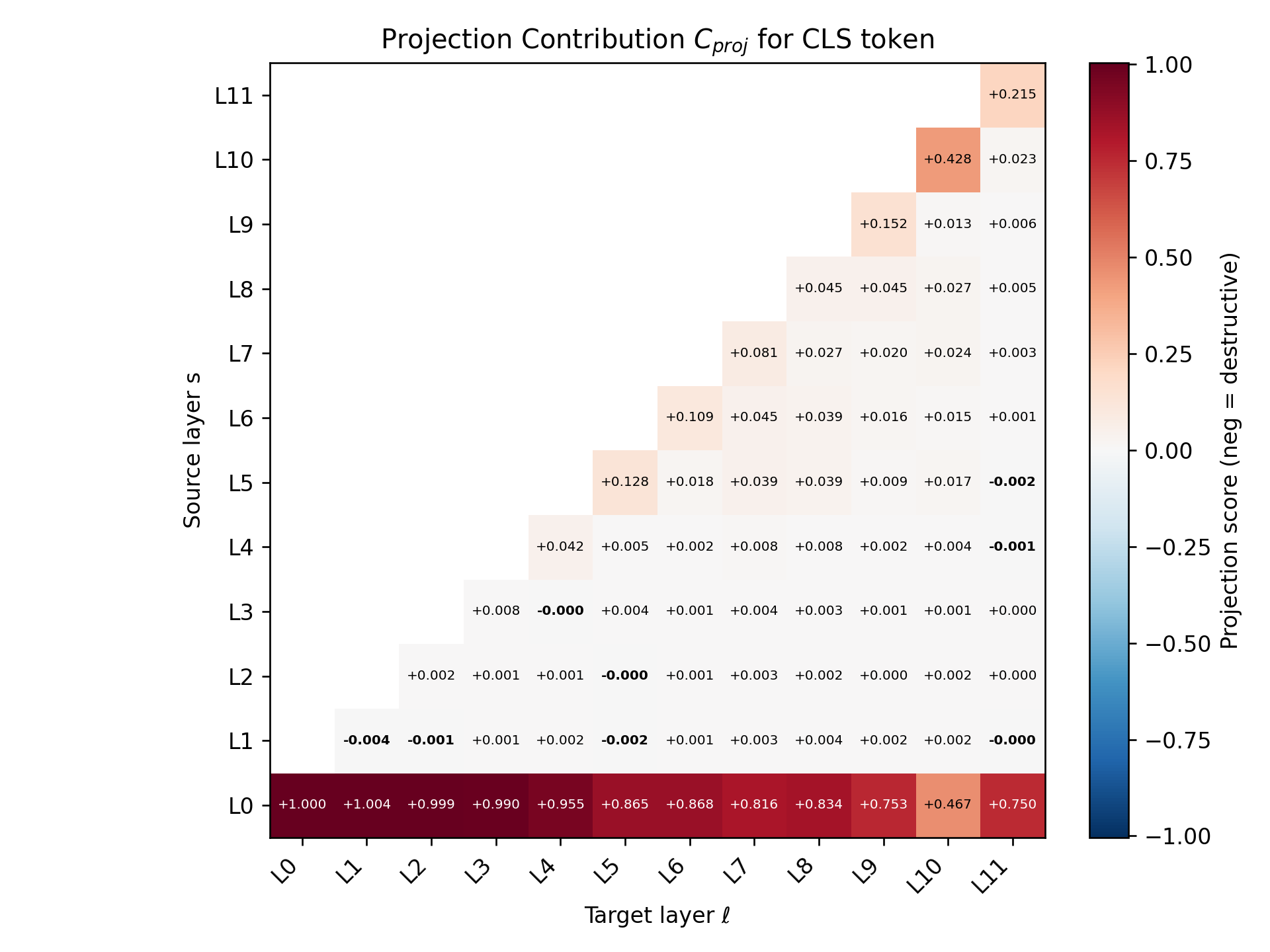}
    \caption{Abs-Top-$k$, CLS}
    \label{fig:contrib-v32-imnet-atk-cls}
  \end{subfigure}

  \vspace{0.6em}

  \begin{subfigure}[t]{0.32\linewidth}
    \centering
    \includegraphics[width=\linewidth]{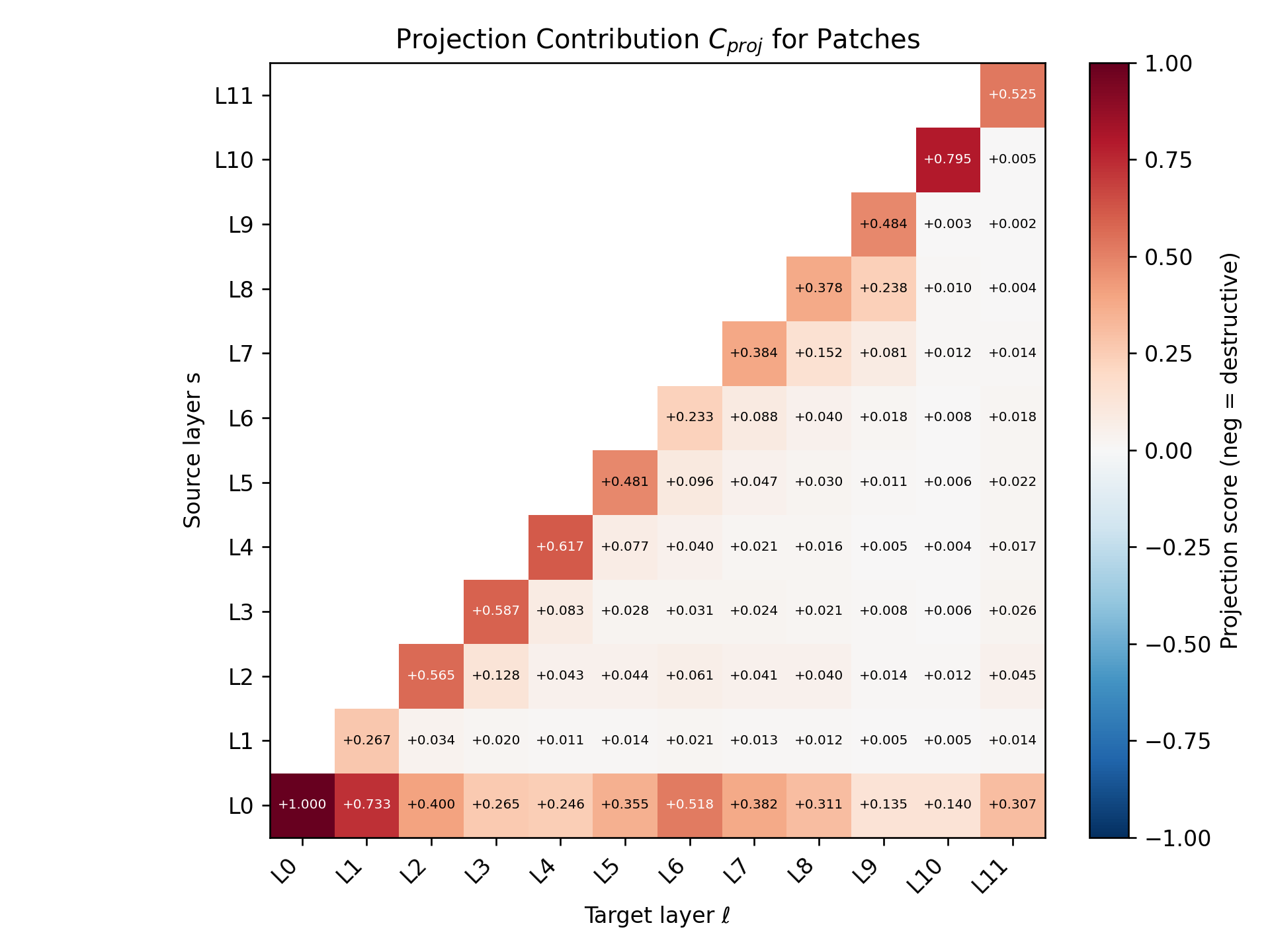}
    \caption{JumpReLU, patches}
    \label{fig:contrib-v32-imnet-jr-patches}
  \end{subfigure}
  \hfill
  \begin{subfigure}[t]{0.32\linewidth}
    \centering
    \includegraphics[width=\linewidth]{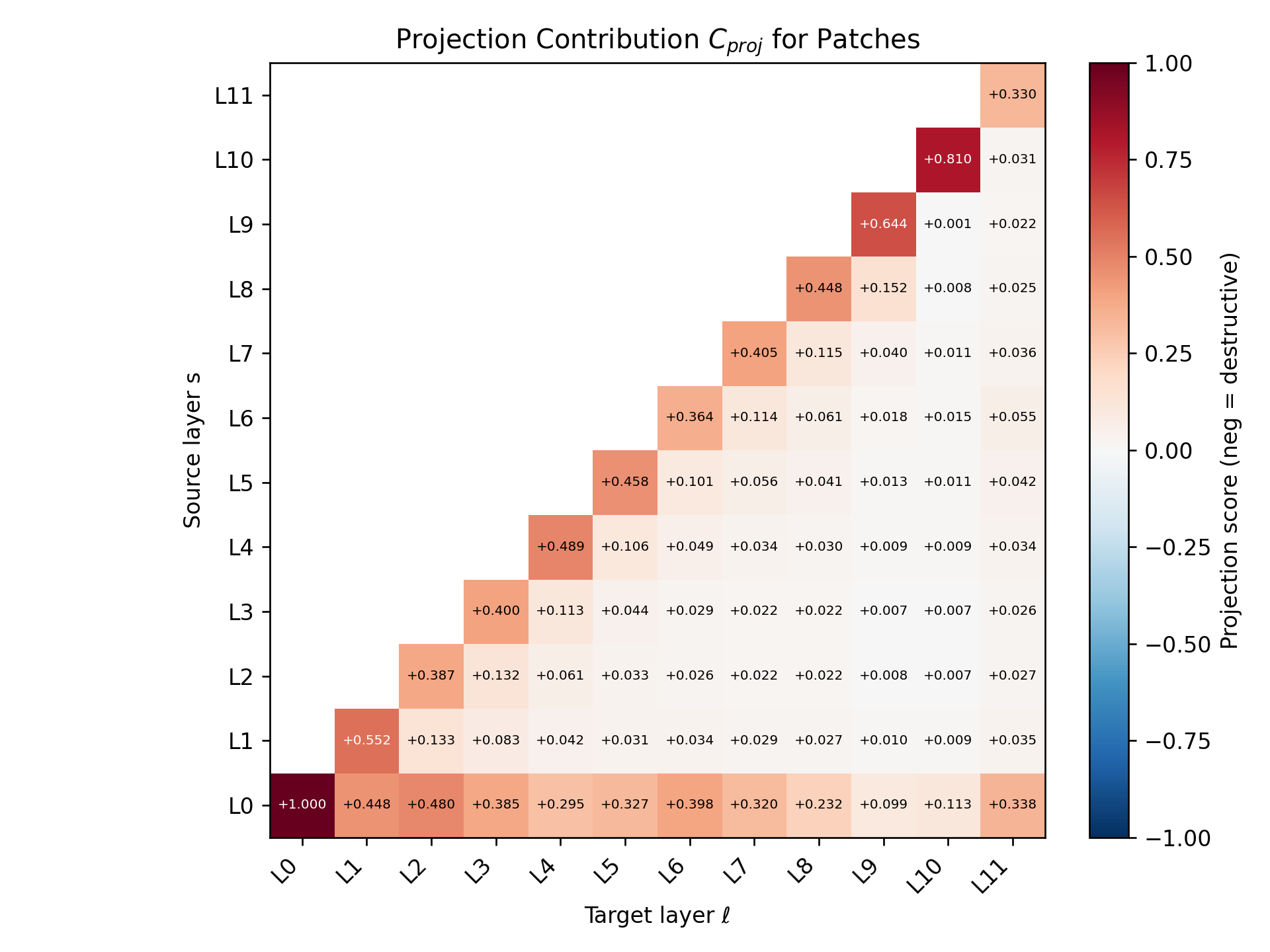}
    \caption{ReLU-Top-$k$, patches}
    \label{fig:contrib-v32-imnet-rtk-patches}
  \end{subfigure}
  \hfill
  \begin{subfigure}[t]{0.32\linewidth}
    \centering
    \includegraphics[width=\linewidth]{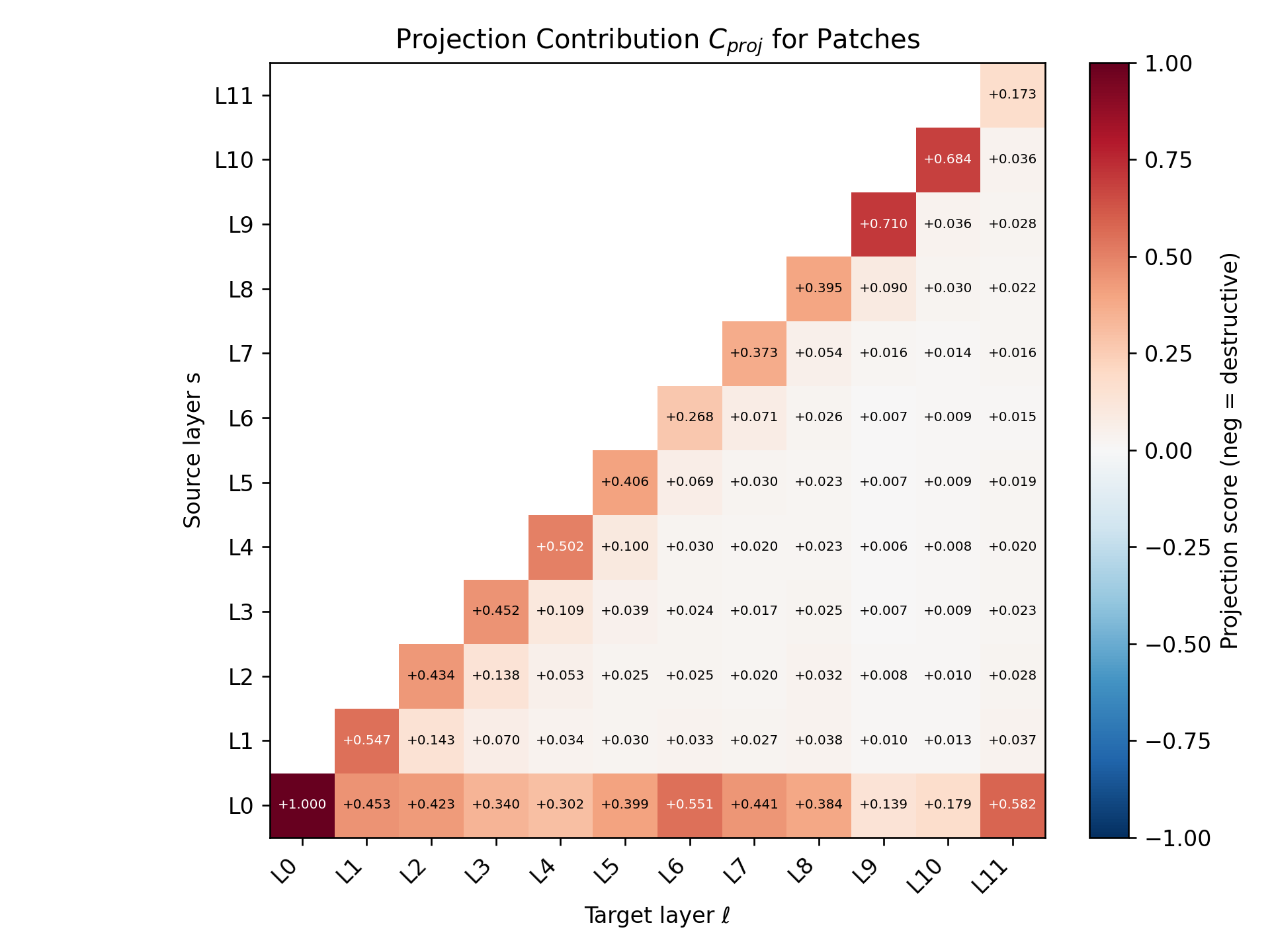}
    \caption{Abs-Top-$k$, patches}
    \label{fig:contrib-v32-imnet-atk-patches}
  \end{subfigure}
\caption{
  Cross-layer contribution scores $C_{s \rightarrow \ell}$ on ImageNet-100 with ViT-B/32. Columns vary the sparsifier (JumpReLU, ReLU-Top-$k$, Abs-Top-$k$)
  and rows show CLS (top) and patch tokens (bottom). Each heatmap visualizes the
  proportional contribution of source layer $s$ to the reconstructed activation at
  target layer $\ell$, averaged over the validation set.
  }
  \label{fig:contrib-v32-imnet-grid}
\end{figure*}

\begin{figure*}[t]
  \centering
  \begin{subfigure}[t]{0.32\linewidth}
    \centering
    \includegraphics[width=\linewidth]{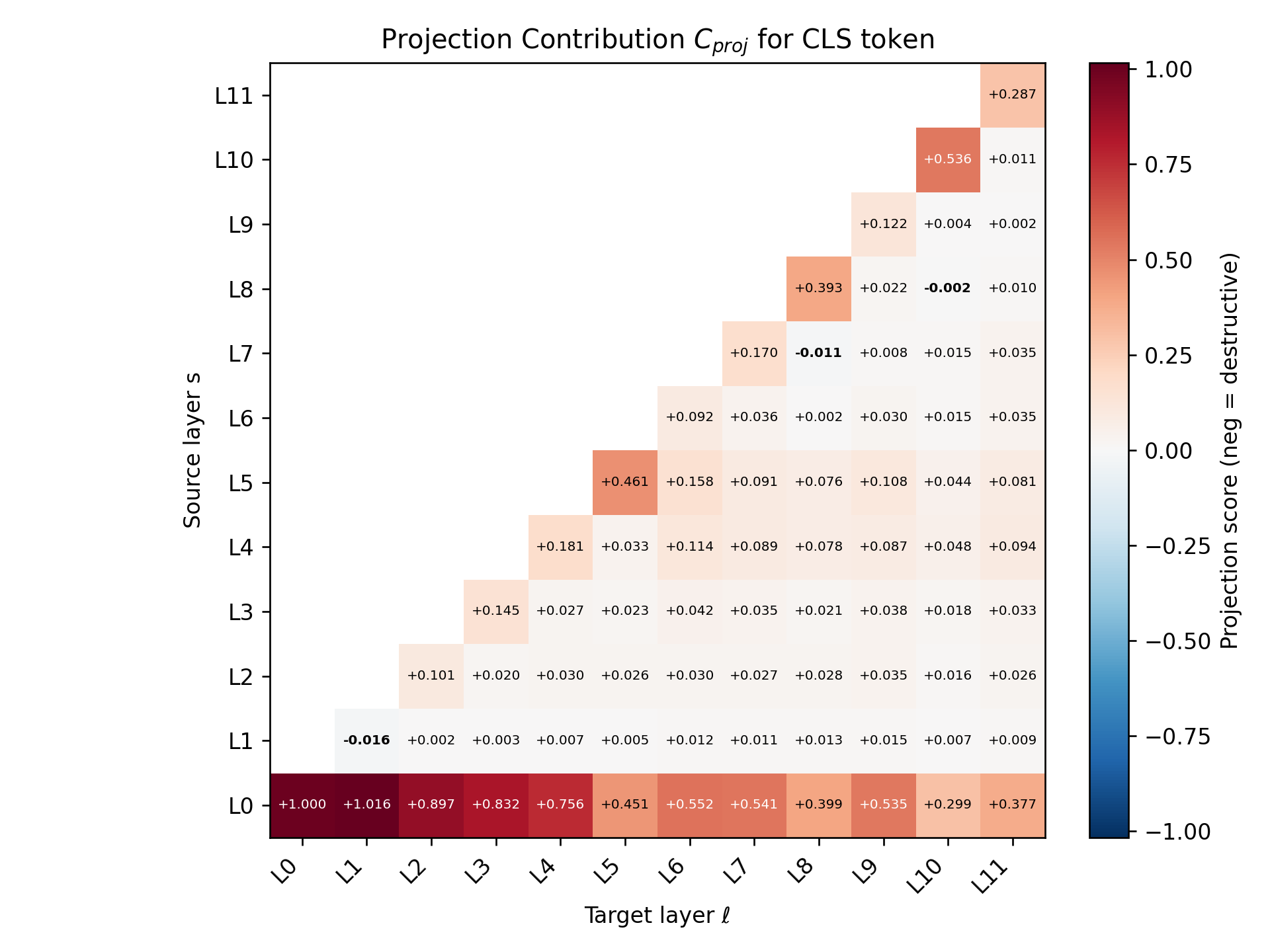}
    \caption{JumpReLU, CLS}
    \label{fig:contrib-v16-imnet-jr-cls}
  \end{subfigure}
  \hfill
  \begin{subfigure}[t]{0.32\linewidth}
    \centering
    \includegraphics[width=\linewidth]{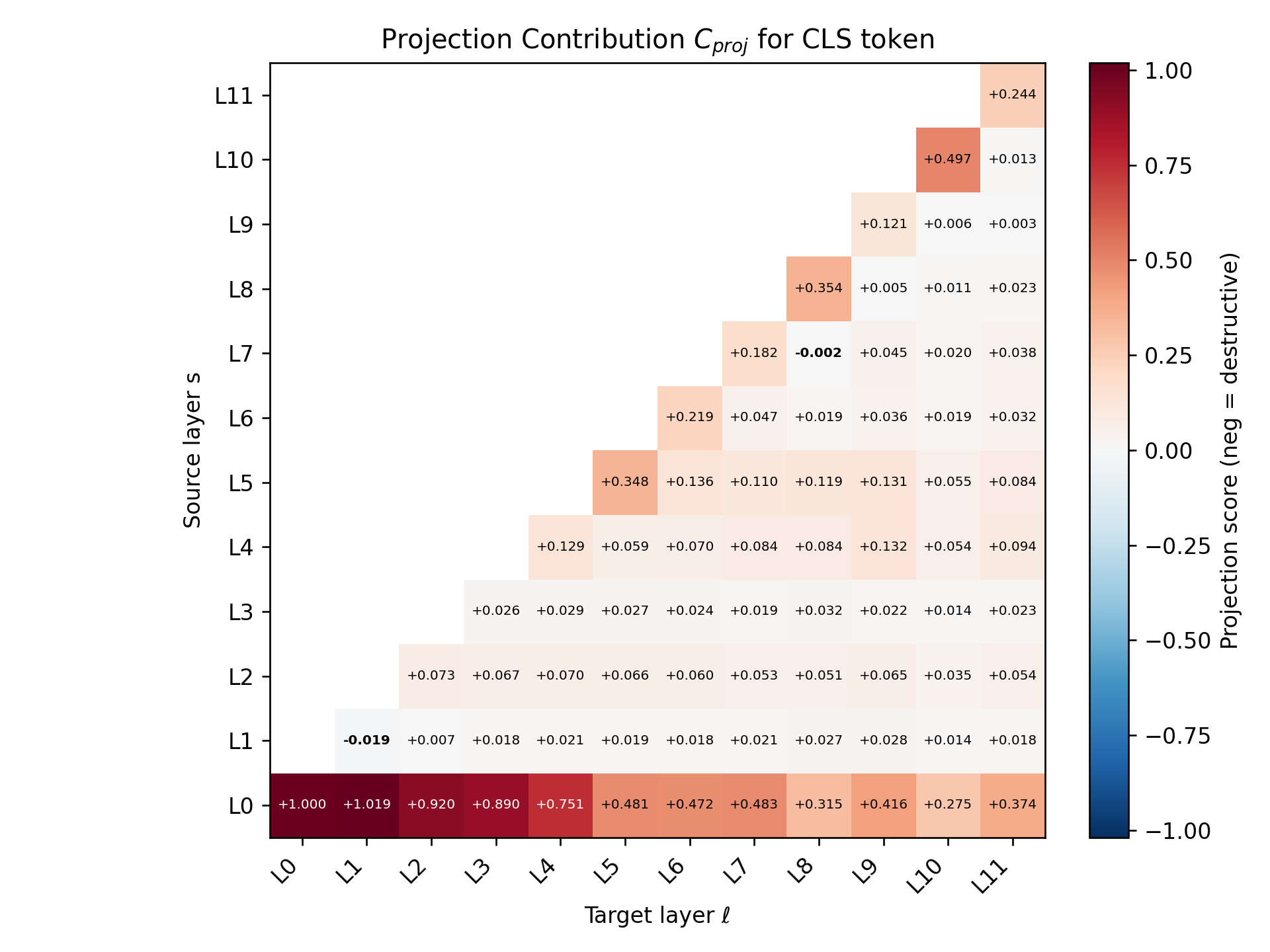}
    \caption{ReLU-Top-$k$, CLS}
    \label{fig:contrib-v16-imnet-rtk-cls}
  \end{subfigure}
  \hfill
  \begin{subfigure}[t]{0.32\linewidth}
    \centering
    \includegraphics[width=\linewidth]{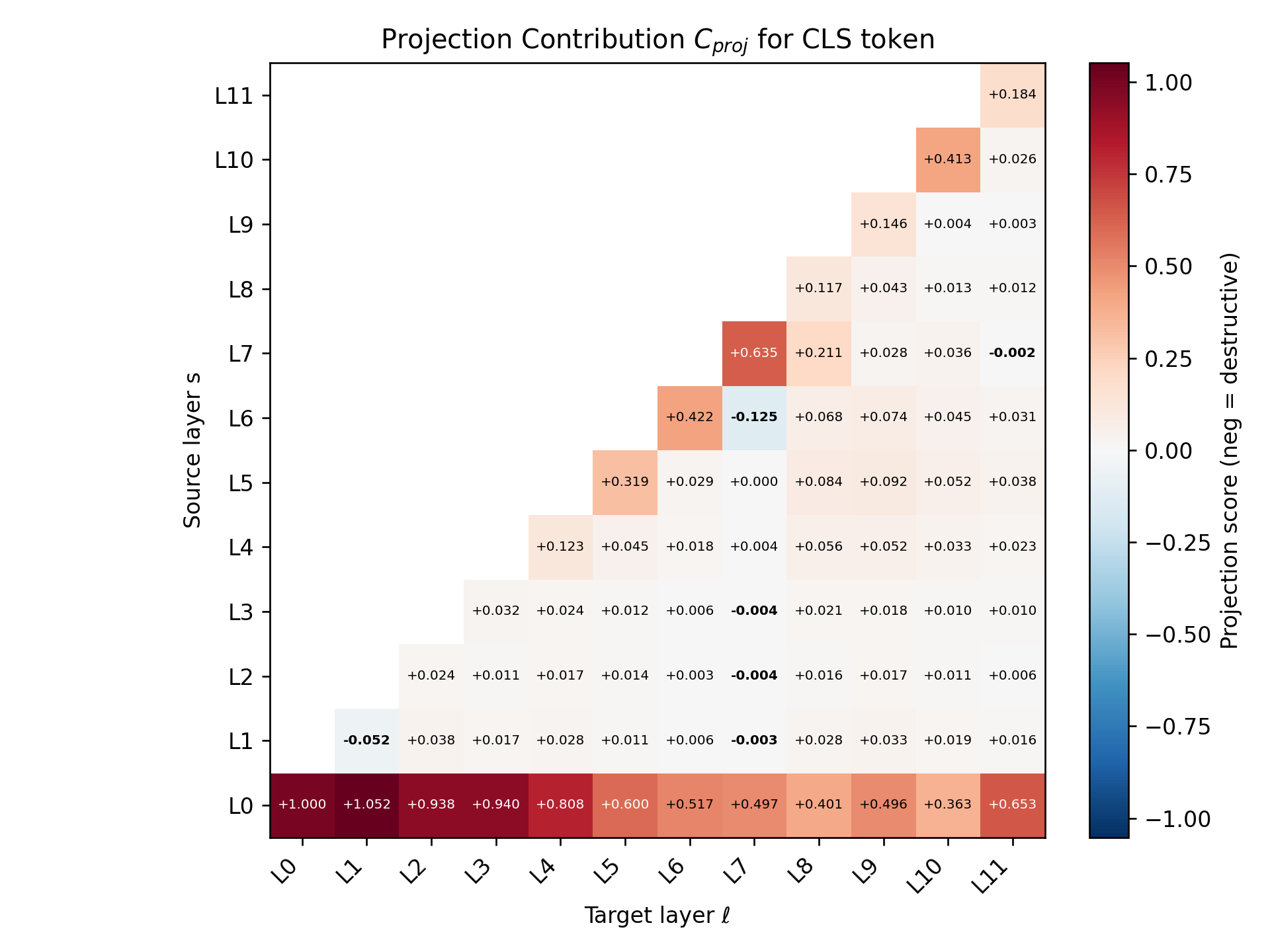}
    \caption{Abs-Top-$k$, CLS}
    \label{fig:contrib-v16-imnet-atk-cls}
  \end{subfigure}

  \vspace{0.6em}

  \begin{subfigure}[t]{0.32\linewidth}
    \centering
    \includegraphics[width=\linewidth]{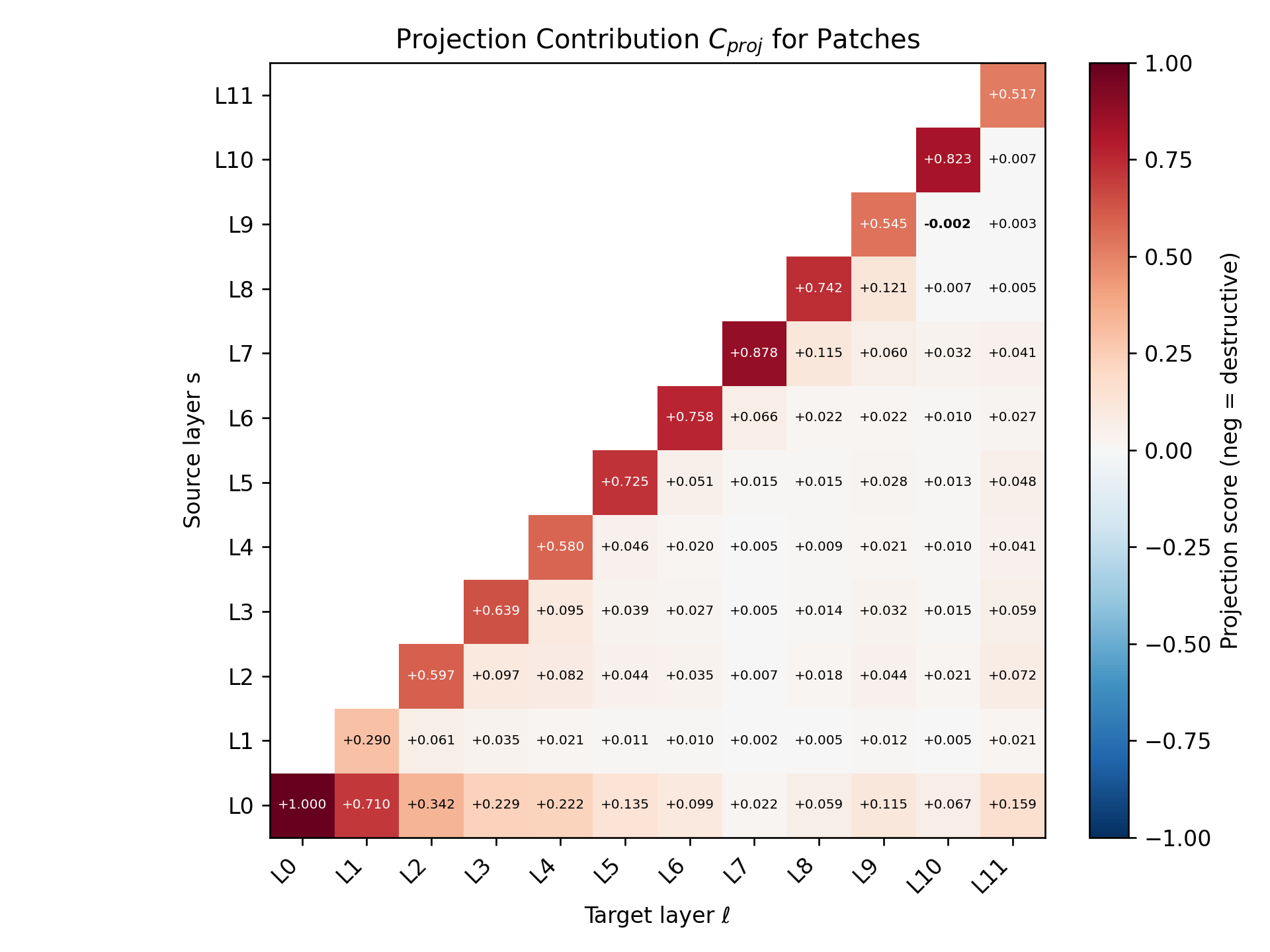}
    \caption{JumpReLU, patches}
    \label{fig:contrib-v16-imnet-jr-patches}
  \end{subfigure}
  \hfill
  \begin{subfigure}[t]{0.32\linewidth}
    \centering
    \includegraphics[width=\linewidth]{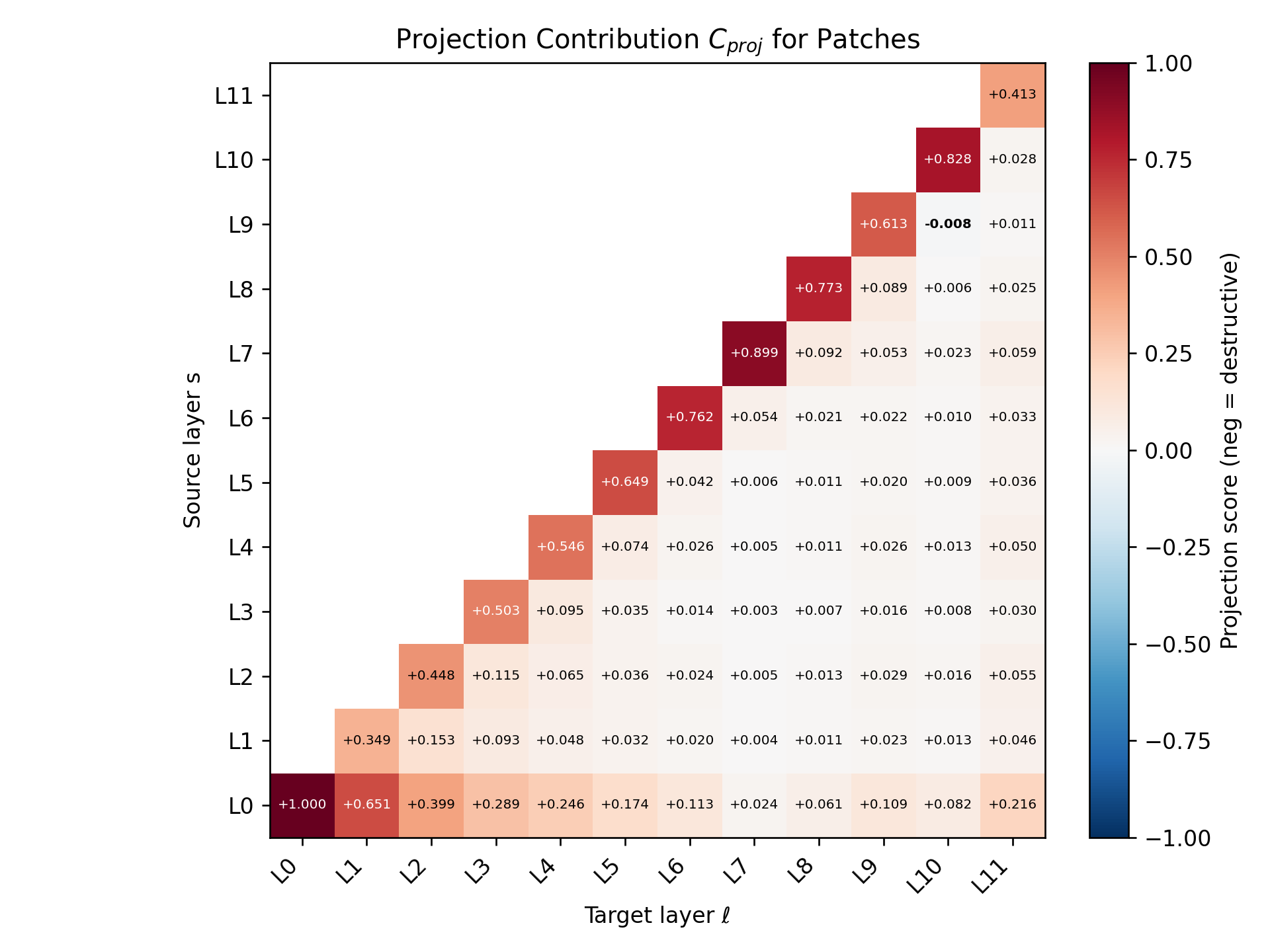}
    \caption{ReLU-Top-$k$, patches}
    \label{fig:contrib-v16-imnet-rtk-patches}
  \end{subfigure}
  \hfill
  \begin{subfigure}[t]{0.32\linewidth}
    \centering
    \includegraphics[width=\linewidth]{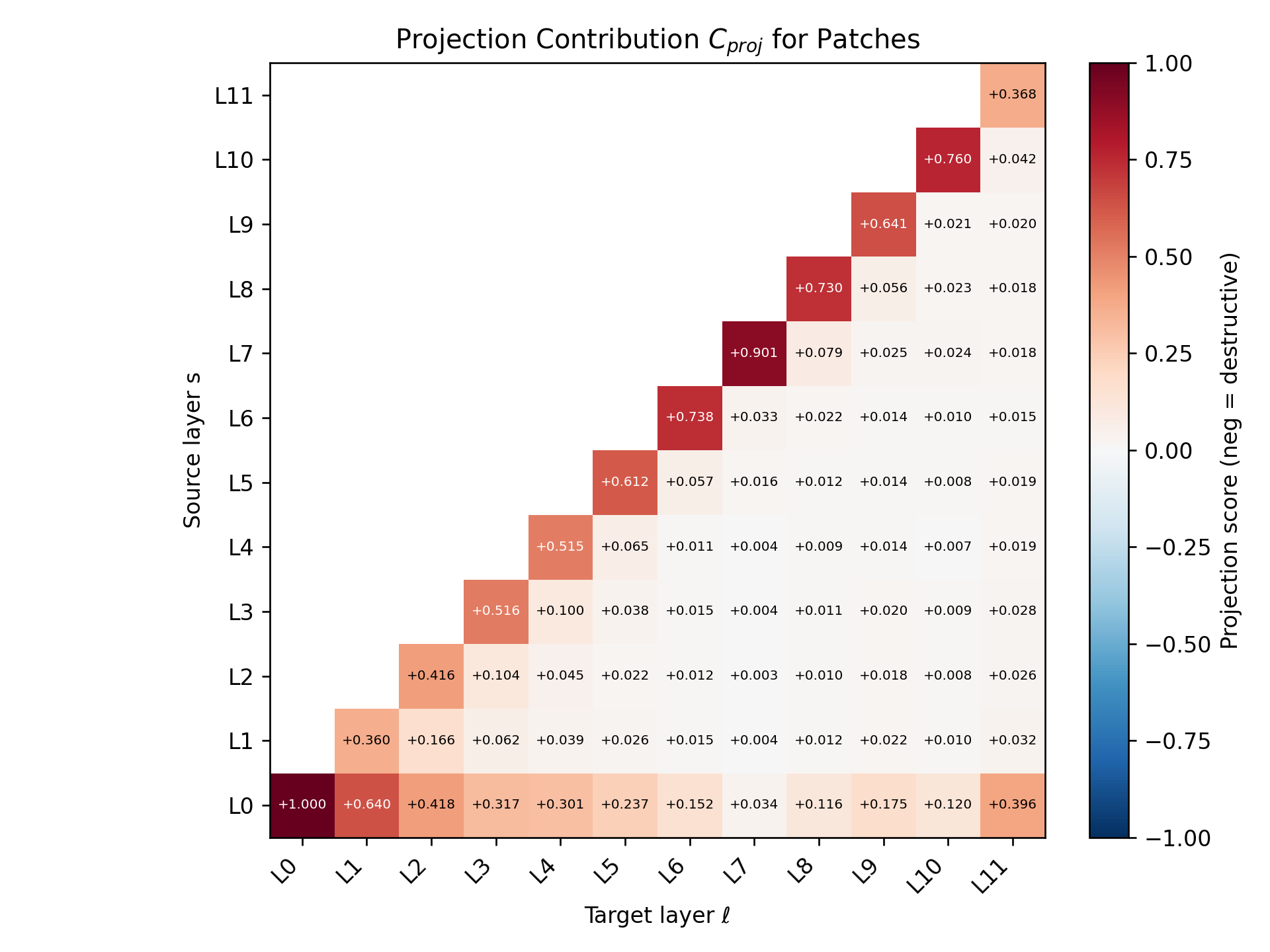}
    \caption{Abs-Top-$k$, patches}
    \label{fig:contrib-v16-imnet-atk-patches}
  \end{subfigure}
\caption{
  Cross-layer contribution scores $C_{s \rightarrow \ell}$ on ImageNet-100 with ViT-B/16. Columns vary the sparsifier (JumpReLU, ReLU-Top-$k$, Abs-Top-$k$)
  and rows show CLS (top) and patch tokens (bottom). Each heatmap visualizes the
  proportional contribution of source layer $s$ to the reconstructed activation at
  target layer $\ell$, averaged over the validation set.
  }
  \label{fig:contrib-v16-imnet-grid}
\end{figure*}

To better understand the internal attribution structure of Cross-Layer Transcoders (CLTs), we visualize in Figures \ref{fig:contrib-v32-cifar100-grid}–\ref{fig:contrib-v16-imnet-grid} the contribution scores $C_{s \rightarrow \ell}$, which quantify the influence of each source layer $s$ on the reconstruction of activations at target layer $\ell$. These heatmaps reveal a clear depth-aware structure across all datasets and backbones, where contributions are strongest from temporally proximal layers. Notably, CLS tokens exhibit more distributed contributions spanning earlier layers, while patch tokens show sharper, more localized attribution.


\end{document}